\documentclass{article} 
\usepackage{iclr2026_conference}
\usepackage{times}

\usepackage{amsmath,amsfonts,bm}

\def\eqref#1{equation~\ref{#1}}

\def\1{\bm{1}}

\DeclareMathAlphabet{\mathsfit}{\encodingdefault}{\sfdefault}{m}{sl}
\SetMathAlphabet{\mathsfit}{bold}{\encodingdefault}{\sfdefault}{bx}{n}

\usepackage{times} 
\usepackage[utf8]{inputenc} 
\usepackage[T1]{fontenc}    
\usepackage{hyperref}       
\usepackage{url}            
\usepackage{booktabs}       
\usepackage{siunitx}
\usepackage{amsfonts}       
\usepackage{nicefrac}       
\usepackage{microtype}      
\usepackage{xcolor}         
\usepackage{amsmath}
\usepackage{array}
\usepackage{graphicx} 
\usepackage{caption}  
\usepackage{amssymb}
\usepackage{subcaption}
\usepackage{enumitem}
\usepackage{amsthm, algorithm, algpseudocode, mathtools} 
\usepackage{multirow}
\usepackage{longtable} 
\usepackage{pifont}
\usepackage{wrapfig} 
\usepackage{xltabular}
\usepackage[nameinlink,capitalize,noabbrev]{cleveref}

\usepackage{chngcntr}

\newcolumntype{A}{S[table-format=3.2(2)]}   
\newcolumntype{D}{S[table-format=+3.2(2)]}  

\title{On the Stability of Growth in Structural Plasticity: Forward-Active yet Backward-Starved}

\author{Lute Lillo \& Nick Cheney \\
Department of Computer Science\\
University of Vermont\\
Burlington, VT 05401, USA \\
\texttt{\{elillopo,ncheney\}@uvm.edu}
}

\iclrfinalcopy 

\begin{document}
\maketitle
\vspace{-1.0em}
\begin{abstract}
\vspace{-1.0em}
Standard deep-learning pipelines usually choose the network architecture before training and keep it fixed throughout optimization.
In contrast, a model can also be adapted by editing its structure during training, for example by pruning existing hidden units or growing new ones; however, growth is not simply the inverse of pruning. 
Pruning selects among units trained from initialization, whereas growth inserts new capacity into an already specialized optimization trajectory. 
We isolate this insertion problem and show that newborn units can be forward-active yet backward-starved, receiving substantially weaker gradient signal than incumbent units. 
This asymmetry is weak in small MLPs but emerges in more challenging convolutional settings, where \textsc{Grow} reaches competitive final architectures despite weaker trajectory-level performance. 
Extending structural edits to residual networks reveals a strong dependence on edit location: head-localized growth produces concentrated but task-aligned representations and faster bounded label remapping, whereas growth throughout the residual hierarchy yields lower effective feature support, and a lower bounded adaptation endpoint.
Interventions targeting optimizer state, insertion, selection, and trainability improve newborn integration, but not necessarily final subnetwork quality. 
Across continual-learning benchmarks, growth is most competitive when newborn 
units have sufficient time and trainability to integrate. 
Therefore, \textsc{Grow} should be evaluated not only by the architecture it ultimately discovers, but by whether late-arriving capacity becomes usable before a new structural change occurs during training.
\end{abstract}

\vspace{-1.5em}
\section{Introduction}
\label{sec:introduction}
\vspace{-1.0em}

Structural plasticity---the ability to modify network architectures during training---is a natural primitive for automated machine learning under explicit resource constraints such as parameters, FLOPs, or latency. 
We distinguish this architectural notion from the classical use of \emph{plasticity}, which refers to a network's capacity, from its current state, to improve under a changed learning signal. 
The two concepts are related but not equivalent: changing a model's structure creates new capacity, but that capacity contributes to plasticity only if it can be effectively integrated and optimized.
The two basic operators of structural plasticity are \emph{pruning}, which removes capacity from an overparameterized model \citep{han2015learning,han2015deep}, and \emph{growth}, which adds capacity to a compact one \citep{yang2021grown,wu2020firefly,dai2019nest,rusu2016progressive}. 
Both can be viewed as search operators in architecture space, and dynamic sparse training (DST) shows that structure can be updated online while respecting a fixed parameter budget \citep{lasby2023dynamic,mocanu2018set,bellec2017deep,dettmers2019sparse}. 
In practice, however, structural adaptation remains dominated by pruning-based approaches.

Growth and pruning are not optimization-symmetric. 
Pruning selects among units that have participated in training from initialization, consistent with the lottery-ticket view that dense networks may contain sparse, trainable subnetworks \citep{frankle2019lottery,frankle2020stabilizing}. 
Growth adds capacity as needed, making it attractive for adaptive and continual systems; here we study shared-parameter growth, where incumbent and newborn units remain jointly trainable rather than modular or frozen \citep{miconi2016neural,yoon2017lifelong,li2019learn,yang2021grown}. 
Because newborn units enter after specialization, they must integrate into a mature optimization trajectory before the next shift; otherwise, repeated expansion can itself destabilize learning \citep{zhao2024overcoming}. 
Consequently, \textsc{Grow}--\textsc{Prune} comparisons can conflate final sparse-architecture quality with the efficiency of the insertion process used to obtain it.

Therefore, we study growth as a \emph{structural plasticity primitive} and frame this process through \emph{insertion stability}: the inability of a model to absorb a structural addition within the available training window.
From this perspective, newly added units face three birth-time disadvantages, depending on the growth mechanism: (i) \emph{function shock}, where insertion perturbs the learned input--output mapping; (ii) optimizer \emph{cold start} when new parameters lack optimizer state; and (iii) \emph{weak learning signal}, where newborn units receive disproportionately little credit relative to incumbent units. 
Throughout, ``units'' refers to hidden neurons added or removed at the neuron level, not individual connections; our study therefore concerns unit-level structural edits rather than unstructured synapse-level rewiring \citep{cheney2017robustness}.

The central claim is not that growth is primarily limited because it fails to discover useful sparse architectures, but that growth is a time-sensitive optimization process whose success depends on whether late-arriving capacity gets incorporated before the next change in the available training signal. 
This can make the adaptive process less stable and more path-dependent, even when the final retrained mask is competitive with pruning. 
We isolate this insertion primitive and investigate when \textsc{Grow} and \textsc{Prune} differ, asking whether this discrepancy stems from the quality of the final sparse architecture or the adaptive process used to generate it.
\footnote{Code available at: \url{https://anonymous.4open.science/r/structural_plasticity-1544}} 

\begin{itemize}
    \item We show that the \textsc{Grow}--\textsc{Prune} gap is not monolithic: in small MLPs, growth and pruning produce similarly retrainable masks, while in convolutional feature-learning regimes they can produce similarly strong final subnetworks but differ substantially in how well the model behaves as its architecture is modified (Sec.~\ref{sec:results}).
    Then, in residual networks, edit locus further determines future adaptability: head-localized \textsc{Grow} adapts fastest under class-permutation remapping, while broader residual growth reaches a lower bounded endpoint (Sec.~\ref{sec:resnet_locus}).
    \item We identify insertion-time optimization disadvantage as a process-level bottleneck for growth, showing that newborn units can be forward-active while remaining backward-starved (Sec.~\ref{sec:allocation_mechanisms}).
    \item We use interventions on optimizer state, insertion, selection, and activation-level trainability as probes of this bottleneck (Sec.~\ref{sec:integration_interventions}), and show that under continual shift, growth is most effective when new units have time to integrate before the next distributional change (Sec.~\ref{sec:cl_extension}).
\end{itemize}

\vspace{-1.5em}
\section{Background \& Related Work}
\label{sec:background_related_work}

\vspace{-1.0em}
\paragraph{Dynamic sparsity and structural operators.}
Dynamic sparse training (DST) methods maintain a fixed parameter budget while updating sparse connectivity online, combining pruning and regrowth as intertwined structural operators \citep{mocanu2018set,bellec2017deep,dettmers2019sparse}. 
This line of work views pruning and growth as search moves in architecture space, and highlights that the \emph{allocation rule}---which connections or units receive structure---interacts strongly with learning dynamics \citep{evci2020rigl}. 
We do not focus on the allocation rule, but on the stability of the \emph{insertion} event itself.

\vspace{-1.0em}
\paragraph{Insertion-compatible growth.}
Architectural expansion can disrupt a learned function and introduce optimization asymmetries between incumbent and newborn parameters.
Net2Net and Network Morphism widen networks while approximately preserving the current input--output mapping \citep{chen2016net2net,wei2016networkmorphism}, while MorphNet couples structural adaptation to resource constraints \citep{gordon2018morphnet}.
Other methods use activations or gradients to select expansion locations \citep{wu2019splitting,evci2022gradmax}, or modify the optimization of late-arriving parameters through variance transfer and age-dependent learning rates \citep{yuan2023accelerated}. 
These approaches target distinct parts of the insertion process. 
More broadly, warm-up, layer-wise modulation, and adaptive optimizer state all point to the same issue: a newly inserted unit may be disadvantaged not only by where it is placed, but by \emph{when} it enters training \citep{you2019large,xiong2020layer,mosbach2020stability,kingma2014adam,reddi2019convergence,zhuang2020adabelief}.
We study these mechanisms jointly through event-aligned unit diagnostics and targeted interventions.

\vspace{-1.0em}
\paragraph{Plasticity and plasticity loss under continual shift.}
Two distinct concepts are relevant. 
Following the continual-learning literature, \emph{plasticity} refers to a network's capacity, from its current state, to continue learning---that is, to reduce loss and acquire new mappings when the learning signal changes \citep{lyle2023understanding, abbas2023loss}. 
\emph{Loss of plasticity} refers to the decline of this ability over the course of training \citep{dohare2024loss, klein2024plasticity}, and is different from forgetting, which concerns the deterioration of performance on information that was previously learned \citep{mccloskey1989catastrophic, kirkpatrick2017overcoming}.
In contrast, \emph{structural plasticity} denotes the capability to modify the network architecture itself during training. 
Consequently, architectural expansion increases the structural capacity of the model, but it enhances plasticity only if the newly introduced units are functionally integrated into the network and remain effectively utilizable.

\vspace{-1.0em}
\paragraph{Growth in continual learning.}
Many continual-learning methods rely on architectural expansion to accommodate new tasks, with design choices centered on \emph{when}, \emph{where}, and \emph{what} to grow \citep{rusu2016progressive,fernando2017pathnet,yoon2017lifelong,li2019learn,yang2021grown}. 
Growth is appealing because it allocates fresh capacity for new information, potentially easing the stability--plasticity dilemma. 
At the same time, strong isolation-based approaches such as PackNet, Piggyback, and winning-subnetwork methods achieve low forgetting by assigning task-specific subnetworks within shared weights \citep{mallya2018packnet,mallya2018piggyback,wsn-pmlr-v162-kang22b}, often at the cost of task identity, routing, or selection at inference time. 
Recent work further shows that poorly controlled expansion can itself induce forgetting in task-agnostic settings \citep{zhao2024overcoming}. 
Our work instead asks whether newly added units can integrate stably within a shared, fully plastic representation, rather than through task-isolated or frozen modules.
\vspace{-1.0em}
\section{Experimental Setup}
\label{sec:methods}

\vspace{-1.0em}
We compare three model families under matched data streams, optimizers, and compactness targets:
\begin{itemize}[leftmargin=1.25em,itemsep=0.2em]
\vspace{-0.5em}
    \item \textbf{\textsc{Dense}} uses the full architecture without structural
    edits.
    \item \textbf{\textsc{Prune}} applies Iterative Magnitude Pruning (\textsc{IMP}) to editable layers and rewinds surviving weights after each pruning event. 
    This canonical lottery-ticket baseline separates the quality of the selected sparse mask from the trajectory used to obtain it (App.~\ref{app:prune}).
    \item \textbf{\textsc{Grow}} starts from a sparse seed mask and iteratively \emph{activates} dormant units until reaching the target compactness.
    Candidates are ranked by how frequently their post-activations would exceed a small threshold if unmasked, estimating whether they would participate meaningfully if recruited (App.~\ref{app:grow}).
    Growth updates the mask in place without rewind: dormant parameters remain at initialization until activation, newborn units do not inherit optimizer state, and all active parameters remain jointly trainable rather than frozen or task-isolated.
    Gradient-based recruitment gives similar qualitative conclusions (App.~\ref{app:app_gradient_grow_ablation}).
\vspace{-0.5em}
\end{itemize}

We begin with a three-layer MLP containing two editable hidden layers and an unmasked classifier, so structural edits change hidden capacity without altering the output mapping. 
\textsc{Grow} and \textsc{Prune} follow opposite compactness trajectories, with \textsc{Grow} increasing from 5\% and \textsc{Prune} decreasing from 100\% until reaching retained compactness $c\in\{20,30,40,50\}\%$. 
During the first \(T-1\) cycles, the method adds or removes units and then trains; the final cycle trains at the target compactness.

Each procedure produces a final binary mask, which we evaluate in two ways:
\begin{enumerate}
\vspace{-0.5em}
    \item \textbf{In-place structural-edit}: performance measured during the structural-editing trajectory itself, measured at the end of each cycle. 
    It captures how well the model performs while architecture changes are being made online during training.
    \item \textbf{Winning-Ticket (WT)}: performance measured after freezing the final discovered mask, rewinding the model, and retraining it from scratch \citep{frankle2019lottery}. 
    This tests the sparse architecture independently of the path used to discover it.
\vspace{-0.5em}
\end{enumerate}

For both evaluations, we report final cumulative accuracy after the last task (\textbf{ACC}) reflecting end-of-stream performance.

\vspace{-1.0em}
\section{Feature Learning Exposes a Newborn-Integration Bottleneck}
\label{sec:results}

\vspace{-0.5em}
\paragraph{Minimal MLPs do not expose a stable growth gap.}
On IID MNIST \citep{lecun1998mnist}, all methods achieve similarly high accuracy. 
Under class-incremental Split-MNIST with a shared 10-way head, all methods collapse near chance without stabilization, indicating that forgetting dominates the architectural comparison. 
Therefore, we add a small experience replay buffer (\textsc{+ER}; $50$ samples per class) to make structural differences interpretable, and use the SGD optimizer with learning rate $\eta=0.01$. 
Across compactness levels, \textsc{Grow}, \textsc{Prune}, and \textsc{Dense} remain closely matched in both In-place structural-edit and Winning-Ticket (WT) metrics, with no consistent ordering; the same holds for Split-Fashion-MNIST \citep{xiao2017fashion}. In particular, \textsc{Grow} and \textsc{Prune} produce final masks that retrain to similar winning-ticket accuracy and TAA. Thus, neither sparsity nor class-incremental training alone produces a stable \textsc{Grow}--\textsc{Prune} asymmetry.
Complete results are reported in Table~\ref{tab:app_ticket_delta_all_datasets}.
\vspace{-0.5em}
\subsection{Newborn units are forward-active but backward-starved}
\label{sec:scaling_convnet}
\vspace{-0.5em}

We next move to CIFAR-100 \citep{Krizhevsky2009LearningML} to increase representation-learning and optimization difficulty while retaining the same unit-level structural abstraction. 
We use a hybrid ConvNet with a dense two-layer convolutional trunk and a four-layer fully connected head. 
Only the hidden layers of the head are editable; the convolutional trunk remains dense. 
Models use SGD with $\eta=0.1$, selected independently for each treatment, and the neutral allocation schedule described in App.~\ref{sec:app_cifar100_bias_ablation}.


\vspace{-1.0em}
\begin{wrapfigure}{l}{0.49\textwidth}
    \vspace{-1.2em}
    \centering
    \includegraphics[width=0.49\textwidth]{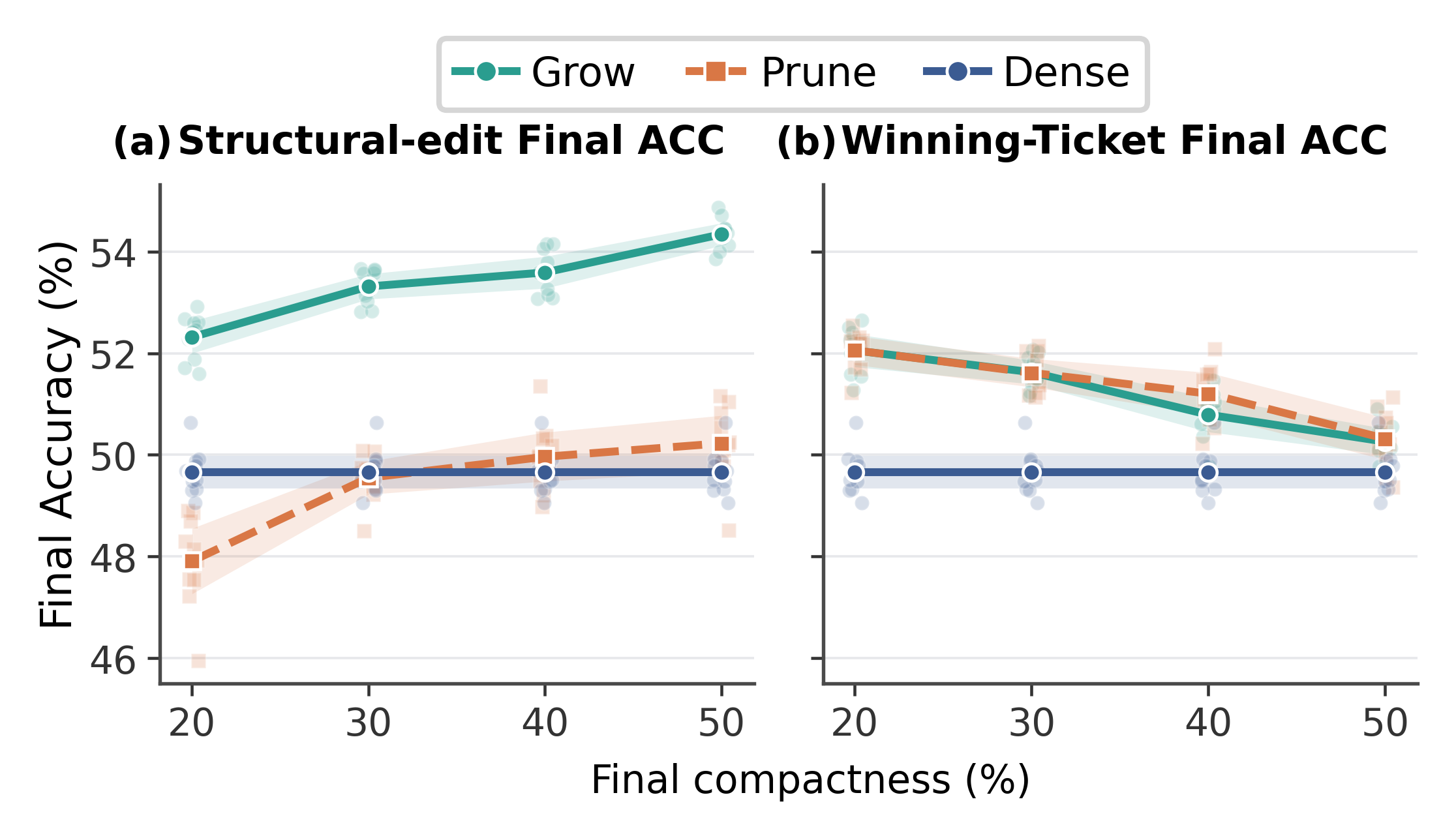}
    \vspace{-2.0em}
    \caption{
        \textbf{\textsc{Grow} reaches a substantially stronger structural-edit endpoint despite discovering no better final architecture.}
        \textbf{(a)} During structural adaptation, \textsc{Grow} consistently exceeds \textsc{Prune}, reaching $54.3\%$ accuracy at $50\%$ compactness versus $50.2\%$ for \textsc{Prune} and $49.7\%$ for \textsc{Dense}.
        \textbf{(b)} After rewinding and retraining the discovered masks, however, \textsc{Grow} and \textsc{Prune} become indistinguishable.
        Thus, the advantage comes from the growth trajectory, not a better final sparse architecture, prompting our analysis of how newborn capacity integrates during training.
    }
    \label{fig:cifar100_cycle_vs_ticket_sgd}
    \vspace{-1.5em}
\end{wrapfigure}

\paragraph{Edit-time learning differs from final mask quality.}
Figure~\ref{fig:cifar100_cycle_vs_ticket_sgd} shows that \textsc{Grow} reaches the strongest final accuracy during in-place structural training across compactness levels, outperforming both \textsc{Prune} and \textsc{Dense}.
To separate this trained-state advantage from the quality of the discovered compact architecture, we use the standard winning-ticket test: when final masks are rewound and retrained from scratch, \textsc{Grow} and \textsc{Prune} are statistically indistinguishable.
This comparison is conservative with respect to \textsc{Grow}: \textsc{Prune} begins at full capacity and subsequently decreases capacity, thereby experiencing substantially greater exposure to active parameters over the course of the structural trajectory, whereas \textsc{Grow} increases capacity from an initially sparse seed; excluding IMP rewinding reduces this discrepancy only modestly (App.~\ref{app:prune_protocol}). Thus, the superior in-place endpoint achieved by \textsc{Grow} cannot be attributed solely to a higher-quality final mask, motivating a study of how newly added capacity is integrated across training.


\vspace{-1.0em}
\subsection{Edit-time integration of newborn units}
\label{sec:allocation_mechanisms}
\vspace{-0.5em}

The gap between final performance and the warm-started training trajectory in Sec.~\ref{sec:scaling_convnet} suggests that \textsc{Grow} successfully identifies high-quality final subnetworks, but is less effective at integrating them during the warm-start editing process. 
To pinpoint this bottleneck, we analyze each growth event, comparing newly activated units to previously active ones.
We measure forward participation by post-activation magnitude and backward credit by pre-activation gradient magnitude. 
For each newborn cohort, we compare activation and gradient signals to already-active units. 
A value of 100\% means equal signal; below 100\% means newborn units receive less activation or gradient signal than incumbent units.

\vspace{-1.0em}
\begin{figure}[ht]
    \centering
    \includegraphics[width=\linewidth]{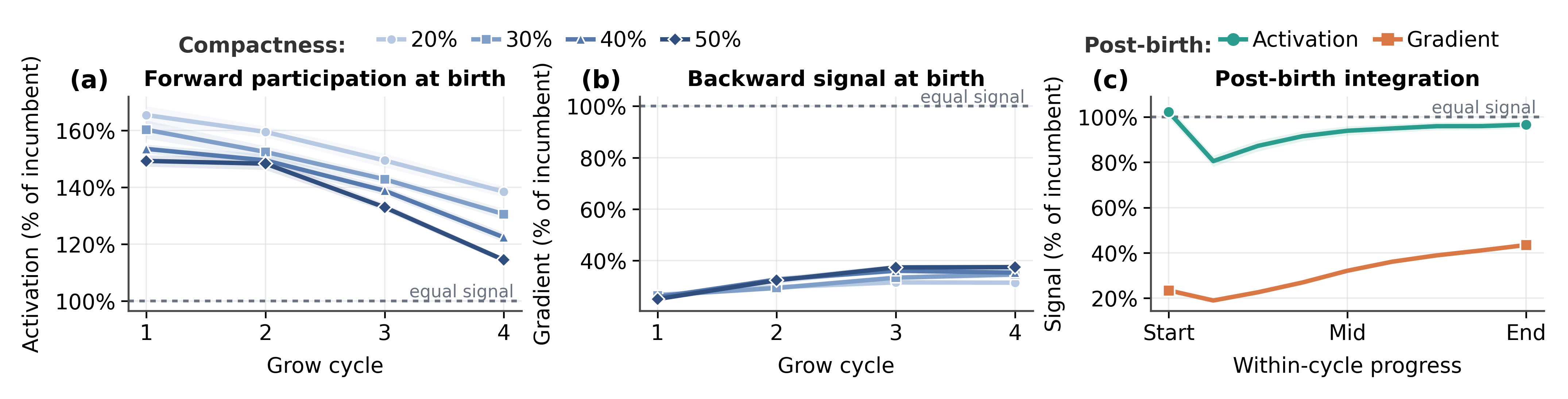}
    \vspace{-2.5em}
    \caption{
    \textbf{Newborn units become forward-active but remain gradient-disadvantaged.}
    Event-aligned diagnostics for the editable ConvNet head on CIFAR-100.
    \textbf{(a)} Newborn activation is near or above the incumbent level immediately after insertion.
    \textbf{(b)} In contrast, newborn units receive only a fraction of the incumbent gradient signal.
    \textbf{(c)} During subsequent training, forward participation approaches parity while backward signal remains substantially weaker.
    Lines and shaded regions show means and 95\% confidence intervals; dashed references denote equal signal.
    }
    \label{fig:cifar100_newborn_integration}
    \vspace{-15pt}
\end{figure}

\paragraph{Growth is limited by backward integration.}
Figure~\ref{fig:cifar100_newborn_integration} rules out a simple dead-unit explanation.
Forward participation is partly encouraged by the default activation-based selector, but the resulting backward disadvantage is not: newborn units receive substantially weaker gradients despite being active.
Newborn units are forward-active at birth and approach incumbent activation levels, yet receive substantially weaker gradients throughout the cycle, indicating that architectural capacity appears before comparable credit assignment.
This helps explain why \textsc{Grow} can discover winning tickets comparable to \textsc{Prune} while exhibiting weaker edit-time performance.
By contrast, \textsc{Prune} initially removes less-active units, though this selectivity weakens and survivor readjustment increases across cycles (App.~\ref{app:prune_protocol}).
The bottleneck also persists under gradient-based recruitment and worsens with more frequent growth under a fixed training budget (App.~\ref{app:app_gradient_grow_ablation}).
Because these diagnostics edit only the classifier head, Sec.~\ref{sec:resnet_locus} tests whether the same limitation strengthens when growth enters the learned feature hierarchy.
\vspace{-0.5em}
\subsection{Where can new structure integrate? Growth inside a feature hierarchy}
\label{sec:resnet_locus}
\vspace{-0.5em}

Our previous experiments isolate structural allocation in a fully connected classifier head, leaving open whether \textsc{Grow}–\textsc{Prune} asymmetry persists when edits modify the learned feature hierarchy. 
We study this on CIFAR-100 using a CIFAR-style ResNet where structural units are channels inside residual branches. 
This lets us distinguish \emph{initial structural plasticity} from \emph{realized structural plasticity}: activating a dormant channel changes the architecture immediately, but yields useful plasticity only if the channel is integrated into the learned representation and aids further optimization.

\vspace{-0.5em}
\paragraph{Residual-channel editing.}
For residual block $b$, let
\[
u_b = \phi(\mathrm{BN}_{b,1}(W_{b,1}h_b)),
\qquad
\tilde{u}_b = m_b \odot u_b,
\qquad \text{and} \qquad
h_{b+1}
=
\phi\!\left(
\mathrm{BN}_{b,2}(W_{b,2}\tilde{u}_b)
+
S_b(h_b)
\right),
\]
where $m_b\in\{0,1\}^{d_b}$ is a binary channel mask applied after the first convolution and nonlinearity, and $S_b$ denotes the identity or projection shortcut.
Therefore, a masked structural unit comprises an output filter of $W_{b,1}$ and the corresponding input-channel slice of $W_{b,2}$. 
The shortcut pathway remains intact, so edits modify the residual transformation without deleting the skip connection.
We compare four edit loci:
(i) \textsc{Head}, which edits only the classifier head;
(ii) \textsc{Late}, which edits residual channels in the final stage;
(iii) \textsc{Mid+Late}, which edits the final two stages; and
(iv) \textsc{All}, which edits every residual stage.
All methods use the same optimization budget and per-mask active-channel target.
Thus, compactness is matched within editable masks, not in whole-network parameter count: broader loci place structural pressure on more of the feature hierarchy (App.~\ref{app:budgets_compactness}).
Accordingly, cross-locus trends reflect both edit depth and the amount of feature-extractor capacity being edited.


\vspace{-0.5em}
\paragraph{Current-task performance degrades as editing broadens.}
At $50\%$ compactness, extending structural edits through more of the residual hierarchy lowers in-place ACC for both methods:
\textsc{Grow} falls from $61.2\%$ under \textsc{Head} to $58.5\%$ under \textsc{All}, while \textsc{Prune} falls from $61.9\%$ to $58.0\%$.
Within each locus, however, their endpoints remain close, and the same holds after retraining the final masks from scratch; under \textsc{All}, WT accuracy is $59.2\%$ for \textsc{Grow} and $59.0\%$ for \textsc{Prune}.
Thus, broader editing reduces current-task performance, but endpoint accuracy alone reveals little \textsc{Grow}--\textsc{Prune} separation.
We next ask whether similarly performing models nevertheless realize their structural capacity differently in the learned representation.
Complete results are reported in App.~\ref{app:resnet_summary}.

\vspace{-0.5em}
\paragraph{Realized representations: scale, support, and utility.}
Adding channels increases architectural capacity, but it need not produce a broader or more useful representation.
Therefore, we ask whether growth at different edit loci changes (i) how much the learned features vary, (ii) how broadly that variation is distributed across feature directions, and (iii) how easily current-task information can be read out.
At the final structural cycle, we spatially average \texttt{stage3\_out}---the output of the final residual stage---to obtain one $64$-dimensional feature vector per example (Apps.~\ref{app:nn_archs}, \ref{app:resnet_representation_robustness}).
Let $H\in\mathbb{R}^{n\times d}$ contain held-out feature vectors, let $\bar H$ denote their centered form, and define $\Sigma = \frac{1}{n-1}\bar H^\top \bar H$.
First, total covariance variance, $V(H)=\operatorname{tr}(\Sigma)=\sum_i\lambda_i$, is the variance summed across all feature directions, and therefore measures the overall spread of the representation across examples.
Second, normalized effective rank measures how broadly that variation is distributed across directions, following entropy-based effective dimensionality measures used in representation and plasticity analyses \citep{roy2007effective,dohare2024loss}.
With $p_i = \frac{\lambda_i}{\sum_j\lambda_j+\epsilon}$, we compute $\widetilde r_{\mathrm{eff}}(H) =
\frac{1}{r_{\max}} \exp\!\left(-\sum_i p_i\log(p_i+\epsilon)\right),$ where $r_{\max}$ is the maximum supported covariance rank.
Unlike total variance, this quantity is insensitive to uniform rescaling: high values indicate variation spread across many directions, low values indicate concentration into a smaller feature subspace.
Third, we train a linear classifier on frozen features and report held-out accuracy, measuring whether current-task information is accessible through a simple readout independent of the original classifier.

\begin{figure*}[ht]
    \centering
    \includegraphics[width=\textwidth]{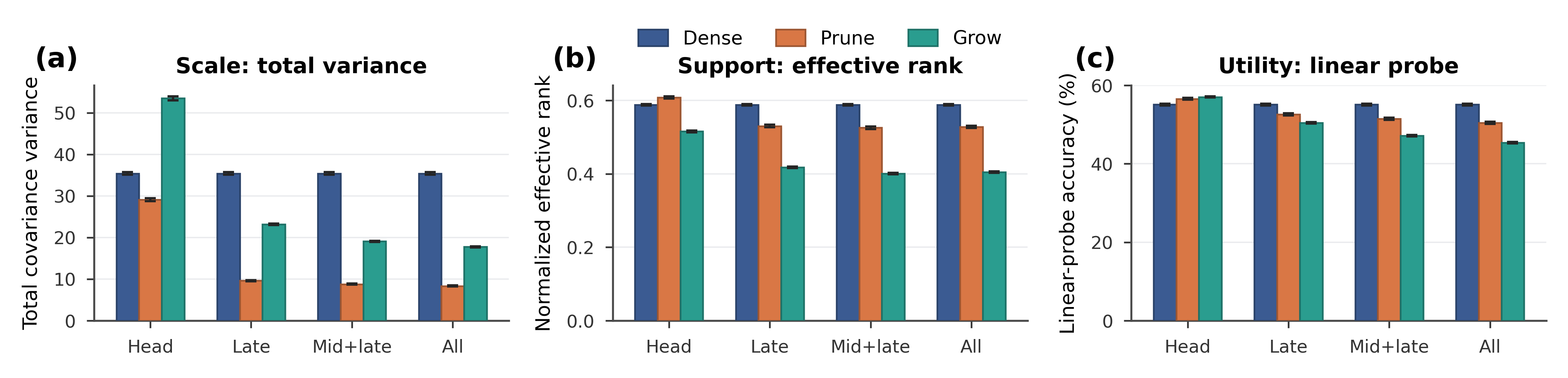}
    \vspace{-2.0em}
    \caption{
        \textbf{Does added structural capacity become useful representational capacity?}
        We examine the final-cycle \texttt{stage3\_out} representation at $50\%$ compactness along three complementary axes:
        (a) \emph{scale}, how much the features vary across examples;
        (b) \emph{support}, how broadly that variation is distributed across feature directions; and
        (c) \emph{utility}, how well current-task information can be recovered with a frozen linear probe.
        \textsc{Head}-localized \textsc{Grow} increases scale and linear utility despite concentrating variation into fewer directions, whereas growth through the residual hierarchy yields narrower support and weaker linear utility.
        Error bars show $95\%$ confidence intervals.
        }
    \label{fig:resnet_representation_state}
\end{figure*}

Figure~\ref{fig:resnet_representation_state} reveals two growth regimes. 
Under \textsc{Head} editing, \textsc{Grow} raises total variance from $35.4$ to $53.5$ and linear-probe accuracy from $55.1\%$ to $57.0\%$, despite a lower normalized effective rank than \textsc{Dense} ($0.516$ versus $0.588$). 
Therefore, \textsc{Head} growth amplifies a concentrated set of task-relevant directions rather than creating a uniformly broader feature basis. 
Growth within the residual hierarchy behaves differently. 
Under \textsc{All}, \textsc{Grow} retains more variance than \textsc{Prune} ($17.7$ versus $8.3$), but has lower effective rank ($0.405$ versus $0.528$) and probe accuracy ($45.4\%$ versus $50.5\%$), with the same pattern under \textsc{Late} and \textsc{Mid+Late}. 
Thus, \textsc{Grow} produces a larger but more directionally concentrated feature cloud than \textsc{Prune}: added channels increase architectural capacity without yielding broadly usable feature support.

\vspace{-1.1em}

\vspace{-1.5em}
\subsubsection{Bounded future reachability.}
\vspace{-0.5em}
We use \emph{bounded future reachability} to mean how readily a trained checkpoint can be adapted to a changed learning signal within a fixed optimization budget.
Specifically, current-task representation quality does not necessarily determine whether a checkpoint remains useful for future learning.
Therefore, we ask whether structural growth changes the \emph{speed and endpoint of subsequent adaptation}.
Using a controlled \emph{class-permutation remapping probe}, we clone selected checkpoints and fine-tune them on a fixed permutation of CIFAR-100 labels. Inputs and visual classes stay the same while output assignments change, isolating functional remapping—whether learned features can be quickly reused and reorganized without adapting to a new visual domain.
All parameters are fine-tuned for the same budget of $B=2400$ gradient steps with batch size $256$.
Let $0=b_0<\cdots<b_J=B$ denote adaptation steps and $a_{b_j}$ the corresponding probe accuracies. 
We report the normalized area under the adaptation curve,
$\mathrm{AUC}_B(\theta)=\frac{1}{B}\sum_{j=0}^{J-1}
\frac{a_{b_j}+a_{b_{j+1}}}{2}(b_{j+1}-b_j)$, which rewards checkpoints that achieve useful performance earlier within the fixed budget. 
Final probe accuracy separately measures the adaptation endpoint.

\vspace{-1.0em}
\begin{figure*}[ht]
    \centering
    \includegraphics[width=\textwidth]{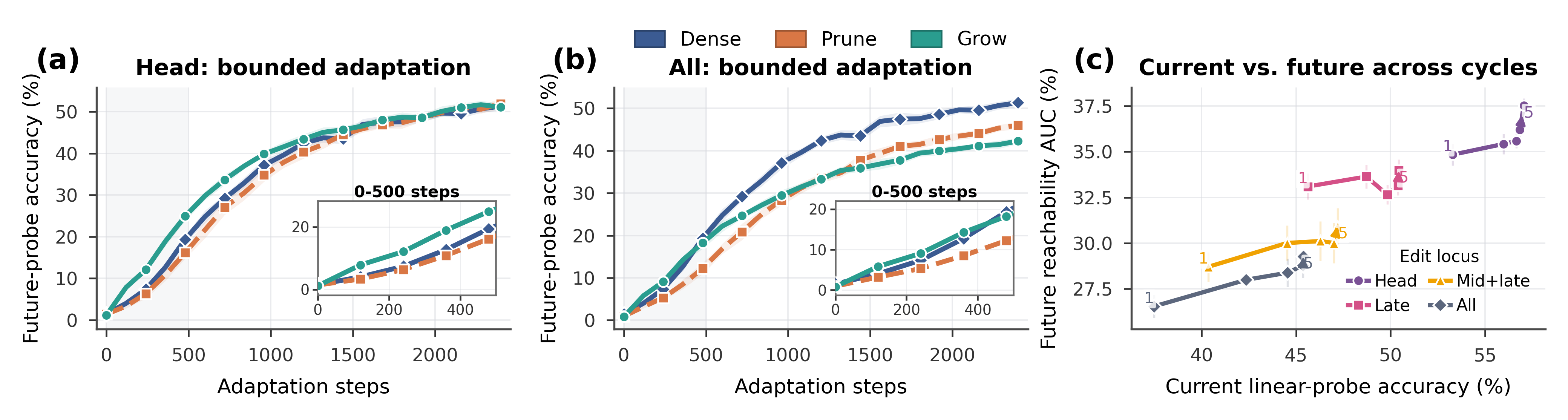}
    \vspace{-2.0em}
    \caption{
    \textbf{The effect of growth on future reachability is locus- and budget-dependent.}
    We clone final-cycle checkpoints at $50\%$ compactness and fully fine-tune them for $2400$ steps after a fixed label permutation.
    \textbf{(a)} \textsc{Head}-localized \textsc{Grow} adapts fastest during the early probe and reaches approximately the same endpoint as \textsc{Dense} and \textsc{Prune}.
    \textbf{(b)} When all residual stages are editable, \textsc{Dense} retains a substantially higher adaptation trajectory; \textsc{Grow} has a small early advantage over \textsc{Prune} but reaches a lower final accuracy.
    Insets magnify the first $500$ adaptation steps.
    \textbf{(c)} For \textsc{Grow}, each trajectory follows structural cycles $1$--$5$ in the plane defined by current linear-probe accuracy and future reachability AUC. Therefore, higher and farther right indicates greater current utility and future adaptability.
    Error bars and shaded regions show $95\%$ confidence intervals.
    }
    \label{fig:resnet_future_reachability}
\end{figure*}

\paragraph{Head-localized growth enables faster remapping.}
Figure~\ref{fig:resnet_future_reachability}(a) reveals a marked early advantage for \textsc{Grow} under \textsc{Head}: after $500$ steps it reaches roughly $25\%$ accuracy, versus $19\%$ for \textsc{Dense} and $16\%$ for \textsc{Prune}.
Yet all three finish near $51\%$ after $2400$ steps.
Thus, head-localized growth does not reach a better final solution; it reaches a comparable one substantially faster.
When edits extend throughout the residual hierarchy, this pattern changes (Fig.~\ref{fig:resnet_future_reachability}(b)):
\textsc{Grow} retains some early mobility but finishes at $41.9\%$, below \textsc{Prune} ($46.2\%$) and \textsc{Dense} ($51.3\%$).
Thus, \textsc{Grow} retains strong early mobility but broad residual editing lowers its bounded adaptation endpoint.

\vspace{-0.5em}
\paragraph{Current feature utility tracks this locus dependence.}
Figure~\ref{fig:resnet_future_reachability}(c) asks whether checkpoints that are more useful on the current task are also easier to repurpose.
Each point represents a \textsc{Grow} checkpoint, with current linear-probe accuracy on the $x$-axis and future-remapping AUC on the $y$-axis; connected points trace structural cycles within each edit locus.
Therefore, moving up and right means becoming both more useful now and easier to adapt later.
\textsc{Head} occupies the favorable upper-right region, while broader loci remain progressively lower in both quantities.
At the final cycle, AUC declines from $36.5$ under \textsc{Head} to $32.9$ (\textsc{Late}), $29.5$ (\textsc{Mid+Late}), and $28.4$ (\textsc{All}) (Table~\ref{tab:resnet-main-summary}).
Thus, where growth occurs strongly shapes both the representation available now and how readily it can be reused under a changed learning signal.

\vspace{-1.0em}
\section{Interventions on the newborn integration bottleneck}
\label{sec:integration_interventions}
\vspace{-1.0em}

Sections~\ref{sec:allocation_mechanisms} and \ref{sec:resnet_locus} identify two linked limitations of growth. 
Newborn units can participate in the forward computation while receiving weak backward signal, and this integration deficit becomes more consequential when growth modifies the learned feature hierarchy. 
Therefore, we evaluate four possible sources of this integration bottleneck: 
(i) \emph{optimizer state} mismatch due to lacking accumulated states, 
(ii) disruptive \emph{insertion} operations that perturb the current function, 
(iii) suboptimal \emph{selection} of inactive units, and 
(iv) restricted gradient flow from \emph{activation functions}. 
These interventions test which mechanisms improve newborn integration and whether the strongest gains transfer from the controlled ConvNet head to residual-channel growth.

\vspace{-0.5em}
\paragraph{Optimizer-state interventions: Two-Speed and Moment Transplant.}
Newborn parameters enter training without the optimization history accumulated by incumbent units.
We test this cold-start hypothesis with two complementary interventions. 
\textsc{Two-Speed}, inspired by learning-rate adaptation for incrementally grown networks \citep{yuan2023accelerated}, temporarily amplifies updates to the incoming and outgoing weights associated with newborn units.
If $\theta_t$ denotes these parameters and $\theta_{t+1}^{\mathrm{base}}$ is the update proposed by the base optimizer, we apply $\theta_{t+1}=\theta_t+r(\theta_{t+1}^{\mathrm{base}}-\theta_t)$ with $r>1$ during a short post-birth warm-up window. 
\textsc{Moment Transplant} instead initializes the newborn optimizer buffers from an incumbent donor unit. 
This tests whether missing accumulated state, rather than an insufficient update magnitude alone, limits early integration \citep{asadi2023resetting,behrouz2025nested}.

\vspace{-0.5em}
\paragraph{Insertion intervention: Net2Wider.}
A second possibility is that activating a new unit perturbs an already learned function before the unit has acquired a useful role. 
Therefore, we implement \textsc{Net2Wider} \citep{chen2016net2net}, which duplicates an incumbent unit and redistributes its outgoing weights so that the network function is preserved at insertion. 
This directly tests whether reducing insertion shock improves newborn integration. Function preservation does not, however, guarantee useful differentiation: the duplicate initially shares its donor's representation and may remain redundant or receive similarly weak credit.

\vspace{-0.5em}
\paragraph{Selection intervention: GradMax-style recruitment.}
The integration deficit may also depend on which dormant units are recruited.
Our default activation-based rule selects candidates according to their prospective forward participation. 
In contrast, the \textsc{GradMax}-style selector \citep{evci2022gradmax} scores dormant candidates by the learning signal they would receive on the mini-batch available at the growth event and activates the top-ranked units within each editable layer. 
This changes only the \emph{selection policy}; initialization, optimizer handling, and post-insertion training remain unchanged. 
Therefore, it tests whether weak newborn gradients arise because the default rule recruits candidates that are forward-active but poorly coupled to the current loss.

\vspace{-0.5em}
\paragraph{Activation-function intervention: Rand.\ Smooth-Leaky.}
Finally, the bottleneck may reflect restricted gradient transmission rather than optimizer state, insertion shock, or candidate selection. 
Because newborn units are forward-active yet gradient-disadvantaged, changing the nonlinearity provides a direct intervention on their backward pathway while leaving the structural procedure unchanged. 
We use \textsc{Rand.\ Smooth-Leaky}, a randomized smooth leaky activation function previously shown to preserve plasticity by maintaining non-zero negative-side gradients and a gradual transition around the origin \citep{lillo2025activation}. 
This tests whether improving generic activation-level trainability is sufficient to help newborn units acquire useful roles. 
Additional activation controls and hyperparameters are reported in Appendix~\ref{app:rsl_newborn_integration}.

\vspace{-1.0em}
\paragraph{Interventions decouple current performance from future adaptability.}
The birth interventions reveal that improving newborn integration does not necessarily improve the current endpoint.
Across the controlled ConvNet and ResNet experiments, changes in newborn backward signal are only weakly reflected in in-place ACC, indicating that better local integration is not sufficient to determine current-task performance (App.~\ref{app:interventions_transfer}).
The clearest case is \textsc{Rand.\ Smooth-Leaky}: it improves newborn gradient signal while yielding lower current accuracy, yet produces substantially higher future-reachability AUC, particularly when growth extends through the residual hierarchy.
Thus, an intervention can improve the optimization state left by structural growth without improving the solution currently expressed by the network.
This distinction motivates Sec.~\ref{sec:cl_extension}, where we test whether such improved trainability becomes useful when the learning signal repeatedly changes.

\vspace{-1.0em}
\section{Continual Learning Across Shift Regimes}
\label{sec:cl_extension}
\vspace{-1.0em}

If interventions increase a grown network’s future adaptability without improving its current endpoint, the key test is whether the gain persists under repeated changes in the learning signal. 
Our earlier experiments showed that growth is limited not only by which units are added, but by whether newborn units can integrate into a trained representation. 
We use the following benchmarks to compare how the same structural operators behave under continual shift, rather than to benchmark complete continual-learning methods against one another. We study two regimes: (i) sequential class accumulation, with longer intervals between shifts, and (ii) repeated-shift benchmarks that reveal continuing-learning degradation \citep{dohare2024loss}.


\vspace{-1.25em}
\begin{figure*}[ht]
    \centering
    \includegraphics[width=\textwidth]{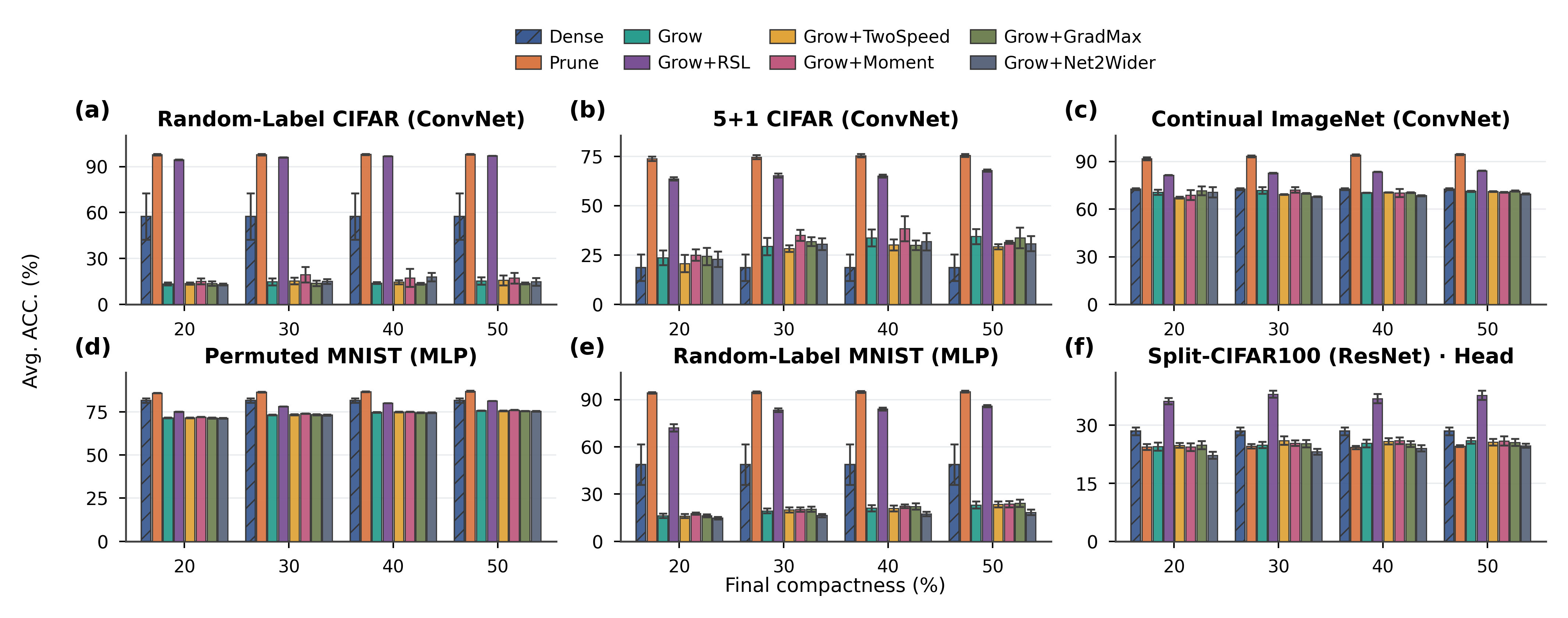}
    \vspace{-2em}
    \caption{
        \textbf{Structural editing across sequential and repeated-shift continual learning.}
        We report average online accuracy across five repeated-shift benchmarks and ResNet Split-CIFAR100. 
        \textsc{Prune} is the most reliable structural
        baseline under rapid shifts, whereas \textsc{Grow + Rand.\ Smooth-Leaky (RSL)} is the most consistent growth variant and strongly improves over vanilla \textsc{Grow} on Split-CIFAR100; activation-matched \textsc{Dense} and \textsc{Prune} controls are reported in App.~\ref{app:activation_control}.
    }
    \label{fig:cl_plasticity_benchmarks}
    \vspace{-15pt}
\end{figure*}

\paragraph{Growth under sequential class accumulation.}
We evaluate Split-CIFAR100 using the ResNet architecture from Sec.~\ref{sec:resnet_locus}. 
Task boundaries are known during training, but tasks share one classifier and task identity is unavailable at test time. 
At each boundary, the model either stays dense or is structurally edited before learning the next subset. 
Since Split-CIFAR100 provides longer intervals between shifts, it tests whether newborn units become useful when given more time to integrate.
Figure~\ref{fig:cl_plasticity_benchmarks} shows that \textsc{Grow}+\textsc{Rand. Smooth-Leaky} substantially improves average online accuracy over vanilla \textsc{Grow} and the other birth interventions.
Although Rand.\ Smooth-Leaky also benefits non-growing models (App.~\ref{app:activation_control}), its gain is largest for \textsc{Grow}, consistent with improved newborn integration when sufficient adaptation time is available.


\vspace{-1.00em}
\paragraph{Growth under repeated non-stationary shifts.}
To distinguish this effect from changes in capacity or interference alone, we also evaluate five benchmarks designed to stress repeated-shift plasticity \citep{kumar2023maintaining}. \textbf{Permuted MNIST} \citep{goodfellow2013empirical} applies a new fixed pixel permutation at each task; \textbf{Random-Label MNIST} and \textbf{Random-Label CIFAR} \citep{lyle2023understanding} repeatedly assign random labels to a fixed data subset; \textbf{CIFAR 5+1} alternates harder five-class and easier single-class tasks drawn from CIFAR-100; and \textbf{Continual ImageNet} \citep{dohare2024loss,imagenet15russakovsky} presents non-repeating binary classification tasks drawn from distinct ImageNet classes. 
Together, these benchmarks span input and concept shift while repeatedly testing whether the current network remains capable of learning the next task.

\vspace{-1.0em}
\begin{wrapfigure}{r}{0.47\textwidth}
    \vspace{-1.2em}
    \centering
    \includegraphics[width=0.46\textwidth]{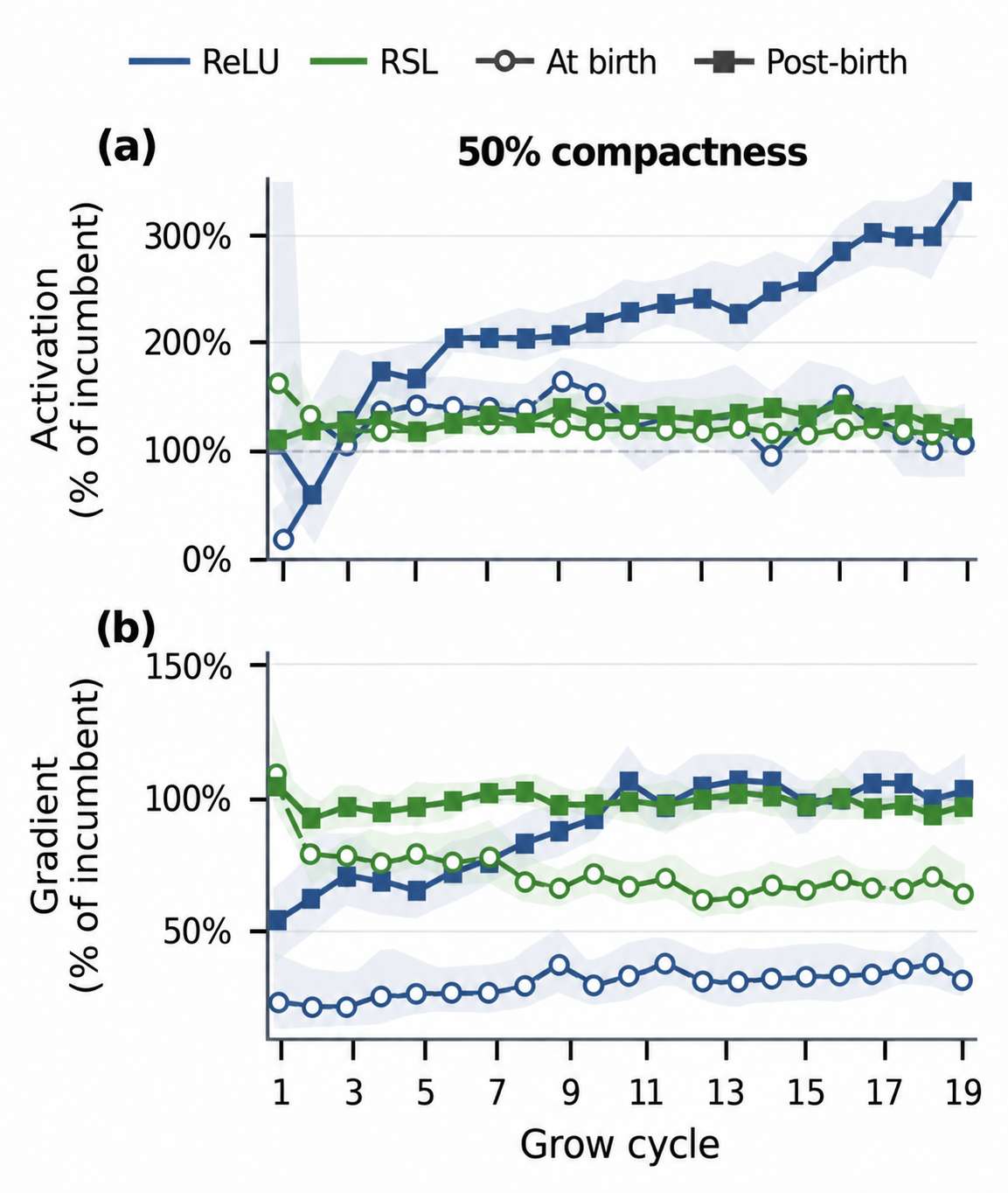}
    \vspace{-1.0em}
    \caption{
    \textbf{Rand.\ Smooth-Leaky stabilizes repeated birth events on Split-CIFAR100.}
    At $50\%$ compactness, we compare newborn vs. incumbent activations and gradient signal relative to incumbent units across 20 growth cycles.
    With ReLU, newborn gradients are low at birth and their activations rise over cycles.
    Rand.\ Smooth-Leaky (RSL) keeps newborn activations near incumbent levels and boosts post-birth gradients toward parity, improving usability under sequential growth.
    }
    \label{fig:split_cifar100_rsl_birth_dynamics}
    \vspace{-1.5em}
\end{wrapfigure}


\paragraph{Growth is competitive but more regime-sensitive under continual shift.}
Across the repeated-shift benchmarks, \textsc{Prune} is generally the strongest or near-strongest structural method, consistent with the advantage of retaining units that are already integrated into the representation and optimization state.
Vanilla \textsc{Grow} can match or exceed \textsc{Dense} in some cases, including CIFAR 5+1, but its gains are less stable.
Among the growth variants, \textsc{Grow}+\textsc{Rand.\ Smooth-Leaky} is the most reliable.
Because this activation also improves \textsc{Dense} and \textsc{Prune} in several benchmarks, App.~\ref{app:activation_control} compares its effect within each structural method; the gains are often largest for \textsc{Grow}.
Figure~\ref{fig:split_cifar100_rsl_birth_dynamics} helps explain this advantage on sequential class accumulation:
on Split-CIFAR100, Rand.\ Smooth-Leaky keeps newborn activations near incumbent levels while substantially improving newborn gradient signal across repeated growth cycles, making new units more trainable under repeated structural edits.
Moment Transplant and GradMax provide occasional gains, while TwoSpeed and Net2Wider are less effective, consistent with Fig.~\ref{fig:birth_interventions}.
In this controlled comparison, growth can match or beat pruning in some regimes, but works best when newborn units stabilize before the next shift.

\vspace{-1.0em}
\section{Conclusion \& Future Work}
\label{sec:conclusion}
\vspace{-1.0em}

\paragraph{Conclusion}
Growth and pruning are complementary forms of structural plasticity but not optimization-symmetric.
Given similar compactness and compute, both can yield competitive sparse architectures; they mainly differ in the trajectories that produce them.
New units are added to an already specialized network and may activate before receiving comparable backward credit, creating a lag between architectural expansion and functional integration.
The consequences of this lag are weak in small MLPs, become visible in convolutional feature learning, and become increasingly consequential when edits enter the residual hierarchy.
Head-localized growth can produce concentrated, task-aligned representations and enable rapid functional remapping, whereas broader residual growth yields narrower feature support and lower bounded future adaptability despite similar current endpoints.
Our interventions further show that improving newborn trainability need not improve immediate accuracy: it can instead change the optimization state left by growth and make future solutions more reachable.
Under continual shift, \textsc{Grow} becomes more competitive when newborn units are sufficiently trainable and have enough time to integrate.
Overall, structural growth should be judged not only by the architecture it produces, but by whether the capacity it adds becomes usable for subsequent optimization.

\vspace{-5pt}
\paragraph{Future Work}
These findings suggest that growth rules should optimize for \emph{realized}, not merely added, capacity.
A natural next step is to replace fixed growth schedules with integration-aware policies that decide when and where to add structure using signals such as newborn-to-incumbent gradient balance, representational support, or short-horizon bounded reachability.
Such policies could delay subsequent edits until newly added capacity has stabilized, or redirect growth toward loci where additional structure remains readily trainable.
A second direction is to test whether these principles persist in larger residual and attention-based architectures and against end-to-end continual-learning and modular expansion methods.
More broadly, structural adaptation should be evaluated jointly by the architecture it produces, its immediate task utility, and the optimization state it leaves behind.


\bibliographystyle{iclr2026_conference}
\bibliography{iclr2026_conference}


\newpage
\appendix

\counterwithin{figure}{section}
\counterwithin{table}{section}


\section{Datasets, Benchmarks and Hyperparameters}
\label{app:repro}

\vspace{-0.5em}
\subsection{Datasets and Benchmarks}
\label{app:datasets_streams}
\vspace{-0.5em}
We evaluate structural adaptation under three data regimes: (i) stationary i.i.d.\ supervised learning, (ii) non-stationary single-head class-incremental streams, and (iii) continual-learning benchmarks designed to stress plasticity under repeated shift. In all cases, we use the standard train/test splits provided with each dataset unless otherwise specified, and the test set is used only for evaluation.

\vspace{-0.5em}
\subsubsection{Independently and identically distributed (i.i.d.)}
\label{app:datasets_iid}

\vspace{-0.5em}
\paragraph{MNIST}
MNIST is treated as a stationary 10-class classification problem. We train on the full training set and evaluate on the standard MNIST test set. Each run uses a fixed budget of $5$ \textsc{Grow} or \textsc{Prune} cycles, with $20$ epochs per cycle. Winning-ticket retraining runs for $100$ epochs to match the total cycle-training budget.

\vspace{-0.5em}
\paragraph{CIFAR-100}
CIFAR-100 is treated as a stationary 100-class classification problem. 
We train on the full training set and evaluate on the standard CIFAR-100 test set. 
Each run uses $200$ epochs total for dense training and winning-ticket retraining. Cycle training is divided into $5$ \textsc{Grow} or \textsc{Prune} cycles, with $40$ epochs per cycle. 
This setting does not use experience replay.


\vspace{-0.5em}
\subsubsection{Class-Incremental Continual Learning}
\label{app:datasets_c_il}

\vspace{-0.5em}
\paragraph{Class-incremental Split-MNIST and Split-FashionMNIST}
MNIST and FashionMNIST each contain 10 classes. 
We construct a stream of $K=5$ tasks by partitioning the classes into five disjoint class pairs. 
We use a single shared 10-way classifier head throughout training, with no task-id routing and no multi-head evaluation. 
At task $t\in\{1,\dots,K\}$, training uses only the two classes assigned to task $t$, while evaluation uses the cumulative test set containing all classes observed up to task $t$. 
Unless otherwise noted, the class-pair ordering is randomized per run using a fixed seed, and the same ordering is shared across methods within each seed.

\vspace{-0.5em}
\paragraph{Experience replay}
To avoid forgetting-dominated collapse in the single-head class-incremental setting, we optionally use experience replay. 
The replay buffer stores up to $M=50$ examples per class, with a maximum total size of $200$ examples. 
During training on tasks $t>1$, each mini-batch mixes current-task samples with replay samples. 
After finishing task $t$, we add up to $M$ examples for each newly observed class. 
Unless explicitly stated as ``no replay,'' Split-MNIST and Split-FashionMNIST use TinyER with a replay fraction of $0.5$ and a fixed budget of $20$ epochs per task ($100$ epochs total across 5 tasks), aligned with the structural-edit cycles.


\vspace{-0.5em}
\paragraph{Split-CIFAR100 (class-incremental)}
We also evaluate class-incremental Split-CIFAR100 as a sequential accumulation benchmark. 
We partition CIFAR-100 into 20 disjoint 5-class tasks and train for 10 epochs per task with batch size 64, without experience replay, and the learner is trained without task-identity information at test time. 
As in the other single-head stream settings, we use a shared 100-way classifier head throughout training and evaluate cumulatively over all classes observed so far where identity is unavailable at test time. 
At each task boundary, the model either remains dense or undergoes a structural edit before continuing optimization on the new task. 
Results for this setting are reported in the main text continual-learning section.

\vspace{-0.5em}
\subsubsection{Plasticity-stressing continual-learning benchmarks}
\label{app:datasets_plasticity}
Following \citet{kumar2023maintaining}, we evaluate five supervised continual image-classification benchmarks spanning two shift types: \emph{input-distribution shift} (Permuted MNIST, 5+1 CIFAR, and Continual ImageNet) and \emph{concept shift} (Random Label MNIST and Random Label CIFAR). 
Across all settings, training proceeds as a sequence of tasks without task-identity signals: the model is never told when a task switch occurs. 
Within each task, the learner receives mini-batches for a fixed duration and is updated incrementally with cross-entropy on the arriving batches. 
Summary hyperparameters are reported in Table~\ref{tab:bench_hp_optimal_app}.

\vspace{-0.5em}
\paragraph{Permuted MNIST}
Permuted MNIST \citep{goodfellow2013empirical} is used as an input-shift benchmark. We first sample a fixed subset of $10{,}000$ images from the MNIST training set. Each task is defined by drawing a new fixed random permutation over pixel indices and applying it to every image in the subset. The permutation remains constant within a task and is independent across tasks. Each task presents exactly one pass over its $10{,}000$ permuted images in mini-batches of size $16$, after which the next task begins with a new permutation. We train for $500$ tasks in total.

\vspace{-0.5em}
\paragraph{Random Label MNIST}
Random Label MNIST \citep{lyle2023understanding} is used as a concept-shift benchmark. We fix a subset of $1{,}200$ MNIST images once, and for each task generate a fresh random label for every image in the subset. The inputs are unchanged across tasks, but the input-label mapping changes completely. To encourage memorization under an arbitrary target function, the model is trained for $400$ epochs per task with batch size $16$. After each task, a new independent random labeling is sampled. We run $50$ tasks in sequence.

\vspace{-0.5em}
\paragraph{Random Label CIFAR}
Random Label CIFAR follows the same protocol as Random Label MNIST, but uses images drawn from CIFAR-10. We again fix a subset of $1{,}200$ images, reassign random labels independently for each task, and train for $400$ epochs per task with batch size $16$ over $50$ tasks.

\vspace{-0.5em}
\paragraph{5+1 CIFAR}
5+1 CIFAR is an input-shift benchmark with alternating task difficulty. Tasks are constructed from CIFAR-100 and alternate between \emph{hard} tasks containing $5$ classes ($2{,}500$ images total, $500$ per class) and \emph{easy} tasks containing a single class ($500$ images). Classes do not repeat across the sequence. Each task lasts $780$ parameter-update steps. With batch size $32$, this corresponds to approximately $10$ epochs on hard tasks and approximately $50$ epochs on easy tasks. We report performance on the hard tasks only, since single-class tasks are near ceiling for all methods.

\vspace{-0.5em}
\paragraph{Continual ImageNet}
Continual ImageNet \citep{dohare2024loss,imagenet15russakovsky} is used as an input-shift benchmark. Each task is a binary classification problem between two distinct ImageNet classes. For every task, we draw $1{,}200$ images total ($600$ per class) and downsample them to $32{\times}32$, following \citet{dohare2024loss}, to reduce compute while preserving semantic variability. Classes do not repeat across tasks. We train for $10$ epochs per task with batch size $100$ and report task accuracy.

\vspace{-0.5em}
\begin{table}[ht]
\centering
\small
\begin{tabular}{lccccl}
\toprule
\textbf{Benchmark} & \textbf{\shortstack{Per-Task\\Data Size}} & \textbf{Batch} & \textbf{Epochs} & \textbf{Timesteps} & \textbf{\# Tasks} \\
\midrule
Permuted MNIST & $10{,}000$ images & $16$ & $1$ & $625$ & $500$ \\
Random Label MNIST & $1{,}200$ images& $16$ & $400$ & $30{,}000$ & $50$ \\
Random Label CIFAR & $1{,}200$ images & $16$ & $400$ & $30{,}000$ & $50$ \\
\multirow{2}{*}{5+1 CIFAR} & \begin{tabular}[t]{@{}l@{}}Hard: $2{,}500$ images \\ (5 classes, 500/class)\end{tabular} & \multirow{2}{*}{$32$} & \begin{tabular}[t]{@{}l@{}}Hard: $\approx 10$\end{tabular} & \multirow{2}{*}{$780$} & \multirow{2}{*}{15} \\
& \begin{tabular}[t]{@{}l@{}}Easy: $500$ images \\ (1 class)\end{tabular} & & \begin{tabular}[t]{@{}l@{}}Easy: $\approx 50$\end{tabular} & & 15 \\
Continual ImageNet & $1{,}200$ images/task (600/class) & $100$ & $10$ & $120$ & 500 \\
\bottomrule
\end{tabular}

\footnotesize
\caption{Hyperparameters and schedule per benchmark. \emph{Timesteps} denote parameter-update steps (i.e., mini-batches) within a task. For 5+1 CIFAR, a fixed timestep budget per task implies approximate epochs depending on data size.}
\textbf{Notes.} (i) In 5+1 CIFAR, classes do not repeat across tasks; tasks alternate easy/hard. $780$ timesteps $\approx 10$ epochs on the hard set (since $2{,}500/32 \approx 78.125$ batches/epoch) and $\approx 50$ epochs on the easy set (since $500/32 \approx 15.625$). (ii) In Continual ImageNet, images are downsampled to $32{\times}32$ to reduce compute; classes do not repeat across tasks. (iii) Timesteps are computed as the number of mini-batches per task.
\label{tab:bench_hp_optimal_app}
\end{table}

\vspace{-1.5em}
\subsection{Hyperparameters, Optimizers and Learning Rates}
\label{app:opt_schedules}
\vspace{-0.5em}
For each MLP, ConvNet, and ResNet dataset--architecture setting, we sweep optimizers in $\{\textsc{SGD},\textsc{Adam},\textsc{AdamW}\}$ and learning rates $\eta\in\{0.1,0.01,0.001,0.0001\}$ independently for each treatment.
For each method, we select the configuration with the highest mean final accuracy.
Thus, all methods receive the same hyperparameter search space and tuning budget, rather than being evaluated under a configuration favored by another treatment.
Within each selected configuration, the remaining optimizer hyperparameters, training budget, and schedule are fixed; they differ only through their prescribed structural-edit procedure and resulting active units (App.~\ref{app:nn_archs}). 
Unless stated otherwise, all models use ReLU activations, and all \texttt{Conv2d} and \texttt{Linear} layers use Kaiming uniform initialization, except ResNet \texttt{Linear} layers which use Xavier uniform initialization.

The continual-learning benchmarks in Sec.~\ref{sec:cl_extension} use a separate tuning protocol: \textsc{Dense} uses the dataset-specific reference learning rate, while non-dense methods select the best learning rate by average accuracy from $\{0.01,0.001,0.0001\}$.
For the activation-control analysis in App.~\ref{app:activation_control}, learning rates are selected independently for ReLU and Rand.\ Smooth-Leaky before computing activation-specific differences.

\vspace{-0.5em}
\paragraph{Learning-rate schedule}
For all experiments in Sections \ref{sec:results} and \ref{sec:integration_interventions} we apply cosine annealing over the full training horizon with $T_{\max}=N_{\text{epochs}}$ and $\eta_{\min}=0$, so the learning rate decays smoothly from its initial value to zero over $200$ epochs.

\subsection{Performance Metrics: ACC, trajectory-average accuracy (TAA), and TAOA}
\label{app:metrics}

We use the same metric names and evaluation conventions as in the main text. 
Our primary evaluation index is defined by \textsc{Grow}/\textsc{Prune} checkpoints: in class-incremental streams these checkpoints coincide with task boundaries, while in i.i.d.\ settings they simply mark successive structural-edit cycles.

\vspace{-0.5em}
\paragraph{Checkpoint indexing}
Let $t\in\{1,\dots,T\}$ index evaluation checkpoints along training.
In our implementation, checkpoints occur at the end of each \textsc{Grow}/\textsc{Prune} cycle. 
In class-incremental streams, this aligns $t$ with task time; in i.i.d.\ settings, $t$ indexes successive structural-edit events.

\vspace{-0.5em}
\paragraph{Class-incremental settings}
When tasks exist, let $K_t$ be the number of tasks encountered up to checkpoint $t$, and let $\mathrm{Acc}_{t,k}$ denote test accuracy on task $k\in\{1,\dots,K_t\}$ measured at checkpoint $t$. 
We define cumulative accuracy at checkpoint $t$ as
\begin{equation}
\mathrm{Acc}^{\mathrm{cum}}_t
\;=\;
\frac{1}{K_t}\sum_{k=1}^{K_t}\mathrm{Acc}_{t,k}.
\label{eq:acc_cum}
\end{equation}
We then report
\begin{align}
\mathrm{ACC} \;&=\; \mathrm{Acc}^{\mathrm{cum}}_{T}, \label{eq:acc_final_stream}\\
\mathrm{TAA} \;&=\; \frac{1}{T}\sum_{t=1}^{T}\mathrm{Acc}^{\mathrm{cum}}_t. \label{eq:acc_auc_stream}
\end{align}
Thus, $\mathrm{ACC}$ summarizes end-of-stream performance, while $\mathrm{TAA}$ summarizes performance over the full trajectory.

\vspace{-0.5em}
\paragraph{I.i.d.\ settings}
When there are no tasks, let $\mathrm{Acc}_t$ be the test accuracy at checkpoint $t$. 
We report
\begin{align}
\mathrm{ACC} \;&=\; \mathrm{Acc}_{T}, \label{eq:acc_final_iid}\\
\mathrm{TAA} \;&=\; \frac{1}{T}\sum_{t=1}^{T}\mathrm{Acc}_t. \label{eq:acc_auc_iid}
\end{align}
In this case, $\mathrm{ACC}$ is the standard final test accuracy and $\mathrm{TAA}$ is the average test accuracy over the checkpoint-defined training trajectory.

\vspace{-0.5em}
\paragraph{In-place vs.\ Winning-Ticket evaluation}
Both \emph{In-place} and \emph{Winning-Ticket} results report $\mathrm{ACC}$ and $\mathrm{TAA}$ using the definitions above. 
In-place metrics are measured during the \textsc{Grow}/\textsc{Prune} procedure while the mask changes across checkpoints. 
Winning-Ticket metrics are measured after freezing the discovered mask, reinitializing the model, and retraining it from scratch under the same data stream and replay setting, following standard Lottery Ticket evaluation \citep{frankle2019lottery}.

\vspace{-0.5em}
\paragraph{Total Average Online Accuracy (TAOA)}
On the continual-learning benchmarks, we additionally use Total Average Online Accuracy (TAOA), following prior online continual-learning work \citep{cai2021online, ghunaim2023real, prabhu2023online, kumar2023maintaining}. 
Unlike $\mathrm{ACC}$ and $\mathrm{TAA}$, which are checkpoint-based, TAOA aggregates online accuracy over all mini-batches seen so far. 
Let $B_{\le T}=\sum_{i=1}^{T} M_i$ be the total number of processed mini-batches up to task $T$, and let $a_t$ denote online accuracy at global batch index $t$. 
We define
\begin{equation}
\mathrm{TAOA}_{\le T}
=
\frac{1}{B_{\le T}}\sum_{t=0}^{B_{\le T}-1} a_t.
\end{equation}
If all tasks have equal length $M_i\equiv M$, this reduces to
\begin{equation}
\mathrm{TAOA}_{\le T}
=
\frac{1}{MT}\sum_{t=0}^{MT-1} a_t.
\end{equation}
We use TAOA in Section \ref{sec:cl_extension} to capture how quickly the agent learns the current task (plasticity) and distinguish it from the checkpoint-based $\mathrm{ACC}$ and $\mathrm{TAA}$ reported in the rest of the main paper.       
\vspace{-1.0em}
\subsection{Neural Network Architectures}
\label{app:nn_archs}
\vspace{-0.5em}

All methods (\textsc{Dense}, \textsc{Grow}, \textsc{Prune}) share the same underlying parameterization and differ only in the binary unit masks that determine which hidden units are active during forward and backward passes. 
This ensures that comparisons isolate the effect of structural adaptation rather than changes in the base architecture.

\vspace{-0.5em}
\paragraph{Unit-wise masking}
For a hidden layer $\ell$ with pre-activation
\begin{equation}
z^{(\ell)} = W^{(\ell)} h^{(\ell-1)} + b^{(\ell)},
\end{equation}
and activation function $\phi(\cdot)$, we define
\begin{equation}
h_{\mathrm{raw}}^{(\ell)} = \phi\!\left(z^{(\ell)}\right),
\qquad
h^{(\ell)} = h_{\mathrm{raw}}^{(\ell)} \odot m^{(\ell)},
\end{equation}
where $m^{(\ell)} \in \{0,1\}^{d_\ell}$ is a unit-wise binary mask and $\odot$ denotes element-wise multiplication. Mask entries set to zero fully deactivate the corresponding hidden units, affecting both activations and gradients. Masks are stored as non-trainable buffers and are therefore serialized with the model state.

\vspace{-0.5em}
\subsubsection{MLP for MNIST and FashionMNIST}
\label{app:mlp_arch}

Our MLP backbone is a 3-layer fully connected network:
\begin{align}
\texttt{fc1}:&\;\; 784 \rightarrow H, \\
\texttt{fc2}:&\;\; H \rightarrow H, \\
\texttt{fc3}:&\;\; H \rightarrow 10,
\end{align}
with two masked hidden layers (\texttt{fc1}, \texttt{fc2}) and an unmasked output layer (\texttt{fc3}). Unless otherwise noted, $H=256$. 

Dense training corresponds to $m^{(1)}=\mathbf{1}$ and $m^{(2)}=\mathbf{1}$ throughout. 
Growth begins from a small active fraction (default $10\%$) and expands masks over time by flipping selected zeros to ones. 
Pruning starts fully active and removes units by setting mask entries to zero.

\vspace{-0.5em}
\subsubsection{ConvNet for CIFAR-10 \& CIFAR-100}
\label{app:cifar_arch}

For both CIFAR-10 and CIFAR-100 we use a fully active convolutional trunk and a growable/prunable MLP head.

\vspace{-0.5em}
\paragraph{Convolutional trunk}
The trunk is
\begin{align}
\texttt{conv1}:&\;\; 3 \rightarrow 32,\; k=3,\; p=1 \;\rightarrow\; \phi \;\rightarrow\; \texttt{maxpool}(2),\\
\texttt{conv2}:&\;\; 32 \rightarrow 64,\; k=3,\; p=1 \;\rightarrow\; \phi \;\rightarrow\; \texttt{maxpool}(2),
\end{align}
followed by flattening to $8 \times 8 \times 64 = 4096$ features. 
No structural adaptation is applied in the convolutional trunk.

\vspace{-0.5em}
\paragraph{Masked fully connected head}
The head has four fully connected layers:
\begin{align}
\texttt{fc1}:&\;\; 4096 \rightarrow H_1 \;\rightarrow\; \phi \;\rightarrow\; \odot\, m^{(1)},\\
\texttt{fc2}:&\;\; H_1 \rightarrow H_2 \;\rightarrow\; \phi \;\rightarrow\; \odot\, m^{(2)},\\
\texttt{fc3}:&\;\; H_2 \rightarrow H_3 \;\rightarrow\; \phi \;\rightarrow\; \odot\, m^{(3)},\\
\texttt{fc4}:&\;\; H_3 \rightarrow C,
\end{align}
where $C$ is the number of output classes. 
By default, $(H_1,H_2,H_3)=(512,512,256)$. 
In all cases, structural adaptation is confined to the fully connected head.

\vspace{-0.5em}
\subsubsection{ResNet for CIFAR-100}
\label{app:resnet_arch}

For the CIFAR-100 feature-hierarchy experiments, we use a CIFAR-style ResNet with a $3\times3$ convolutional stem followed by three residual stages. 
The stage widths are $(16,32,64)$, with three basic blocks per stage. 
The first block of \texttt{stage2} and \texttt{stage3} uses stride two; all remaining convolutions use stride one. 
Projection shortcuts are used whenever the spatial resolution or channel dimension changes.

The resulting feature hierarchy is
\begin{align}
\texttt{stem}:&
\quad 3\times32\times32
\rightarrow 16\times32\times32,\\
\texttt{stage1}:&
\quad 16\times32\times32
\rightarrow 16\times32\times32,\\
\texttt{stage2}:&
\quad 16\times32\times32
\rightarrow 32\times16\times16,\\
\texttt{stage3}:&
\quad 32\times16\times16
\rightarrow 64\times8\times8.
\end{align}
After \texttt{stage3}, global average pooling produces a $64$-dimensional vector. 
The classifier head is
\begin{align}
    \texttt{fc1}:&\quad 64 \rightarrow 256 \rightarrow \phi,\\
    \texttt{fc2}:&\quad 256 \rightarrow C,
\end{align}
where $C=100$ for CIFAR-100.

\vspace{-0.5em}
\paragraph{Masked residual units.}
Each basic block contains two $3\times3$ convolutions. 
Writing $S_b(\cdot)$ for the identity or projection shortcut, the block computes
\begin{align}
u_b
&=
\phi\!\left(
\operatorname{BN}_{b,1}(W_{b,1}h_b)
\right),\\
\widetilde u_b
&=
m_b\odot u_b,\\
h_{b+1}
&=
\phi\!\left(
\operatorname{BN}_{b,2}(W_{b,2}\widetilde u_b)
+
S_b(h_b)
\right),
\end{align}
where $m_b\in\{0,1\}^{d_b}$ is broadcast across the spatial dimensions. 
The mask is applied after the first batch-normalization and activation and before the second convolution.

A masked residual unit consists of one output channel of $W_{b,1}$, its associated batch-normalization channel, and the corresponding input-channel slice of $W_{b,2}$. 
Setting its mask entry to zero removes that channel from the residual computation and blocks its forward contribution and gradient signal. 
The skip pathway remains active, so masking changes the learned residual transformation without removing the block's identity connection. 
The stem, second-convolution output channels, shortcut pathways, and final output layer are never masked.

\vspace{-0.5em}
\paragraph{Structural edit loci.}
We control where structural changes occur through four mutually exclusive edit-locus configurations:
\begin{description}[leftmargin=1.5em,itemsep=0.15em]
    \item[\textsc{Head}:] only the $256$ hidden units of \texttt{fc1} are masked; all residual stages remain fully active.

    \item[\textsc{Late}:] the hidden residual channels in every block of \texttt{stage3} are masked; the classifier head and earlier stages remain fully active.

    \item[\textsc{Mid+Late}:] the hidden residual channels in every block of \texttt{stage2} and  \texttt{stage3} are masked; \texttt{stage1} and the classifier head remain fully active.

    \item[\textsc{All}:] the hidden residual channels in every block of \texttt{stage1}, \texttt{stage2}, and \texttt{stage3} are masked; the classifier head remains fully active.
\end{description}
Thus, the loci are not cumulative with respect to \textsc{Head}: feature-level edit loci modify residual channels but do not additionally mask \texttt{fc1}.

\vspace{-0.5em}
\paragraph{Diagnostic feature readouts.}
The model records representations after the stem and after the final block of each residual stage:
\begin{align}
\texttt{stem\_out}&\in\mathbb{R}^{16\times32\times32},\\
\texttt{stage1\_out}&\in\mathbb{R}^{16\times32\times32},\\
\texttt{stage2\_out}&\in\mathbb{R}^{32\times16\times16},\\
\texttt{stage3\_out}&\in\mathbb{R}^{64\times8\times8}.
\end{align}

\vspace{-0.5em}
\subsection{Compactness and Sparsity}
\label{app:budgets_compactness}
\vspace{-0.5em}

Our structural interventions operate through unit-wise binary masks applied to selected hidden layers. 
Therefore, we define compactness budgets in terms of active units and match these budgets across \textsc{Dense}, \textsc{Grow}, and \textsc{Prune}.

\vspace{-0.5em}
\paragraph{Layer compactness}
For a masked layer $\ell$ with $d_\ell$ units and mask $m^{(\ell)} \in \{0,1\}^{d_\ell}$, the number of active units is
\begin{equation}
    a_\ell = \|m^{(\ell)}\|_0 = \sum_{j=1}^{d_\ell} m^{(\ell)}_j,
\end{equation}
and the layer compactness is
\begin{equation}
    c_\ell = \frac{a_\ell}{d_\ell} \in (0,1].
\end{equation}

\vspace{-0.5em}
\paragraph{Global weight-space compactness for MLPs and ConvNets.}
For the MLP and ConvNet experiments, the final compactness target $c\in(0,1]$ specifies the fraction of structured weights retained or activated across all masked layers. 
We allocate this global weight budget across layers and convert the resulting allocations into integer unit targets.
Consequently, $c$ is primarily a weight-space budget and need not produce the same active-unit fraction in every masked layer.

\vspace{-0.5em}
\paragraph{Per-mask channel compactness for ResNets.}
The ResNet edit-locus experiment instead uses a per-mask channel target.
Let $\mathcal{G}_{\mathcal{L}}$ denote the collection of masks editable under locus $\mathcal{L}$. 
For each mask $g\in\mathcal{G}_{\mathcal{L}}$ with $d_g$ structural units, the target compactness $c$ determines
\begin{equation}
a_g^\star = \left\lfloor c\,d_g \right\rceil,
\qquad
c_g = \frac{\|m_g\|_0}{d_g} \approx c,
\end{equation}
up to integer rounding. The same target fraction is therefore applied independently to every editable residual block, or to \texttt{fc1} in the \textsc{Head} condition.

For example, at $c=0.5$:
\begin{itemize}[leftmargin=1.5em,itemsep=0.1em]
    \item \textsc{Head} retains approximately $128$ of the $256$ \texttt{fc1} units;
    \item \textsc{Late} retains approximately $32$ of $64$ hidden channels in each \texttt{stage3} block;
    \item \textsc{Mid+Late} retains approximately $16$ of $32$ channels in each \texttt{stage2} block and $32$ of $64$ in each \texttt{stage3} block;
    \item \textsc{All} additionally retains approximately $8$ of $16$ channels in each \texttt{stage1} block.
\end{itemize}

The corresponding mask sparsity is $s_g=1-c_g$. 
Because each locus exposes a different number and width of masks, the same reported compactness does \emph{not} imply the same whole-network parameter sparsity across \textsc{Head}, \textsc{Late}, \textsc{Mid+Late}, and \textsc{All}. 
Instead, it matches the fraction of active structural units within every editable block. 
This avoids confounding edit depth with a global allocator that could redistribute the budget toward particular stages.

\vspace{-0.5em}
\paragraph{Matched structural budgets.}
Within a dataset--architecture setting and edit locus, \textsc{Grow} and \textsc{Prune} are budget-matched when they reach the same integer targets $\{a_g^\star\}_{g\in\mathcal{G}_{\mathcal{L}}}$. Growth approaches these targets from a sparse initial mask, whereas pruning approaches them from a fully active mask. 
Dense keeps all units active and is locus-agnostic.

Masking induces \emph{effective structured sparsity}: inactive producer filters and consumer slices remain allocated in the parameter tensors but do not participate in the forward or backward computation. 
Reported compactness therefore describes active computational structure rather than compressed model-storage size.         
\vspace{-1.0em}
\subsection{Structural Adaptation Procedure}
\label{app:procedures}
\vspace{-0.5em}

Both \textsc{Grow} and \textsc{Prune} operate on the same masked backbone (App.~\ref{app:nn_archs}) and train for $T$ cycles.
Structural edits occur during cycles $t=1,\ldots,T-1$, reaching the target compactness before the final cycle; cycle $T$ therefore trains the fixed target architecture without an additional edit.
Each cycle consists of 
(i) selecting units to activate or deactivate, 
(ii) updating the masks and any associated bookkeeping, and 
(iii) training for a fixed budget before the next cycle.

\vspace{-0.5em}
\subsubsection{Grow}
\label{app:grow}

\vspace{-0.5em}
\paragraph{Selection rule}
At cycle $t$, for each masked layer $\ell$, we compute a score $s_j^{(\ell)}$ for each currently inactive unit $j$ with $m_j^{(\ell)}=0$. 
Then, we activate the top-$n_{t,\ell}$ inactive units in that layer.
Let $a_\ell=\|m^{(\ell)}\|_0$ be the current active count and $a^\star_\ell$ the target. 

For $t\in\{1,\ldots,T-1\}$, the remaining
number of units to add is
\[ 
q^{\mathrm{grow}}_{t,\ell} = \max(0,a^\star_\ell-a_\ell). 
\]
and we activate
\[
n_{t,\ell}
=
\min\!\left(
q^{\mathrm{grow}}_{t,\ell},
\left\lceil
\frac{q^{\mathrm{grow}}_{t,\ell}}{T-t}
\right\rceil,
d_\ell-a_\ell
\right)
\]
highest-scoring inactive units.

For the MLP and ConvNet experiments, the layer targets are obtained from the global structured-weight allocator; for the ResNet, the same active-channel fraction is applied independently to every mask in the selected edit locus (App.~\ref{app:budgets_compactness}).

\vspace{-0.5em}
\paragraph{Activation-frequency heuristic}
Our default growth heuristic scores an inactive unit by the fraction of post-activation values exceeding a threshold $\tau$ (i.e. $\tau =0.05$). 
Given a mini-batch $B$, we define
\begin{equation}
s^{\mathrm{Act}}_j \;=\;
\begin{cases}
\frac{1}{|B|}\sum\limits_{x\in B}\mathbf{1}\!\left\{A_j(x) > \tau\right\},
& \text{fully connected}, \\[8pt]
\frac{1}{|B|}\sum\limits_{x\in B}\left(\frac{1}{HW}\sum\limits_{u\in[H]\times[W]}
\mathbf{1}\!\left\{A_j(x,u) > \tau\right\}\right),
& \text{convolutional}.
\end{cases}
\label{eq:grow_cycle_quota}
\end{equation}
Here $A_j$ denotes the \emph{raw} post-activation value computed before the binary mask is applied. 
This permits dormant units to be evaluated even though their masked contribution to the network output is zero. 
For convolutional units, the score averages over both examples and spatial locations. 
Intuitively, the score, $s_j^{\mathrm{Act}}$, estimates how often an inactive unit would produce a non-negligible response if activated.

\vspace{-0.5em}
\subsubsection{Prune}
\label{app:prune}

\vspace{-0.5em}
\paragraph{Selection rule.}
\textsc{Prune} starts from a fully active network and removes units over $T$ cycles until reaching the target per-layer unit counts. 

For $t\in\{1,\ldots,T-1\}$, the remaining number to remove is
\[
q_{t,\ell}=a_\ell-a^\star_\ell ,
\]
and we prune
\[
k_{t,\ell}
=
\min\!\left(
q_{t,\ell},
\left\lceil
\frac{q_{t,\ell}}{T-t}
\right\rceil,
a_\ell
\right)
\]
units from the active set by selecting the lowest-scoring units.

\vspace{-0.5em}
\paragraph{Magnitude score.}
Our default pruning score is the mean absolute weight magnitude per unit. 
For a \texttt{Linear} layer,
\begin{equation}
s^{\mathrm{Mag}}_j = \frac{1}{d_{\mathrm{in}}}\sum_{i=1}^{d_{\mathrm{in}}}|W_{j,i}|,
\end{equation}
and for a \texttt{Conv2d} layer,
\begin{equation}
s^{\mathrm{Mag}}_j = \frac{1}{C_{\mathrm{in}}k_Hk_W}\sum_{c,u,v}|W_{j,c,u,v}|.
\end{equation}
We prune the smallest-magnitude units among the currently active set.

\vspace{-0.5em}
\paragraph{IMP rewind.}
Our default pruning procedure uses IMP-style rewinding. 
After updating the mask, we rewind surviving parameter slices to their initialization (or stored rewind snapshot) before retraining, using the current mask to select the surviving rows, filters, and input columns as needed. 
This isolates the effect of subnetwork selection from continued fine-tuning dynamics.

\vspace{-1.0em}
\section{Additional Experimental Studies and Ablations}
\label{sec:ablations_appendix}

\vspace{-0.5em}
\subsection{Compactness-resolved MLP winning-ticket results}
\label{sec:app_mlp_results}
\vspace{-0.5em}

Table~\ref{tab:app_ticket_delta_all_datasets} expands the MLP results from Sec.~\ref{sec:results} across compactness levels.
On IID MNIST, Split-MNIST+ER, and Split-Fashion+ER, \textsc{Grow} and \textsc{Prune} achieve closely matched winning-ticket ACC and TAA with no stable ordering.
The $\Delta_{\mathrm{WT-C}}$ values vary by dataset, but overall these results confirm that small MLP settings do not expose a robust \textsc{Grow}--\textsc{Prune} asymmetry.

\vspace{-0.5em}
\begin{table*}[ht]
\centering
\scriptsize
\renewcommand{\arraystretch}{1.05}
\setlength{\tabcolsep}{2.8pt}
\resizebox{\linewidth}{!}{%
\begin{tabular}{l *{4}{r r r}}
\toprule
\multirow{3}{*}{\textbf{Method}} \\
\cmidrule(lr){2-13}
& \multicolumn{3}{c}{\textbf{20\%}}& \multicolumn{3}{c}{\textbf{30\%}}& \multicolumn{3}{c}{\textbf{40\%}}& \multicolumn{3}{c}{\textbf{50\%}} \\
\cmidrule(lr){2-4}\cmidrule(lr){5-7}\cmidrule(lr){8-10}\cmidrule(lr){11-13}
& {\textbf{WT Final}} & {\textbf{WT TAA}} & {\textbf{$\Delta_{\mathrm{WT-C}}$}}
& {\textbf{WT Final}} & {\textbf{WT TAA}} & {\textbf{$\Delta_{\mathrm{WT-C}}$}}
& {\textbf{WT Final}} & {\textbf{WT TAA}} & {\textbf{$\Delta_{\mathrm{WT-C}}$}}
& {\textbf{WT Final}} & {\textbf{WT TAA}} & {\textbf{$\Delta_{\mathrm{WT-C}}$}} \\
\midrule
\multicolumn{13}{l}{\textbf{MNIST (IID)} \quad \textbf{\textit{Dense (100\%)}}: \textbf{WT Final} \num{95.98 +- 0.09}, \textbf{WT TAA} \num{93.13 +- 0.06}} \\
\cmidrule(lr){1-13}
\textbf{Grow} & \num{95.39 +- 0.13} & \num{92.94 +- 0.13} & \num{-1.16 +- 0.17} & \num{95.96 +- 0.07} & \num{93.71 +- 0.07} & \num{-0.57 +- 0.10} & \num{96.19 +- 0.08} & \num{94.12 +- 0.07} & \num{-0.38 +- 0.11} & \num{96.40 +- 0.07} & \num{94.40 +- 0.06} & \num{-0.22 +- 0.22} \\
\textbf{Prune} & \num{95.33 +- 0.12} & \num{92.87 +- 0.12} & \num{-0.63 +- 0.20} & \num{95.87 +- 0.14} & \num{93.66 +- 0.11} & \num{-0.58 +- 0.15} & \num{96.16 +- 0.13} & \num{94.05 +- 0.10} & \num{-0.62 +- 0.14} & \num{96.39 +- 0.09} & \num{94.39 +- 0.08} & \num{-0.60 +- 0.10} \\
\midrule
\multicolumn{13}{l}{\textbf{Split-MNIST (+ER)} \quad \textbf{\textit{Dense (100\%)}}: \textbf{WT Final} \num{84.36 +- 0.75}, \textbf{WT TAA} \num{90.44 +- 0.77}} \\
\cmidrule(lr){1-13}
\textbf{Grow} & \num{84.53 +- 0.77} & \num{90.19 +- 0.62} & \num{+11.15 +- 1.66} & \num{84.47 +- 0.77} & \num{91.00 +- 0.52} & \num{+11.93 +- 1.16} & \num{84.88 +- 1.09} & \num{91.30 +- 0.66} & \num{+12.21 +- 1.54} & \num{85.81 +- 0.49} & \num{91.41 +- 0.75} & \num{+10.85 +- 1.48} \\
\textbf{Prune} & \num{84.53 +- 0.72} & \num{90.33 +- 0.48} & \num{+8.52 +- 2.14} & \num{85.14 +- 1.17} & \num{90.70 +- 0.84} & \num{+7.92 +- 1.61} & \num{85.69 +- 0.71} & \num{90.98 +- 0.51} & \num{+8.47 +- 1.51} & \num{85.20 +- 0.79} & \num{90.91 +- 0.36} & \num{+7.29 +- 0.98} \\
\midrule
\multicolumn{13}{l}{\textbf{Split-Fashion (+ER)} \quad \textbf{\textit{Dense (100\%)}}: \textbf{WT Final} \num{75.01 +- 1.58}, \textbf{WT TAA} \num{84.34 +- 1.53}} \\
\cmidrule(lr){1-13}
\textbf{Grow} & \num{74.88 +- 1.07} & \num{83.34 +- 2.09} & \num{+8.13 +- 1.07} & \num{74.96 +- 1.66} & \num{85.84 +- 1.34} & \num{+7.21 +- 0.92} & \num{75.70 +- 1.11} & \num{86.02 +- 1.11} & \num{+8.49 +- 1.02} & \num{75.11 +- 2.03} & \num{85.02 +- 2.16} & \num{+7.19 +- 2.12} \\
\textbf{Prune} & \num{75.53 +- 1.57} & \num{86.32 +- 0.76} & \num{+5.01 +- 0.92} & \num{75.70 +- 1.52} & \num{85.72 +- 1.67} & \num{+5.74 +- 2.15} & \num{75.76 +- 1.08} & \num{84.08 +- 2.21} & \num{+4.92 +- 0.97} & \num{75.00 +- 1.48} & \num{84.92 +- 3.04} & \num{+5.71 +- 1.54} \\
\bottomrule
\end{tabular}%
}
\vspace{-0.5em}
\caption{
    \textbf{Compactness-resolved winning-ticket performance for the small MLP benchmarks.}
    Entries are mean $\pm$ 95\% CI over $10$ seeds.
    For each compactness budget, we report Winning-Ticket final accuracy (\textbf{WT Final}), Winning-Ticket trajectory-average accuracy (\textbf{WT TAA}), and 
    $\Delta_{\mathrm{WT-C}} = \mathrm{Final}_{\mathrm{WT}} - \mathrm{Final}_{\mathrm{C}}$.
    Negative $\Delta_{\mathrm{WT-C}}$ values indicate that the in-place structural-edit \emph{cycle} reached a higher endpoint than the retrained ticket, while positive values indicate that the frozen mask retrained from scratch to a higher endpoint than the in-place structural-edit \emph{cycle} procedure achieved.
    Dense is invariant to compactness and is therefore reported once per dataset header.
    }
\label{tab:app_ticket_delta_all_datasets}
\vspace{-2.0em}
\end{table*}             
\subsection{CIFAR-100 ConvNet: Additional Results and Ablations}
\label{app:cifar100_convnet}
\vspace{-0.5em}

Table~\ref{tab:cifar100_cycle_ticket_deltas} makes the CIFAR-100 in-place--vs.--ticket dissociation explicit across compactness.
\textsc{Grow} consistently attains higher in-place final ACC than \textsc{Prune}, yet their Winning-Ticket performance is statistically indistinguishable after retraining from scratch.
As compactness increases, \textsc{Grow} shows increasingly negative $\Delta\mathrm{ACC}$, indicating stronger dependence of its cycle endpoint on the adaptive path, whereas \textsc{Prune} remains near-zero or positive.
Thus, the main separation lies in the structural-editing process rather than final retrainable architecture quality.

\vspace{-0.5em}
\begin{table}[ht]
\centering
\small
\setlength{\tabcolsep}{3.5pt}
\resizebox{\linewidth}{!}{%
\begin{tabular}{r l r r r r r r}
\toprule
& & \multicolumn{2}{c}{\textbf{Cycle Eval.}} & \multicolumn{2}{c}{\textbf{Winning-ticket Eval.}} & \multicolumn{2}{c}{\textbf{Ticket $-$ Cycle}} \\
\cmidrule(lr){3-4}\cmidrule(lr){5-6}\cmidrule(lr){7-8}
\textbf{Comp. (\%)} & \textbf{Method} & \textbf{ACC} & \textbf{TAA} & \textbf{ACC} & \textbf{TAA} & $\boldsymbol{\Delta\mathrm{ACC}}$ & $\boldsymbol{\Delta\mathrm{TAA}}$ \\
\midrule
100 & \textbf{\textsc{Dense}} & \textbf{49.660$\pm$0.315} & \textbf{48.062$\pm$0.284} & \textbf{49.660$\pm$0.315} & \textbf{45.998$\pm$0.262} & 0.000 & $-2.065$ \\
\midrule
20 & \textsc{Grow} & \textbf{52.316$\pm$0.317} & 47.288$\pm$0.269 & 52.055$\pm$0.324 & 46.257$\pm$0.188 & $-0.261$ & \textbf{$-1.031$} \\
20 & \textsc{Prune} & 47.911$\pm$0.644 & \textbf{48.519$\pm$0.348} & \textbf{52.057$\pm$0.283} & \textbf{46.298$\pm$0.162} & \textbf{$+4.146$} & $-2.221$ \\
\addlinespace
30 & \textsc{Grow} & \textbf{53.312$\pm$0.247} & 47.871$\pm$0.374 & \textbf{51.626$\pm$0.232} & \textbf{46.909$\pm$0.220} & $-1.686$ & \textbf{$-0.963$} \\
30 & \textsc{Prune} & 49.549$\pm$0.325 & \textbf{48.732$\pm$0.358} & 51.615$\pm$0.280 & 46.880$\pm$0.193 & \textbf{$+2.066$} & $-1.852$ \\
\addlinespace
40 & \textsc{Grow} & \textbf{53.588$\pm$0.313} & 48.338$\pm$0.291 & 50.791$\pm$0.338 & 46.886$\pm$0.239 & $-2.797$ & \textbf{$-1.452$} \\
40 & \textsc{Prune} & 49.963$\pm$0.481 & \textbf{48.869$\pm$0.251} & \textbf{51.205$\pm$0.415} & \textbf{47.195$\pm$0.234} & \textbf{$+1.242$} & $-1.675$ \\
\addlinespace
50 & \textsc{Grow} & \textbf{54.338$\pm$0.221} & 48.990$\pm$0.245 & 50.260$\pm$0.257 & \textbf{46.776$\pm$0.174} & $-4.078$ & \textbf{$-2.213$} \\
50 & \textsc{Prune} & 50.227$\pm$0.548 & \textbf{48.997$\pm$0.240} & \textbf{50.320$\pm$0.415} & 46.749$\pm$0.295 & \textbf{$+0.093$} & $-2.247$ \\
\bottomrule
\end{tabular}%
}
\vspace{-0.5em}
\caption{
\textbf{CIFAR-100 (i.i.d.) ConvNet, ReLU, 200 epochs (SGD, $\eta{=}0.1$).} 
Mean $\pm$ CI95 over $n=10$ seeds. 
We report \emph{cycle} evaluation metrics (measured during in-place structural adaptation) and \emph{winning-ticket} evaluation metrics (mask frozen, weights reinitialized, retrained from scratch). 
Deltas are computed \emph{within method and compactness} as $\Delta = (\text{ticket} - \text{cycle})$ (\%). 
Bold highlights the higher value between \textsc{Grow} and \textsc{Prune} within each compactness for each column.}
\label{tab:cifar100_cycle_ticket_deltas}
\vspace{-1.0em}
\end{table}

\vspace{-0.5em}
\paragraph{Statistical comparisons.}
Welch two-sample tests shown in Table~\ref{tab:pvalues_cycle_wt_acc} and Table~\ref{tab:pvalues_cycle_wt_auc} confirm that the \textsc{Grow}--\textsc{Prune} difference is strongest during structural adaptation. 
In-place final ACC differs significantly at every compactness, while in-place final TAA differs at $20$--$40\%$. In contrast, neither Winning-Ticket ACC nor Winning-Ticket TAA differs significantly between the two methods at any compactness. 
Thus, the statistically reliable separation concerns the adaptive process rather than the retrainability of the final masks.

\begin{table}[ht]
\centering
\small
\begin{tabular}{lrrrrrrrr}
\toprule
 & \multicolumn{2}{c}{\textbf{20\%}} & \multicolumn{2}{c}{\textbf{30\%}} & \multicolumn{2}{c}{\textbf{40\%}} & \multicolumn{2}{c}{\textbf{50\%}} \\
\textbf{Comparison} & \textbf{In-place} & \textbf{WT} & \textbf{In-place} & \textbf{WT} & \textbf{In-place} & \textbf{WT} & \textbf{In-place} & \textbf{WT} \\
\midrule
\textsc{Grow} vs \textsc{Dense} & \textbf{0.0000} & \textbf{0.0000} & \textbf{0.0000} & \textbf{0.0000} & \textbf{0.0000} & \textbf{0.0000} & \textbf{0.0000} & \textbf{0.0038} \\
\textsc{Grow} vs \textsc{Prune} & \textbf{0.0000} & 0.9917 & \textbf{0.0000} & 0.9461 & \textbf{0.0000} & 0.0980 & \textbf{0.0000} & 0.7846 \\
\textsc{Dense} vs \textsc{Prune} & \textbf{0.0001} & \textbf{0.0000} & 0.5862 & \textbf{0.0000} & 0.2514 & \textbf{0.0000} & 0.0615 & \textbf{0.0108} \\
\bottomrule
\end{tabular}
\vspace{-0.5em}
\caption{Welch two-sample t-test p-values for final accuracy (ACC), comparing methods across compactness levels for in-place structural adaptation and Winning-Ticket (WT) evaluations. 
Bold indicates statistical significance at $p<0.05$.}
\label{tab:pvalues_cycle_wt_acc}
\vspace{-1.5em}
\end{table}

\begin{table}[ht]
\centering
\small
\begin{tabular}{lrrrrrrrr}
\toprule
 & \multicolumn{2}{c}{\textbf{20\%}} & \multicolumn{2}{c}{\textbf{30\%}} & \multicolumn{2}{c}{\textbf{40\%}} & \multicolumn{2}{c}{\textbf{50\%}} \\
\textbf{Comparison} & \textbf{In-place} & \textbf{WT} & \textbf{In-place} & \textbf{WT} & \textbf{In-place} & \textbf{WT} & \textbf{In-place} & \textbf{WT} \\
\midrule
\textsc{Grow} vs \textsc{Dense} & \textbf{0.0003} & 0.0876 & 0.3701 & \textbf{0.0000} & 0.1425 & \textbf{0.0000} & \textbf{0.0000} & \textbf{0.0000} \\
\textsc{Grow} vs \textsc{Prune} & \textbf{0.0000} & 0.7154 & \textbf{0.0014} & 0.8277 & \textbf{0.0059} & 0.0517 & 0.9637 & 0.8601 \\
\textsc{Dense} vs \textsc{Prune} & \textbf{0.0340} & \textbf{0.0437} & \textbf{0.0040} & \textbf{0.0000} & \textbf{0.0001} & \textbf{0.0000} & \textbf{0.0000} & \textbf{0.0004} \\
\bottomrule
\end{tabular}
\vspace{-0.5em}
\caption{
Welch two-sample t-test p-values for TAA, comparing methods across compactness levels for in-place structural adaptation and Winning-Ticket (WT) evaluations. 
Bold indicates statistical significance at $p<0.05$.
}
\label{tab:pvalues_cycle_wt_auc}
\vspace{-1.0em}
\end{table}

\subsubsection{Cohort-level integration diagnostics}
\label{app:cifar100_convnet_diagnostics}

Structural plasticity changes \emph{who participates} in the computation and \emph{who receives credit} during optimization. 
Section~\ref{sec:allocation_mechanisms} identifies a consistent asymmetry between structural growth and pruning: newly activated units can participate in the forward pass while remaining weakly integrated into the optimization process. 
The following ablations test whether this result depends on particular design choices, including the growth-selection heuristic, the allocation of the structural budget across layers, and the location at which structural edits are introduced.

\vspace{-0.5em}
\paragraph{Cohorts and measurement snapshots.}
Both are computed from the same per-cycle cohort snapshots, but they answer different questions. Absolute diagnostics ask whether newborn, kept, or pruned units exhibit nontrivial forward participation and learning signal; parity diagnostics ask how those cohorts compare \emph{relative} to one another.

For \textsc{Grow}, at cycle $t$ we define the newborn cohort
\[
\mathcal{N}_t = \{\text{units activated at the \textsc{Grow} step of cycle } t\},
\]
and the incumbent cohort
\[
\mathcal{O}_t = \{\text{units already active before that \textsc{Grow} step}\}.
\]
For \textsc{Prune}, we define the kept cohort
\[
\mathcal{K}_t = \{\text{units kept by the \textsc{Prune} decision at cycle } t\},
\]
and the pruned cohort
\[
\mathcal{P}_t = \{\text{units removed by the \textsc{Prune} decision at cycle } t\}.
\]
We use three time-points: 
(i) \emph{Post}, the immediate snapshot after a structural edit and before further training;
(ii) \emph{Exit}, the end-of-cycle checkpoint at which prune decisions are made; and
(iii) \emph{End}, the end of the subsequent training segment.

\vspace{-0.5em}
\paragraph{Absolute activation and gradient signals.}
For a layer $\ell$ and a unit $j$, we measure:
\begin{itemize}
    \item \textbf{Activation rate} $\mathrm{act}(j)$: the fraction of post-activation values exceeding a small threshold $\tau$ on a single mini-batch.
    \item \textbf{Per-unit gradient magnitude} $\mathrm{grad}(j)$: the mean absolute pre-activation gradient $\left|\partial \mathcal{L}/\partial z_j\right|$ on a single mini-batch, where $z_j$ denotes the unit pre-activation.
\end{itemize}
For convolutional layers, activation rates are averaged over $(B,H,W)$; for linear layers, over $B$. We aggregate within cohort by averaging over units and seeds, and in the main-paper figures we additionally average across layers unless otherwise stated.

For \textsc{Grow}, we report newborn absolute vitality through the post-growth cohort means
\[
A^{\mathrm{abs}}_{\mathrm{grow,act}}(t,\ell)
=
\mathbb{E}_{j\in\mathcal{N}_t(\ell)}[\mathrm{act}(j)],
\qquad
A^{\mathrm{abs}}_{\mathrm{grow,grad}}(t,\ell)
=
\mathbb{E}_{j\in\mathcal{N}_t(\ell)}[\mathrm{grad}(j)].
\]
These quantities serve as sanity checks: they ask whether newborn units are active at birth and whether they receive nonzero learning signal. For \textsc{Prune}, we analogously report absolute contrasts such as $\mathbb{E}[\mathrm{act}]$ for kept and pruned cohorts at exit, together with post-prune changes in survivor activation.

\vspace{-0.5em}
\paragraph{Relative cohort signal and log-parity.}
Absolute activation and gradient magnitudes establish whether structural cohorts receive nontrivial forward and backward signal, but their scales vary with training time, loss magnitude, optimizer state, and representation maturity. 
Therefore, we compare cohorts measured at the same snapshot.
For a focal cohort $A$, reference cohort $B$, and signal $x\in\{\mathrm{act},\mathrm{grad}\}$, we define the relative cohort signal
\begin{equation}
R_x(A,B) =
\frac{\mathbb{E}_{j\in A}[x(j)]}
{\mathbb{E}_{j\in B}[x(j)] + \epsilon}.
\end{equation}
In figures intended for direct interpretation, we report $100R_x$: $100\%$ denotes equal per-unit signal, values below $100\%$ indicate that the focal cohort receives less signal, and values above $100\%$ indicate more.
For analyses where a symmetric scale around parity is useful, we additionally use the log-ratio
\begin{equation}
    \Delta_x(A,B)=\log R_x(A,B),
\end{equation}
for which parity corresponds to $\Delta_x=0$.

For \textsc{Grow}, we compare newborn units $N_t^{(\ell)}$ with previously active incumbents $O_t^{(\ell)}$ at the post-growth snapshot:
\begin{equation}
R_{\mathrm{grow},x}(t,\ell) =
\frac{\mathbb{E}_{j\in N_t^{(\ell)}}[x(j)]}{\mathbb{E}_{j\in O_t^{(\ell)}}[x(j)] + \epsilon}.
\end{equation}
When log-parity is required, we use $\Delta_{\mathrm{grow},x}(t,\ell)=\log R_{\mathrm{grow},x}(t,\ell)$.

For \textsc{Prune}, we compare retained units $K_t^{(\ell)}$ with units selected for removal $P_t^{(\ell)}$ using the pre-mask exit snapshot:
\begin{equation}
R_{\mathrm{prune},x}(t,\ell) =
\frac{\mathbb{E}_{j\in K_t^{(\ell)}}[x(j)]}{\mathbb{E}_{j\in P_t^{(\ell)}}[x(j)] + \epsilon}.
\end{equation}

The corresponding log-parity is $\Delta_{\mathrm{prune},x}(t,\ell)=\log R_{\mathrm{prune},x}(t,\ell)$.
Using the pre-mask snapshot prevents the removed cohort from being assigned trivial zero activation and gradient after pruning.

\vspace{-0.5em}
\paragraph{Prune survivor stability.}
To measure how the retained cohort changes after pruning, we compare its immediate post-prune activity with its end-of-cycle activity:
\begin{equation}
R_{\mathrm{surv},\mathrm{act}}
=
\frac{
\mathbb{E}_{j\in K_t}[\mathrm{act}_{\mathrm{post}}(j)]
}{\mathbb{E}_{j\in K_t}[\mathrm{act}_{\mathrm{end}}(j)] + \epsilon}.
\end{equation}
We report $100R_{\mathrm{surv},\mathrm{act}}$, so $100\%$ denotes no change, whereas values below $100\%$ indicate that survivor activity increases during the subsequent training segment.

\begin{figure}[ht]
    \centering
    \includegraphics[width=\linewidth]{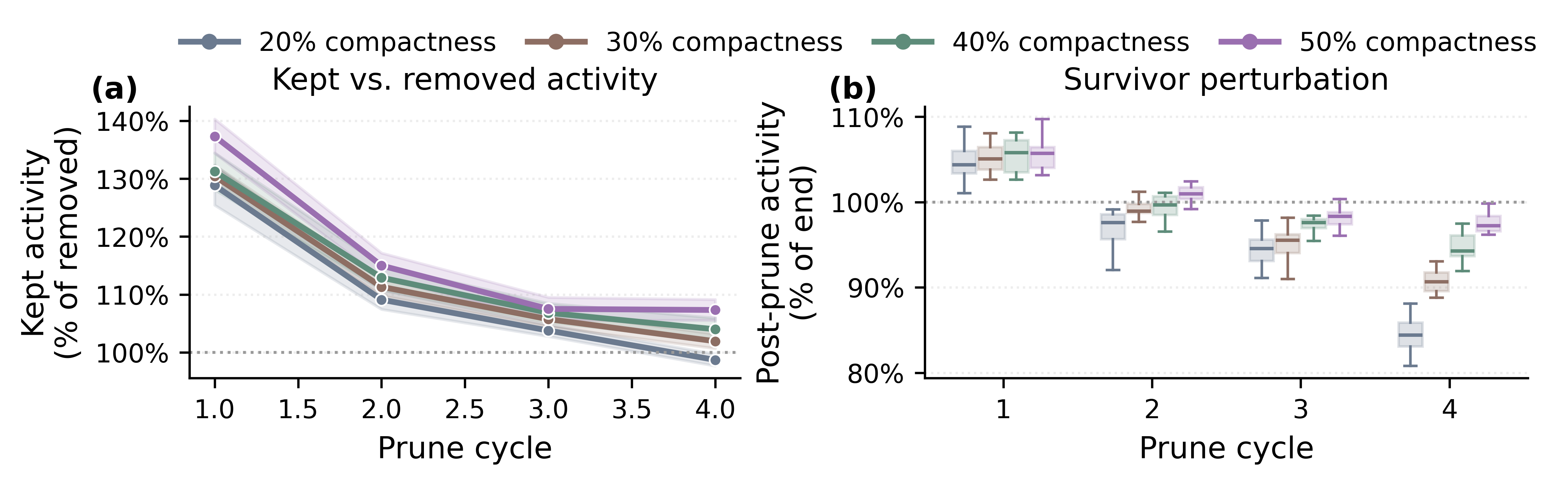}
    \vspace{-2.0em}
        \caption{
        \textbf{Pruning becomes less selective and increasingly perturbs surviving units across structural cycles.}
        \textbf{(a)}  Activity of retained units relative to units selected for removal immediately before each pruning decision, expressed as a percentage of the removed cohort; \(100\%\) denotes equal activity. 
        The decreasing separation across cycles shows that later pruning events have fewer clearly low-activity candidates. 
        \textbf{(b)} Survivor readjustment, measured as post-prune activity relative to end-of-cycle activity; values below \(100\%\) indicate that surviving units increase their activity during the subsequent training segment. 
        Curves, bands, and boxplots summarize means and 95\% confidence intervals across runs.
        }
        \label{fig:convnet_prune_cohort_dynamics}
\end{figure}

\vspace{-0.5em}
\paragraph{Prune selection and survivor readjustment.}
Figure~\ref{fig:convnet_prune_cohort_dynamics} shows the complementary \textsc{Prune} dynamics.
At the first event, retained units reach roughly $128$--$138\%$ of the activity of removed units, but this separation approaches $100\%$ across cycles, indicating increasingly less selective pruning.
Survivors also readjust more strongly: by the final event, immediate post-prune activity falls to about $84\%$ and $90\%$ of end-of-cycle activity at $20\%$ and $30\%$ compactness, respectively.
Thus, \textsc{Prune} initially removes clearly low-activity structure, while later edits become less selective and require greater compensatory adaptation from surviving units.

\vspace{-1.0em}
\subsubsection{Robustness to the \textsc{Grow} selection heuristic}
\label{app:app_gradient_grow_ablation}

The default \textsc{Grow} rule ranks dormant units by their pre-mask activation frequency. 
A natural concern is that this criterion may select units that respond strongly to the input but are poorly aligned with the loss gradient, thereby producing the observed combination of forward participation and weak backward credit.
We test this explanation by replacing activation-based top-$k$ selection with gradient-based top-$k$ selection in the CIFAR-100 ConvNet, while holding the initialization, optimizer, training
budget, compactness schedule, and neutral FC-budget allocation fixed.

Table~\ref{tab:grow_heuristic_ablation_performance} shows no systematic performance advantage from gradient-based selection. 
Across compactness levels, in-place final ACC and in-place TAA remain nearly unchanged, while the winning-ticket results are comparable or slightly lower. 
Therefore, changing the selection signal improves neither the structural-edit trajectory nor the quality of the final retrainable mask.

\begin{table}[ht]
\centering
\small
\setlength{\tabcolsep}{4pt}
\begin{tabular}{r l c c c c}
\toprule
\textbf{Comp. (\%)} & \textbf{\textsc{Grow} heuristic} & \textbf{In-place Final ACC} & \textbf{In-place Final TAA} & \textbf{WT ACC} & \textbf{WT TAA} \\
\midrule
20 & \textsc{Activation} & 52.32 $\pm$ 0.27 & 47.29 $\pm$ 0.23 & 52.05 $\pm$ 0.28 & 46.26 $\pm$ 0.16 \\
20 & \textsc{Gradient} & 52.32 $\pm$ 0.15 & 47.38 $\pm$ 0.12 & 51.72 $\pm$ 0.26 & 46.20 $\pm$ 0.16 \\
30 & \textsc{Activation} & 53.31 $\pm$ 0.21 & 47.87 $\pm$ 0.32 & 51.63 $\pm$ 0.20 & 46.91 $\pm$ 0.19 \\
30 & \textsc{Gradient} & 53.22 $\pm$ 0.28 & 48.09 $\pm$ 0.16 & 51.67 $\pm$ 0.26 & 46.99 $\pm$ 0.12 \\
40 & \textsc{Activation} & 53.59 $\pm$ 0.27 & 48.34 $\pm$ 0.25 & 50.79 $\pm$ 0.29 & 46.89 $\pm$ 0.21 \\
40 & \textsc{Gradient} & 53.78 $\pm$ 0.26 & 48.35 $\pm$ 0.30 & 50.89 $\pm$ 0.31 & 46.93 $\pm$ 0.26 \\
50 & \textsc{Activation} & 54.34 $\pm$ 0.19 & 48.99 $\pm$ 0.21 & 50.26 $\pm$ 0.22 & 46.78 $\pm$ 0.15 \\
50 & \textsc{Gradient} & 54.14 $\pm$ 0.12 & 48.95 $\pm$ 0.18 & 50.12 $\pm$ 0.38 & 46.63 $\pm$ 0.16 \\
\bottomrule
\end{tabular}
\vspace{-1.0em}
\caption{
    \textbf{\textsc{Grow} heuristic ablation.} 
    We compare activation-based and gradient-based top-$k$ \textsc{Grow} under neutral allocation bias. 
    This tests whether the newborn gradient bottleneck is simply caused by selecting growth locations using activation statistics. 
    In-place structural adaptation metrics evaluate the adaptive structural trajectory; WT metrics freeze the final mask and retrain it from scratch. Values are mean $\pm$ CI95 over 20 seeds.
}
\label{tab:grow_heuristic_ablation_performance}
\end{table}

\vspace{-1.0em}
\subsection{ConvNet FC-head allocation-bias ablation}
\label{sec:app_cifar100_bias_ablation}
\vspace{-0.5em}

Because the main CIFAR-100 experiments use a neutral layer-allocation schedule, we test whether the observed \textsc{Grow} behavior depends on how compactness is distributed across the FC head.
Table~\ref{tab:bias_schedules} defines the neutral schedule and alternatives that mildly favor early, late, or both ends of the head.
Table~\ref{tab:grow_bias_cycle_ticket_deltas} shows only moderate sensitivity: different schedules shift the trade-off between in-place performance and ticket retrainability, but none consistently improves both across compactness levels.
Thus, modest FC-head budget redistribution does not remove the newborn-integration limitation.

\vspace{-0.5em}
\begin{table}[ht]
\centering
\small
\begin{tabular}{lccc}
\toprule
Schedule & $b_{\texttt{FC1}}$ & $b_{\texttt{FC2}}$ & $b_{\texttt{FC3}}$ \\
\midrule
Neutral         & 1.0 & 1.0 & 1.0 \\
FC1-Protect     & 1.5 & 1.5 & 0.6 \\
FC3-Protect     & 0.6 & 0.6 & 1.5 \\
Ends-Skewed     & 1.2 & 0.6 & 1.2 \\
\bottomrule
\end{tabular}
\vspace{-1.0em}
\caption{
    \textbf{Bias scalars used to distribute the global kept-weight budget across masked fully connected layers.}
    The neutral schedule allocates proportionally to layer weight mass, while the remaining schedules mildly favor early layers, late layers, or both ends of the head.
    }
\label{tab:bias_schedules}
\vspace{-0.5em}
\end{table}

\begin{table}[ht]
\centering
\small
\setlength{\tabcolsep}{3.5pt}
\resizebox{\linewidth}{!}{%
\begin{tabular}{r l r r r r r r}
\toprule
& & \multicolumn{2}{c}{\textbf{In-place Eval.}} & \multicolumn{2}{c}{\textbf{Winning-ticket Eval.}} & \multicolumn{2}{c}{\textbf{Ticket $-$ In-place}} \\
\cmidrule(lr){3-4}\cmidrule(lr){5-6}\cmidrule(lr){7-8}
\textbf{Comp. (\%)} & \textbf{Method} & \textbf{ACC} & \textbf{TAA} & \textbf{ACC} & \textbf{TAA} & $\boldsymbol{\Delta\mathrm{ACC}}$ & $\boldsymbol{\Delta\mathrm{TAA}}$ \\
\midrule
20 & \textsc{Grow} & 52.316$\pm$0.317 & 47.288$\pm$0.269 & \textbf{52.055$\pm$0.324} & \textbf{46.257$\pm$0.188} & $-0.261$ & \textbf{$-1.031$} \\
20 & \textsc{Grow} (FC3-Protect) & \textbf{53.016$\pm$0.393} & \textbf{47.717$\pm$0.461} & 51.287$\pm$0.314 & 46.085$\pm$0.175 & $-1.729$ & $-1.632$ \\
20 & \textsc{Grow} (Ends-Skewed) & 51.926$\pm$0.153 & 46.987$\pm$0.196 & 51.757$\pm$0.259 & 45.758$\pm$0.190 & \textbf{$-0.169$} & $-1.230$ \\
20 & \textsc{Grow} (FC1-Protect) & 51.794$\pm$0.201 & 46.830$\pm$0.247 & 51.622$\pm$0.276 & 45.721$\pm$0.140 & $-0.172$ & $-1.109$ \\
\addlinespace
30 & \textsc{Grow} & 53.312$\pm$0.247 & 47.871$\pm$0.374 & 51.626$\pm$0.232 & \textbf{46.909$\pm$0.220} & $-1.686$ & $-0.963$ \\
30 & \textsc{Grow} (FC3-Protect) & \textbf{53.564$\pm$0.345} & \textbf{48.235$\pm$0.284} & 50.232$\pm$0.290 & 46.295$\pm$0.161 & $-3.332$ & $-1.940$ \\
30 & \textsc{Grow} (Ends-Skewed) & 52.819$\pm$0.220 & 47.650$\pm$0.326 & \textbf{52.076$\pm$0.257} & 46.876$\pm$0.189 & \textbf{$-0.743$} & \textbf{$-0.775$} \\
30 & \textsc{Grow} (FC1-Protect) & 52.647$\pm$0.395 & 47.469$\pm$0.172 & 51.879$\pm$0.303 & 46.648$\pm$0.194 & $-0.768$ & $-0.821$ \\
\addlinespace
40 & \textsc{Grow} & 53.588$\pm$0.313 & 48.338$\pm$0.291 & 50.791$\pm$0.338 & 46.886$\pm$0.239 & $-2.797$ & $-1.452$ \\
40 & \textsc{Grow} (FC3-Protect) & \textbf{53.761$\pm$0.283} & \textbf{48.424$\pm$0.257} & 49.550$\pm$0.369 & 46.105$\pm$0.219 & $-4.211$ & $-2.319$ \\
40 & \textsc{Grow} (Ends-Skewed) & 53.409$\pm$0.257 & 48.146$\pm$0.389 & \textbf{51.670$\pm$0.311} & \textbf{47.129$\pm$0.231} & \textbf{$-1.739$} & \textbf{$-1.017$} \\
40 & \textsc{Grow} (FC1-Protect) & 53.305$\pm$0.258 & 48.191$\pm$0.101 & 51.263$\pm$0.359 & 46.795$\pm$0.187 & $-2.042$ & $-1.396$ \\
\addlinespace
50 & \textsc{Grow} & \textbf{54.338$\pm$0.221} & \textbf{48.990$\pm$0.245} & 50.260$\pm$0.257 & 46.776$\pm$0.174 & $-4.078$ & $-2.213$ \\
50 & \textsc{Grow} (FC3-Protect) & 54.206$\pm$0.259 & 48.941$\pm$0.230 & 49.415$\pm$0.258 & 46.293$\pm$0.165 & $-4.791$ & $-2.647$ \\
50 & \textsc{Grow} (Ends-Skewed) & 53.759$\pm$0.225 & 48.272$\pm$0.298 & \textbf{50.983$\pm$0.337} & \textbf{47.057$\pm$0.253} & \textbf{$-2.776$} & \textbf{$-1.215$} \\
50 & \textsc{Grow} (FC1-Protect) & 53.791$\pm$0.264 & 48.340$\pm$0.246 & 50.884$\pm$0.258 & 46.887$\pm$0.210 & $-2.907$ & $-1.453$ \\
\bottomrule
\end{tabular}%
}
\vspace{-1.0em}
\caption{\textbf{CIFAR-100 (i.i.d.) ConvNet, ReLU, 200 epochs (SGD, $\eta{=}0.1$).} Mean $\pm$ CI95 over $n=10$ seeds. We report \emph{In-place} evaluation metrics (measured during structural adaptation) and \emph{winning-ticket} evaluation metrics (mask frozen, weights reinitialized, retrained from scratch). Deltas are computed \emph{within method and compactness} as $\Delta = (\text{ticket} - \text{In-place})$ (\%). Bold highlights the highest value among the \textsc{Grow} bias schedules within each compactness for each column.}
\label{tab:grow_bias_cycle_ticket_deltas}
\vspace{-1.5em}
\end{table}



\vspace{-0.5em}
\subsection{CIFAR-100 ResNet: Additional Locus Analyses}
\label{app:cifar100_resnet}
\vspace{-0.5em}

The main text reports the central ResNet finding: the representation and future adaptability produced by structural growth depend strongly on where new channels are introduced. 
This section provides the complete compactness sweeps, event-local integration diagnostics, representation decompositions, and future-probe robustness analyses underlying that result. Architecture, edit-locus, and compactness definitions are given in Apps.~\ref{app:resnet_arch} and \ref{app:budgets_compactness}.

\subsubsection{Cross-metric summary}
\label{app:resnet_summary}

Table~\ref{tab:resnet-main-summary} collects the current-task, representation, and future-adaptation quantities reported in the main text.
The table highlights the progressive separation between head-localized and residual-hierarchy growth: broader edit loci reduce in-place final ACC, effective feature support, and the final future-adaptation endpoint, even when final winning-ticket accuracy remains comparatively close to \textsc{Prune}.

\begin{table*}[ht]
\centering
\scriptsize
\setlength{\tabcolsep}{3.0pt}
\resizebox{\textwidth}{!}{%
\begin{tabular}{llccccccc}
\toprule
Method & Locus & \shortstack{In-place\\ACC} & \shortstack{WT\\ACC} & \shortstack{Total\\variance} & \shortstack{Norm.\\eff. rank} & \shortstack{Linear\\probe} & \shortstack{Future\\AUC} & \shortstack{Final future\\accuracy} \\
\midrule
\textsc{Dense} & -- & 63.1 $\pm$ 0.3 & 63.1 $\pm$ 0.3 & 35.4 $\pm$ 0.4 & 0.588 $\pm$ 0.002 & 55.1 $\pm$ 0.2 & 35.2 $\pm$ 0.7 & 51.3 $\pm$ 1.0 \\
\addlinespace[1.5pt]
\midrule
\textsc{Prune} & Head & 61.9 $\pm$ 0.3 & 61.3 $\pm$ 0.4 & 30.8 $\pm$ 1.0 & 0.603 $\pm$ 0.004 & 56.2 $\pm$ 0.3 & 33.6 $\pm$ 0.6 & 50.6 $\pm$ 0.9 \\
\textsc{Grow} & Head & 61.2 $\pm$ 0.3 & 61.7 $\pm$ 0.4 & 53.6 $\pm$ 0.4 & 0.516 $\pm$ 0.002 & 57.1 $\pm$ 0.1 & 36.5 $\pm$ 0.7 & 50.7 $\pm$ 0.6 \\
\addlinespace[1.5pt]
\textsc{Prune}  & Late & 60.4 $\pm$ 0.2 & 60.5 $\pm$ 0.3 & 13.3 $\pm$ 2.0 & 0.514 $\pm$ 0.009 & 52.1 $\pm$ 0.3 & 32.5 $\pm$ 0.9 & 48.5 $\pm$ 0.7 \\
\textsc{Grow} & Late & 61.4 $\pm$ 0.2 & 60.2 $\pm$ 0.2 & 23.2 $\pm$ 0.1 & 0.416 $\pm$ 0.002 & 50.4 $\pm$ 0.2 & 32.9 $\pm$ 0.4 & 46.6 $\pm$ 0.5 \\
\addlinespace[1.5pt]
\textsc{Prune}  & Mid+late & 59.2 $\pm$ 0.4 & 59.5 $\pm$ 0.4 & 11.6 $\pm$ 1.7 & 0.510 $\pm$ 0.010 & 50.9 $\pm$ 0.4 & 31.9 $\pm$ 0.7 & 47.9 $\pm$ 0.5 \\
\textsc{Grow} & Mid+late & 59.5 $\pm$ 0.3 & 59.7 $\pm$ 0.3 & 19.1 $\pm$ 0.1 & 0.400 $\pm$ 0.002 & 47.0 $\pm$ 0.2 & 29.5 $\pm$ 0.5 & 43.3 $\pm$ 0.5 \\
\addlinespace[1.5pt]
\textsc{Prune}  & All & 58.0 $\pm$ 0.2 & 59.0 $\pm$ 0.3 & 11.3 $\pm$ 1.6 & 0.509 $\pm$ 0.011 & 49.7 $\pm$ 0.5 & 30.9 $\pm$ 1.0 & 46.2 $\pm$ 0.6 \\
\textsc{Grow} & All & 58.5 $\pm$ 0.2 & 59.2 $\pm$ 0.4 & 17.8 $\pm$ 0.1 & 0.403 $\pm$ 0.002 & 45.2 $\pm$ 0.2 & 28.4 $\pm$ 0.4 & 41.9 $\pm$ 0.5 \\
\bottomrule
\end{tabular}%
}
\vspace{-0.5em}
    \caption{
        \textbf{ResNet current-task performance, representation state, and bounded future reachability at 50\% final compactness.} 
        Representation scale and effective rank are reported descriptively and are not ranked.
        Values are means $\pm$ 95\% confidence intervals. Dense is locus-agnostic and is reported once.
    }
\label{tab:resnet-main-summary}
\end{table*}

\vspace{-0.5em}
\subsubsection{Pruning protocol and capacity-path control.}
\label{app:prune_protocol}
The main \textsc{Grow}--\textsc{Prune} comparison contrasts opposite structural operators, but two procedural asymmetries can affect In-place TAA: the default pruning baseline uses IMP-style rewinding whereas \textsc{Grow} updates in place, and pruning descends from a fully active head while growth ascends from a sparse seed. 
Therefore, we repeat the \texttt{CIFAR-100} ResNet HEAD comparison over the 20--30--40--50\% compactness sweep and add \textsc{Prune}-FT, which follows the same pruning schedule without rewinding surviving weights. 
We report in-place TAA (Cycle-TAA) together with in-place final ACC at the matched target compactness. 

\begin{figure*}[ht]
    \centering
    \includegraphics[width=\textwidth]{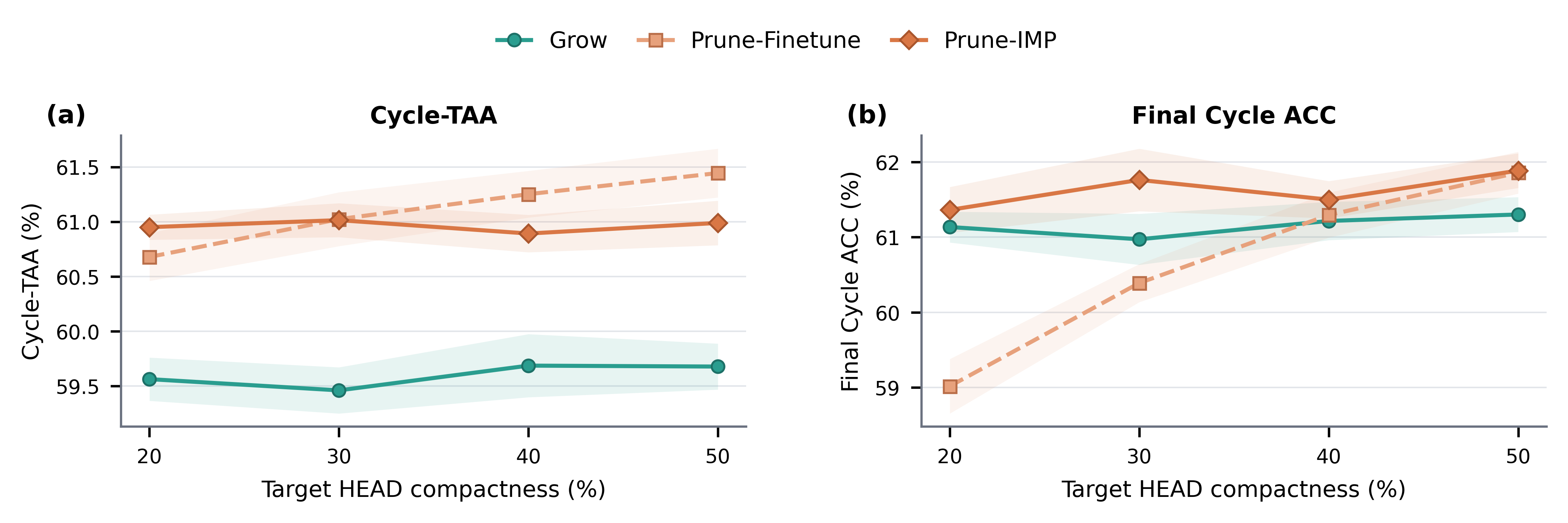}
    \vspace{-2.0em}
    \caption{
        \textbf{Pruning-protocol control across ResNet HEAD compactness.} 
        (a) Cycle-TAA across target compactness for \textsc{Grow}, non-rewind \textsc{Prune}-FT, and rewind-based \textsc{Prune}-IMP. 
        Because Cycle-TAA averages over the complete structural trajectory, it also reflects the greater active capacity available to pruning before the final target is reached. 
        (b) Final Cycle ACC after each method reaches the same target HEAD compactness. 
        Points show means across experiment indices and shaded bands show 95\% confidence intervals. 
        \textsc{Prune}-FT removes the IMP reset while preserving the pruning schedule, separating reset-protocol effects from the capacity-path asymmetry.
    }
    \label{fig:prune_protocol}
\end{figure*}

\vspace{-0.5em}
\paragraph{Results and interpretation.}
At 50\% compactness, removing rewind changes In-place TAA by only +0.46 points and In-place Final ACC by -0.03 points.
\textsc{Grow} remains lower in In-place TAA by 1.77/1.31 points relative to \textsc{Prune}-FT/IMP, but only 0.56/0.58 points lower at the final cycle.
Mean active \textsc{Head} capacity over the trajectory is 38.2\% for \textsc{Grow} versus 65.0\% for both pruning variants.
Thus, In-place TAA reflects unequal capacity exposure as well as editing dynamics; \textsc{Prune}-FT isolates the rewind convention, while In-place Final ACC and WT-ACC provide matched-target endpoint controls.
We retain IMP as the canonical lottery-ticket baseline.

\vspace{-0.5em}
\subsubsection{Event-local integration dynamics}
\label{app:integration_dynamics}

Table~\ref{tab:resnet_integration_representation_correlations} reports partial Spearman correlations between newborn integration and subsequent representation quality, controlling for final compactness.
Associations are strongly locus dependent. 
At \textsc{Head}, stronger newborn activation is associated with greater class separation ($\rho_{\mathrm{partial}}=0.378$) and lower normalized effective rank ($\rho_{\mathrm{partial}}=-0.480$), with weaker but similar trends for gradient signal. 
Under \textsc{Mid+Late}, stronger gradient signal likewise tracks greater class separation and lower effective rank, while relationships are weak or absent for \textsc{Late} and \textsc{All}.
Neither activation nor gradient integration reliably predicts linear-probe accuracy at any locus.
Thus, newborn integration relates more consistently to representation geometry than to downstream feature utility: it constrains how growth is absorbed, but does not by itself determine the usefulness of the resulting representation.

\begin{table}[ht]
\centering
\scriptsize
\setlength{\tabcolsep}{3.4pt}
\begin{tabular}{lllrr}
\toprule
\textsc{Integration signal} & \textsc{Representation outcome} & \textsc{Locus} & \textsc{Partial Spearman $\rho$} & \textsc{$p$} \\
\midrule
\textbf{Activation parity} & \textit{Class separation} & \texttt{Head} & 0.378 & <.001\\
 &  & \texttt{Late} & 0.084 & .120\\
 &  & \texttt{Mid+Late} & -0.007 & .895\\
 &  & \texttt{All} & -0.013 & .810\\
 \midrule
\addlinespace[0.7pt]
 & \textit{Normalized effective rank} & \texttt{Head} & -0.480 & <.001\\
 &  & \texttt{Late} & 0.002 & .968\\
 &  & \texttt{Mid+Late} & 0.037 & .506\\
 &  & \texttt{All} & 0.053 & .331\\
 \midrule
\addlinespace[0.7pt]
 & \textit{Linear probe accuracy (\%)} & \texttt{Head} & -0.083 & .127\\
 &  & \texttt{Late} & 0.009 & .867\\
 &  & \texttt{Mid+Late} & -0.044 & .425\\
 &  & \texttt{All} & -0.040 & .463\\
\midrule
\textbf{Gradient parity} & \textit{Class separation} & \texttt{Head} & 0.184 & <.001\\
 &  & \texttt{Late} & -0.050 & .360\\
 &  & \texttt{Mid+Late} & 0.126 & .022\\
 &  & \texttt{All} & -0.050 & .359\\
 \midrule
\addlinespace[0.7pt]
 & \textit{Normalized effective rank} & \texttt{Head} & -0.191 & <.001\\
 &  & \texttt{Late} & 0.020 & .714\\
 &  & \texttt{Mid+Late} & -0.286 & <.001\\
 &  & \texttt{All} & 0.042 & .441\\
 \midrule
\addlinespace[0.7pt]
 & \textit{Linear probe accuracy (\%)} & \texttt{Head} & -0.040 & .464\\
 &  & \texttt{Late} & -0.015 & .790\\
 &  & \texttt{Mid+Late} & -0.063 & .256\\
 &  & \texttt{All} & -0.086 & .115\\
\bottomrule
\end{tabular}
\caption{
    \textbf{Partial Spearman correlations between newborn integration and representation quality.}
    Integration statistics are first averaged across editable growth layers within each event, and then, over structural cycles within each replicate and compactness, and correlations control for final compactness. 
    For newborn-log diagnostics, integration at birth in cycle $c$ is aligned with representation diagnostics at the subsequent checkpoint in cycle $c+1$. 
    Reported $p$-values are descriptive and are not corrected for multiple comparisons.
}
\label{tab:resnet_integration_representation_correlations}
\end{table}

\vspace{-0.5em}
\subsubsection{Representation robustness across compactness and stages}
\label{app:resnet_representation_robustness}

\begin{figure*}[ht]
    \centering
    \includegraphics[width=\textwidth]{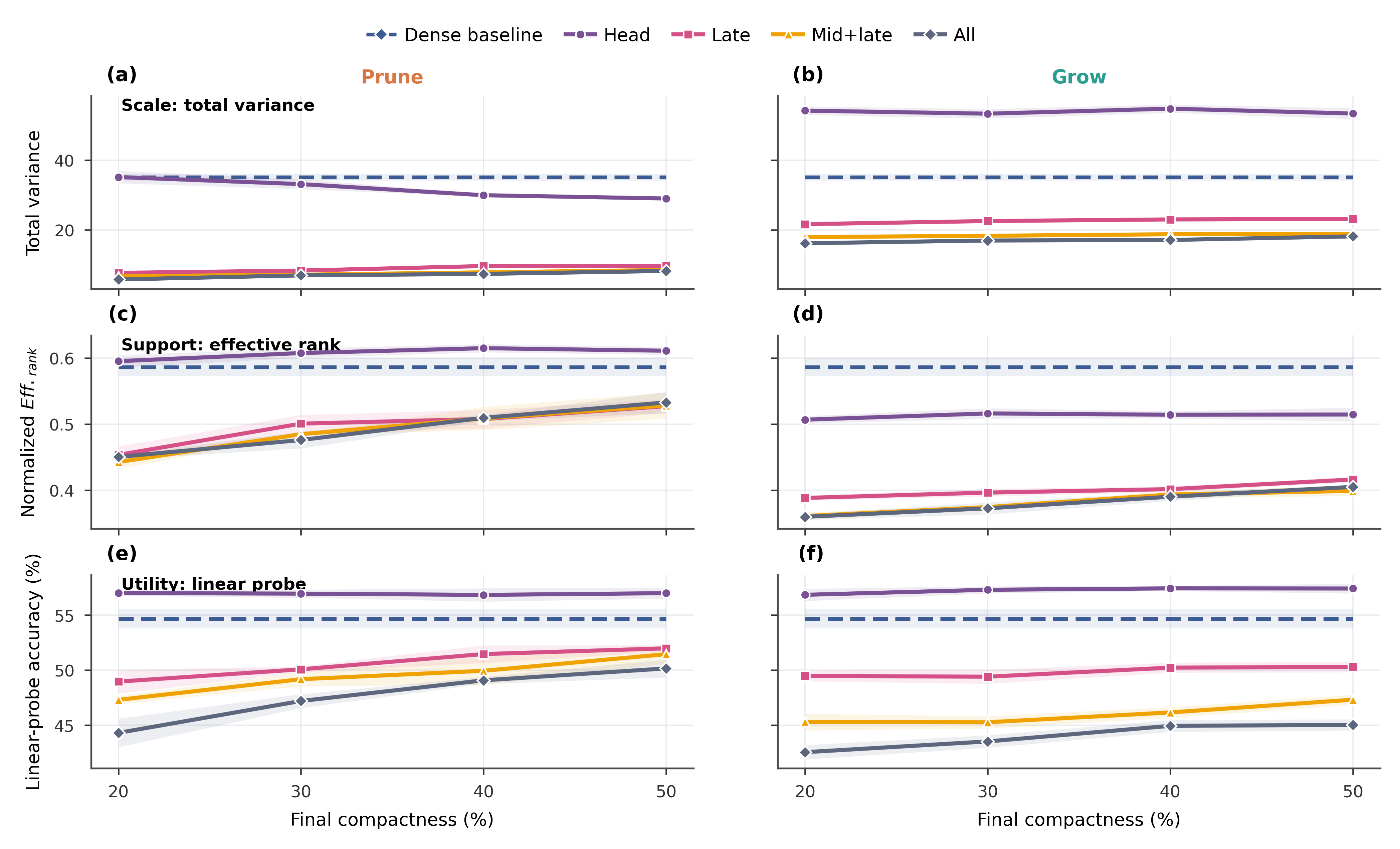}
    \vspace{-2.0em}
    \caption{
        \textbf{Representation robustness across compactness.}
        Final-cycle representation diagnostics at \texttt{stage3\_out} for $20$--$50\%$ compactness. 
        Rows report feature scale (total covariance variance), normalized effective rank, and linear-probe accuracy; columns compare \textsc{Prune} and \textsc{Grow}. The dashed line denotes the
        Dense baseline, which is locus-agnostic. 
        Across compactness values, head-localized growth increases feature scale and probe accuracy while reducing effective support, whereas growth within the residual hierarchy preserves more variance than \textsc{Prune} but yields lower effective rank and lower linear readability. 
        Error bands show $95\%$ confidence intervals.
    }
    \label{fig:resnet_representation_compactness}
\end{figure*}

\begin{figure*}[ht]
    \centering
    \includegraphics[width=\textwidth]{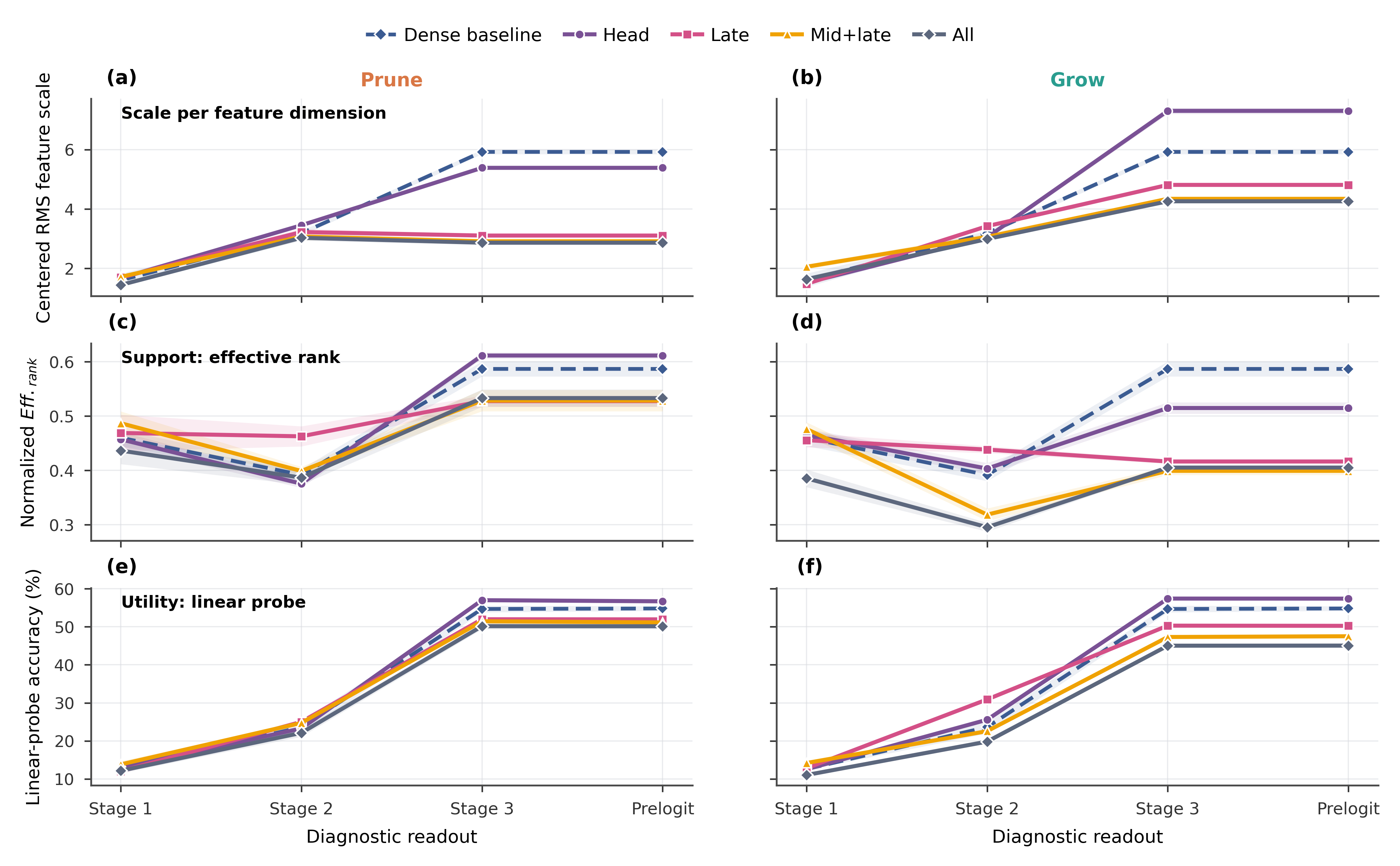}
    \vspace{-2.0em}
    \caption{
        \textbf{Stagewise representation diagnostics at $50\%$ compactness.}
        Final-cycle diagnostics across \texttt{stage1\_out}, \texttt{stage2\_out}, \texttt{stage3\_out}, and \texttt{prelogit}. 
        The dashed line denotes the Dense baseline. Head-localized edits remain close to Dense through the residual hierarchy and diverge mainly at the later readouts, whereas residual-hierarchy edits produce progressively weaker support and utility by \texttt{stage3\_out} and \texttt{prelogit}, especially for \textsc{Grow}. 
        Shaded regions show $95\%$ confidence intervals.
    }
    \label{fig:resnet_stagewise_representation}
\end{figure*}

We focus on final-cycle \texttt{stage3\_out} at $50\%$ compactness in the main text, where the tradeoff between current utility and bounded future adaptability is clearest.
Here we test its robustness across compactness and depth using the same scale--support--utility diagnostics:
(1) a compactness sweep at \texttt{stage3\_out}, and
(2) a $50\%$ stagewise analysis across \texttt{stage1\_out}, \texttt{stage2\_out}, \texttt{stage3\_out}, and \texttt{prelogit}.
Figures~\ref{fig:resnet_representation_compactness} and \ref{fig:resnet_stagewise_representation} show that the main pattern persists across compactness and emerges most strongly in later task-facing stages.

The compactness sweep confirms that the main-text $50\%$ result is not isolated.
Under \textsc{Prune}, \textsc{Head} remains closest to \textsc{Dense} in support and utility, while broader residual loci reduce variance and linear-probe accuracy but retain relatively high effective rank.
Under \textsc{Grow}, \textsc{Head} consistently increases feature scale and probe accuracy while concentrating variance into fewer directions, whereas residual-hierarchy growth yields lower effective rank and linear utility than \textsc{Prune}, most strongly under \textsc{All}.
Thus, the contrast between head-localized and deep feature-hierarchy growth persists across compactness levels.

The stagewise view clarifies how the final \texttt{stage3\_out} differences emerge.
Under \textsc{Prune}, \textsc{Head} stays close to \textsc{Dense}, while deeper loci reduce scale and linear utility but largely recover effective support by \texttt{stage3\_out}.
Under \textsc{Grow}, \textsc{Head} diverges mainly at later readouts, where scale and probe accuracy increase sharply, whereas residual growth degrades representations earlier and the deficit persists downstream.
\textsc{Mid+Late} and \textsc{All} already lose support by \texttt{stage2\_out} and remain below \textsc{Dense} in effective rank and linear-probe accuracy at \texttt{stage3\_out} and \texttt{prelogit}.
Thus, broad growth yields representations that are not only smaller in scale but also less supported and less linearly useful.

\vspace{-0.5em}
\subsubsection{Interventions decouple current performance from future adaptability.}
\label{app:interventions_transfer}

\begin{figure*}[ht]
    \centering
  \includegraphics[width=\textwidth]{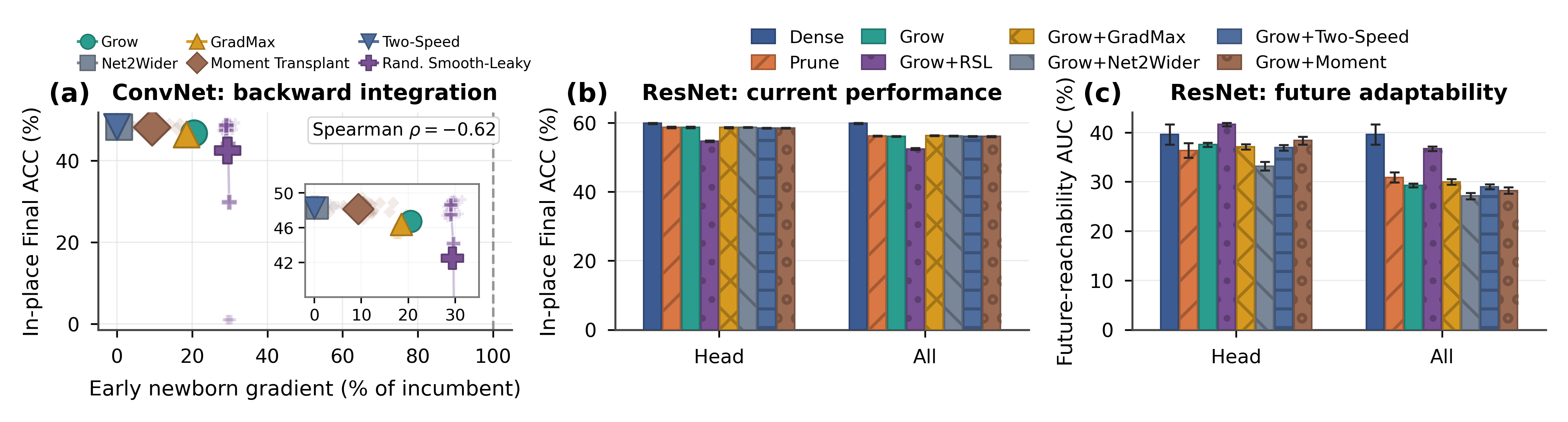}
    \vspace{-2em}
    \caption{
        \textbf{Interventions decouple newborn integration, current performance, and future adaptability.}
        \textbf{(a)} Across ConvNet interventions and compactness levels, stronger early newborn gradient signal is not associated with higher in-place ACC (Spearman $\rho=-0.62$); 100\% denotes equal backward signal.
        Large markers denote method means and lighter points individual runs.
        \textbf{(b)} On ResNet, most interventions produce only modest changes in in-place ACC under \textsc{Head} and \textsc{All}; \textsc{Rand.\ Smooth-Leaky} improves newborn trainability despite lower current accuracy.
        \textbf{(c)} Future-reachability AUC follows a different ordering: \textsc{Rand.\ Smooth-Leaky} is the strongest growth variant at both loci, particularly under \textsc{All}, despite weaker in-place ACC.
        Bars show means with 95\% confidence intervals.
    }
  \label{fig:birth_interventions}
\end{figure*}

Figure~\ref{fig:birth_interventions} shows that the interventions do not improve newborn integration, current performance, and future adaptability in lockstep.
Across ConvNet interventions and compactness levels, early newborn gradient signal is not positively associated with in-place ACC (Spearman $\rho=-0.62$), so better local backward integration does not by itself predict a better current endpoint.
The same decoupling appears in ResNet at $50\%$ compactness.
Under \textsc{Head}, vanilla \textsc{Grow} reaches about $58\%$ in-place ACC and most growth variants remain within roughly one point of it, whereas \textsc{Grow+RSL} drops to about $54\%$.
Under \textsc{All}, the pattern is similar: most variants stay near vanilla \textsc{Grow}, while \textsc{Grow+RSL} is again lower in current accuracy.
Future adaptability follows a different ordering.
Under \textsc{Head}, \textsc{Grow+RSL} reaches about $42$ AUC, above vanilla \textsc{Grow} ($\sim 38$), \textsc{Prune} ($\sim 36$), and the other growth variants.
Under \textsc{All}, it remains the strongest growth variant at about $37$ AUC, compared with roughly $29$--$30$ for vanilla \textsc{Grow}.
Thus, interventions can change the optimization state left by growth---especially its later adaptability---without improving the immediate in-place endpoint.

\vspace{-0.5em}
\subsubsection{Interventions at intermediate edit loci.}
\label{app:interventions_loci}

\begin{figure*}[ht]
    \centering
    \includegraphics[width=\textwidth]{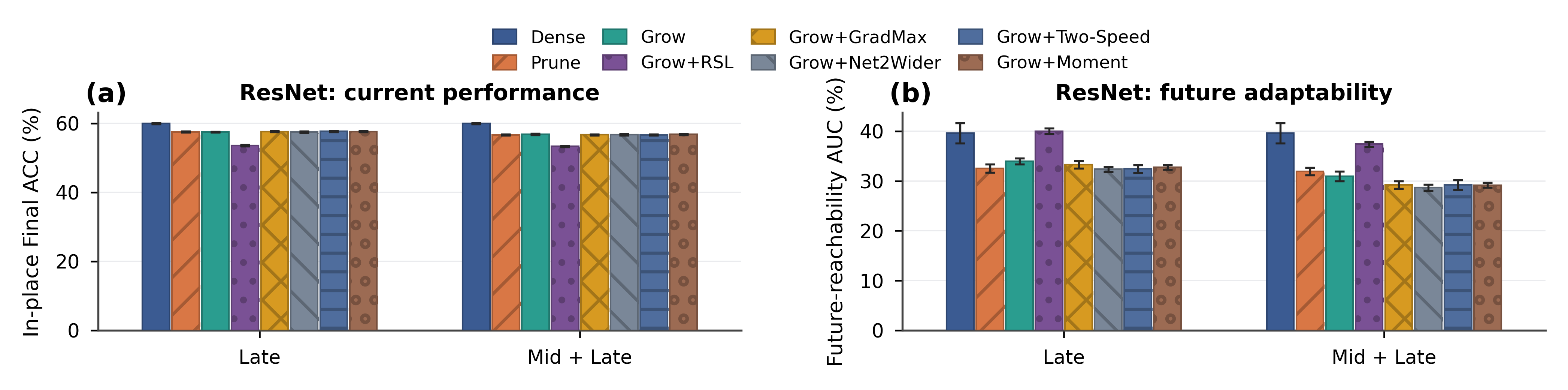}
    \caption{
        \textbf{Intervention effects persist at intermediate ResNet edit loci.}
        Results are shown at $50\%$ compactness for edits confined to the final residual stage (\textsc{Late}) or the final two stages (\textsc{Mid+Late}).
        \textbf{(a)} Most interventions produce only modest changes in in-place final ACC, while \textsc{Rand.\ Smooth-Leaky} lowers current accuracy relative to vanilla \textsc{Grow}.
        \textbf{(b)} The ordering changes for bounded future adaptation: \textsc{Grow+RSL} achieves the highest future-reachability AUC among the growth variants at both loci, approximately matching \textsc{Dense} under \textsc{Late} and substantially narrowing the gap under \textsc{Mid+Late}.
        The remaining interventions provide smaller or less consistent gains over vanilla \textsc{Grow}.
        Bars show means with 95\% confidence intervals.
    }
    \label{fig:resnet_interventions_intermediate}
\end{figure*}

To bridge the \textsc{Head} and \textsc{All} cases from the main text, we repeat the intervention comparison under \textsc{Late} and \textsc{Mid+Late}, reporting in-place final ACC and future-reachability AUC.
Figure~\ref{fig:resnet_interventions_intermediate} extends the same decoupling trend.
Current performance changes little across most interventions: under \textsc{Late}, vanilla \textsc{Grow} reaches about $58\%$ in-place ACC and the other growth variants remain close to it, while \textsc{Grow+RSL} is lower at roughly $54\%$.
Under \textsc{Mid+Late}, the pattern is similar: most growth variants cluster near $57\%$, whereas \textsc{Grow+RSL} remains lower at about $53\%$.
Future adaptability differs more strongly.
\textsc{Grow+RSL} raises AUC from about $34$ to $40$ under \textsc{Late}, and from about $31$ to $37$--$38$ under \textsc{Mid+Late}, while the remaining interventions stay closer to vanilla \textsc{Grow} or below it.
Together with the \textsc{Head}/\textsc{All} results, this shows that activation-level trainability can improve bounded future adaptability without improving the immediate endpoint of structural growth.

\vspace{-0.5em}
\subsubsection{Class-separation decomposition}
\label{app:class_separation}

The main text uses total covariance variance to measure the overall scale of the learned representation, but this quantity does not reveal whether that variation separates classes or instead reflects dispersion within classes.
To distinguish these possibilities, we decompose the feature geometry into between- and within-class scatter. 
Viewing each class as a cloud in feature space, between-class scatter measures the distance among class centers, whereas within-class scatter measures the spread of examples around their own center. 
We evaluate both components at the final cycle and $50\%$ compactness:
\begin{align}
S_B &= \sum_{c=1}^{C} n_c(\mu_c-\mu)(\mu_c-\mu)^\top,\\
S_W &= \sum_{c=1}^{C} \sum_{i:y_i=c} (h_i-\mu_c)(h_i-\mu_c)^\top,
\end{align}
and report the per-dimension traces
$\operatorname{tr}(S_B)/d$ and $\operatorname{tr}(S_W)/d$, together with 
\begin{align}
\operatorname{Sep}(H)= \operatorname{tr}(S_B)/(\operatorname{tr}(S_W)+\epsilon)
\end{align}

\begin{figure*}[ht]
    \centering
    \includegraphics[width=\textwidth]{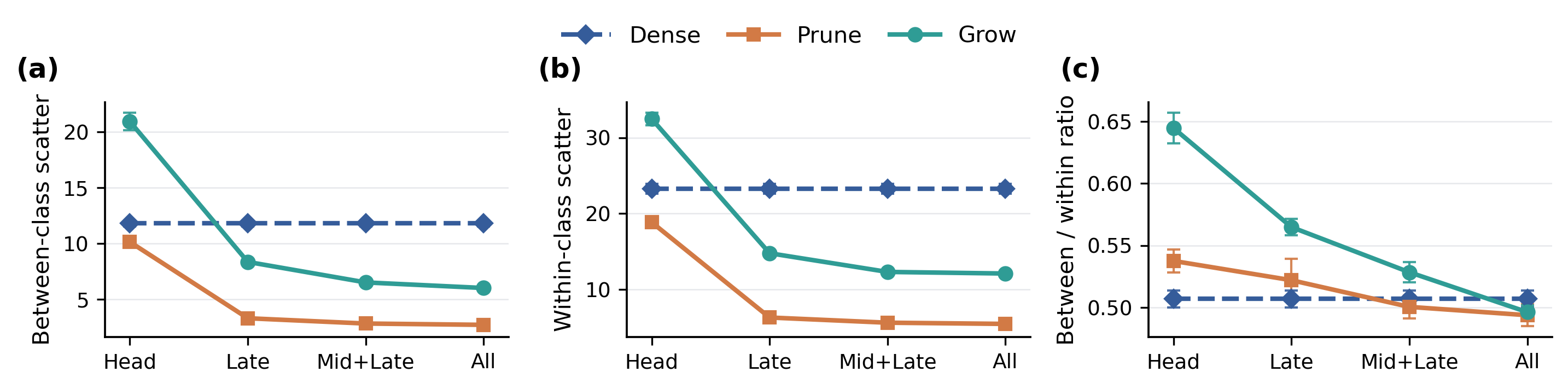}
    \vspace{-2.0em}
    \caption{
        \textbf{Decomposition of class-separation geometry.}
        Final-cycle \texttt{stage3\_out} features at $50\%$ compactness.
        \textbf{(a)} Between-class scatter per feature dimension.
        \textbf{(b)} Within-class scatter per feature dimension.
        \textbf{(c)} Their ratio.
        \textsc{Head}-localized \textsc{Grow} expands both components, with a larger proportional increase in between-class scatter. 
        Residual-hierarchy edits instead contract the feature geometry, and apparently favorable separation ratios can partly result from reduced within-class spread rather than stronger absolute class structure. 
        The dashed line denotes the locus-agnostic \textsc{Dense} baseline; error bars show $95\%$ confidence intervals.
    }
    \label{fig:resnet_class_separation_decomposition}
\end{figure*}

Figure~\ref{fig:resnet_class_separation_decomposition} shows that similar separation ratios can reflect different geometries.
Under \textsc{Head}, \textsc{Grow} increases both between- and within-class scatter relative to \textsc{Dense}, with the larger between-class increase yielding the highest separation ratio.
Residual-hierarchy edits instead reduce both components; \textsc{Grow} can retain a high ratio partly because within-class spread also contracts, without corresponding gains in effective rank or linear-probe accuracy.
Thus, head-localized growth amplifies task-aligned class structure, whereas broad growth weakens the feature hierarchy; \textsc{Prune} preserves mature pathways but compresses the representation.

\vspace{-0.5em}
\subsubsection{Sensitivity to growth-event frequency}
\label{app:resnet_growth_frequency}

We vary growth frequency while fixing the total structural-training budget at $200$ epochs, comparing $K\in\{5,10,20\}$ events with $40$, $20$, and $10$ epochs between edits, respectively.

\begin{figure*}[ht]
    \centering
    \includegraphics[width=\textwidth]{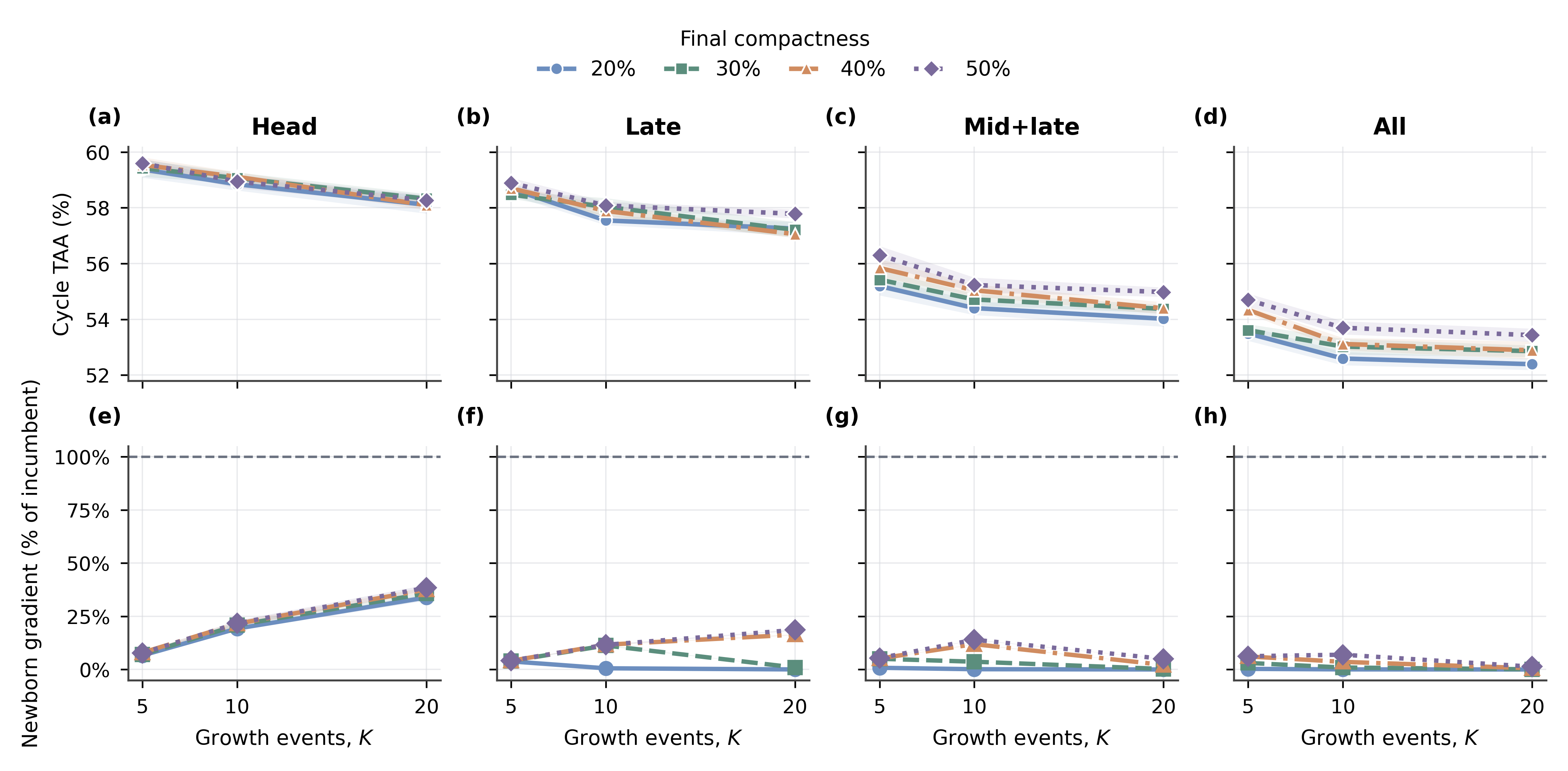}
    \vspace{-2.0em}
    \caption{
        \textbf{ResNet growth-frequency stress test.}
        We vary the number of growth events $K\in\{5,10,20\}$ while fixing the total structural-training budget at $200$ epochs.
        Columns correspond to the \textsc{Head}, \textsc{Late}, \textsc{Mid+Late}, and \textsc{All} edit loci; curves denote final compactness targets.
        \textbf{Top:} Cycle-TAA measures performance throughout the repeated structural-edit trajectory.
        \textbf{Bottom:} newborn gradient signal relative to incumbent units at birth, reported as a percentage of incumbent signal; \(100\%\) denotes equal backward credit. 
        Bands show 95\% confidence intervals
    }
    \label{fig:resnet_growth_frequency_stress}
\end{figure*}

Figure~\ref{fig:resnet_growth_frequency_stress} shows that more frequent growth lowers in-place TAA (Cycle-TAA in the figure), increasingly so as edits extend through the residual hierarchy.
The effect is modest for \textsc{Head} and \textsc{Late}, but stronger for \textsc{Mid+Late} and \textsc{All}.
Newborn gradient signal shows the corresponding integration pressure: head-level newborns approach incumbent levels with more frequent edits, whereas residual-channel newborns become increasingly disadvantaged at lower compactness.
Thus, frequent growth is most disruptive when edits are broad and channel budgets are small, indicating that residual insertion requires sufficient optimization time to integrate.

\vspace{-0.5em}
\subsubsection{Bounded future reachability}
\label{app:resnet_future_reachability}

As in Sec.~\ref{sec:resnet_locus}, we clone each checkpoint and fully fine-tune it under the CIFAR-100 class-permutation remapping probe, preserving inputs and visual classes while changing their output assignments.
We summarize bounded adaptation with normalized AUC over budget $B$ and final probe accuracy.
We select the probe protocol independently of the structural-training optimizer by sweeping
$\{\textsc{SGD},\textsc{Adam},\textsc{AdamW}\}$,
learning rates $\{10^{-1},10^{-2},10^{-3},10^{-4}\}$, and
$B\in\{600,1200,2400,4800\}$.
\textsc{AdamW} with $\eta=10^{-3}$, $B=2400$, and batch size $256$ yields the most stable and discriminative trajectories and is used throughout.

\begin{figure}[ht]
    \centering
    \includegraphics[width=\linewidth]{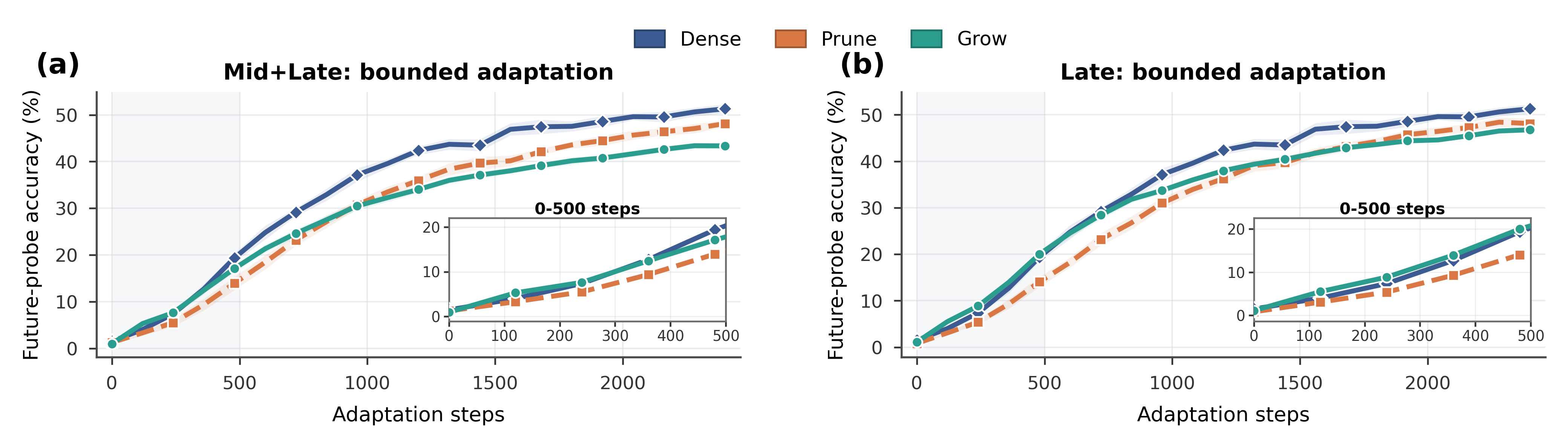}
    \vspace{-2.0em}
    \caption{
        \textbf{Bounded adaptation at intermediate ResNet edit loci.}
        Under both \textsc{Late} and \textsc{Mid+Late} editing, \textsc{Grow} initially matches or exceeds \textsc{Prune}, but \textsc{Prune} subsequently catches up and reaches a higher endpoint. 
        The separation is
        larger under \textsc{Mid+Late}, while \textsc{Dense} retains the strongest overall trajectory. 
        Insets magnify the first $500$ adaptation steps; shaded regions show $95\%$ confidence intervals.
    }
    \label{fig:resnet_intermediate_adaptation}
\end{figure}

\vspace{-0.5em}
\paragraph{Intermediate-loci adaptation trajectories.}
Figure~\ref{fig:resnet_intermediate_adaptation} completes the edit-locus comparison at $50\%$ compactness using the same class-permutation shift and $2400$-step fine-tuning budget.
Under \textsc{Late}, \textsc{Grow} adapts faster than \textsc{Prune} initially, but the trajectories converge and it finishes slightly lower ($47.2\%$ vs.\ $48.1\%$).
Under \textsc{Mid+Late}, this early advantage weakens further and \textsc{Grow} finishes at $44.8\%$, compared with $48.1\%$ for \textsc{Prune} and $51.3\%$ for \textsc{Dense} (Table~\ref{tab:resnet-main-summary}).
Thus, broader feature-hierarchy growth retains some early mobility but reaches a progressively lower endpoint within the fixed adaptation budget.

\vspace{-1.0em}
\section{Activation-Control Benchmark Analysis}
\label{app:activation_control}
\vspace{-0.5em}

The activation-function intervention in Sec.~\ref{sec:integration_interventions} targets a specific mechanistic question: does improving activation-level trainability reduce the newborn integration disadvantage of growth, rather than just boosting performance via changes in insertion, selection, or optimizer-state handling? 
In the main text, Rand.\ Smooth-Leaky was introduced as exactly this intervention, since it modifies the gradient pathway by which new units integrate into a mature network while leaving structural adaptation unchanged.

\vspace{-0.5em}
\paragraph{Protocol}
For each benchmark and method, we apply the same procedure with two activations: standard ReLU and Rand.\ Smooth-Leaky (RSL).
We report the signed difference
\[
\Delta \;=\; \mathrm{RSL} - \mathrm{ReLU},
\]
so positive values mean RSL performs better.
The comparison is done separately for each method, benchmark, and compactness level, then aggregated over the compactness levels shown.
To isolate activation effects from learning-rate choice, we first select the best learning rate independently for each activation, then compute $\Delta$.
The full suite spans seven datasets: the five repeated-shift plasticity benchmarks, Split-CIFAR100 as the sequential-accumulation continual-learning setting, and the i.i.d.\ CIFAR-100 control.

\vspace{-0.5em}
\paragraph{Rand.\ Smooth-Leaky hyperparameters.}
For the continual-learning plasticity benchmarks (Fig.~\ref{fig:cl_plasticity_benchmarks}), we used the default Rand.\ Smooth-Leaky hyperparameters from \citet{lillo2025activation}.
For Split-CIFAR100 and the i.i.d.\ CIFAR-100 control, we instead ran a dedicated sweep over the Rand.\ Smooth-Leaky shape parameters, varying \(c\) and \(p\) over
\[
\{0.1,\,0.3,\,0.5,\,0.8,\,1,\,2,\,3,\,4,\,5\},
\]
and combining them with the following lower/upper-bound pairs:
\[
\begin{aligned}
&(0.01,0.05),\ (0.01,1.00),\ (0.125,0.333),\ (0.40,1.00),\ (0.50,1.00),\\
&(0.60,0.80),\ (0.60,1.00),\ (0.70,1.00),\ (0.673,2.673).
\end{aligned}
\]
Sweeps for Split-CIFAR100 and i.i.d.\ CIFAR-100 were performed under the \textsc{Dense} configuration.
Then, we fixed the best-performing Rand.\ Smooth-Leaky configuration found under \textsc{Dense} for both \textsc{Grow} and \textsc{Prune} avoiding any type of method-specific over-tuning and ensuring that differences reflect the structural operators themselves.
Table~\ref{tab:rsl_hparams} summarizes the Rand.\ Smooth-Leaky settings used in each benchmark.

\begin{table}[ht]
\centering
\small
\setlength{\tabcolsep}{4pt}
\renewcommand{\arraystretch}{1.08}
\begin{tabular}{lcccc|c}
\toprule
\textbf{Benchmark} & \(\mathbf{c}\) & \(\mathbf{p}\) & \textbf{Lower} & \textbf{Upper} & \textbf{LR} \\
\midrule
Permuted MNIST        & 0.8 & 1.0 & 0.3   & 0.6   & 0.001 \\
Random-Label MNIST    & 2.0 & 0.8 & 0.3   & 0.6   & 0.001 \\
Random-Label CIFAR    & 0.8 & 3.0 & 0.5   & 0.5   & 0.001 \\
5+1 CIFAR             & 0.5 & 0.5 & 0.673 & 2.673 & 0.001 \\
Continual ImageNet    & 0.5 & 0.5 & 0.3   & 0.3   & 0.001 \\
Split-CIFAR100        & 0.3 & 0.1 & 0.673 & 2.673 & 0.001 \\
CIFAR-100 (i.i.d.)    & 0.1 & 0.1 & 0.673 & 2.673 & 0.001 \\
\bottomrule
\end{tabular}
    \caption{
    \textbf{Rand.\ Smooth-Leaky hyperparameters used in the activation-control analysis.}
    For the main continual-learning plasticity benchmarks, settings are inherited from \citet{lillo2025activation}. 
    For Split-CIFAR100 and the i.i.d.\ controls, we ran a dedicated sweep over \(c\), \(p\), and lower/upper bounds; Split-CIFAR100 and i.i.d.\ CIFAR-100 report the selected best configuration.}
\label{tab:rsl_hparams}
\end{table}

\vspace{-0.5em}
\paragraph{Results and interpretation}
Table~\ref{tab:activation_delta_core_resnet} shows that Rand.\ Smooth-Leaky is a regime-sensitive rather than uniformly stronger activation.
Its largest gains are often concentrated in \textsc{Grow}, including +35.14 on \texttt{5+1 CIFAR}, +11.91 on \texttt{Continual ImageNet}, +81.64 on \texttt{Random-Label CIFAR}, +61.38 on \texttt{Random-Label MNIST}, and +12.42 on \texttt{Split-CIFAR100}.
On \texttt{Split-CIFAR100}, gains also persist in final accuracy (+14.56) and early-task adaptation (+4.10).
RSL does not rescue every regime: on i.i.d.\ \texttt{CIFAR-100}, average accuracy declines for all methods.
Overall, these results suggest that RSL is most beneficial when newly available capacity must become useful quickly, rather than providing a generic activation advantage.

\begin{table*}[t]
    \centering
    \scriptsize
    \setlength{\tabcolsep}{6pt}
    \renewcommand{\arraystretch}{1.08}
    \begin{tabular}{@{}llrrr@{}}
        \toprule
        Benchmark / locus & Method
        & $\Delta$ Avg.\ ACC.
        & $\Delta$ Final ACC.
        & $\Delta$ Early TAA \\
        \midrule

        \texttt{5+1 CIFAR}
        & \textsc{Dense} & +29.61 & +2.08 & \textbf{+12.99} \\
        & \textsc{Prune} & +3.60 & +0.03 & -1.20 \\
        & \textsc{Grow}  & \textbf{+35.14} & \textbf{+3.32} & +4.38 \\

        \addlinespace[2pt]
        \texttt{CIFAR-100} (Head)
        & \textsc{Dense} & \textbf{-3.03} & \textbf{-0.92} & -3.03 \\
        & \textsc{Prune} & -11.30 & -8.43 & -11.30 \\
        & \textsc{Grow}  & -4.99 & -1.00 & \textbf{-0.62} \\

        \addlinespace[2pt]
        \texttt{Continual ImageNet}
        & \textsc{Dense} & -2.18 & -0.13 & \textbf{+1.66} \\
        & \textsc{Prune} & +2.39 & +0.08 & +0.00 \\
        & \textsc{Grow}  & \textbf{+11.91} & \textbf{+0.10} & +0.04 \\

        \addlinespace[2pt]
        \texttt{Permuted MNIST}
        & \textsc{Dense} & \textbf{+5.43} & -2.91 & -0.99 \\
        & \textsc{Prune} & -0.74 & \textbf{-0.59} & \textbf{-0.44} \\
        & \textsc{Grow}  & +4.79 & -1.49 & -2.61 \\

        \addlinespace[2pt]
        \texttt{Random-Label CIFAR}
        & \textsc{Dense} & +67.30 & +1.75 & +3.43 \\
        & \textsc{Prune} & +1.41 & -0.04 & +0.04 \\
        & \textsc{Grow}  & \textbf{+81.64} & \textbf{+1.79} & \textbf{+6.76} \\

        \addlinespace[2pt]
        \texttt{Random-Label MNIST}
        & \textsc{Dense} & +57.07 & \textbf{+1.72} & +4.07 \\
        & \textsc{Prune} & -2.30 & -0.01 & -0.06 \\
        & \textsc{Grow}  & \textbf{+61.38} & +1.67 & \textbf{+6.27} \\

        \addlinespace[2pt]
        \texttt{Split-CIFAR100} (Head)
        & \textsc{Dense} & +10.76 & +11.63 & -0.63 \\
        & \textsc{Prune} & +5.89 & +5.75 & +3.49 \\
        & \textsc{Grow}  & \textbf{+12.42} & \textbf{+14.56} & \textbf{+4.10} \\

        \bottomrule
    \end{tabular}

    \caption{
        \textbf{Effect of Randomized Smooth-Leaky relative to ReLU.}
        Entries report $\Delta=\mathrm{RSL}-\mathrm{ReLU}$, averaged over compactness after learning-rate selection within each activation.
        Positive values favor RSL; bold denotes the largest delta within each benchmark and metric. \texttt{CIFAR-100} and \texttt{Split-CIFAR100} use the ResNet Head locus.
    }
    \label{tab:activation_delta_core_resnet}
\end{table*}

\vspace{-0.5em}
\subsection{ReLU vs.\ Rand.\ Smooth-Leaky Newborn Integration}
\label{app:rsl_newborn_integration}

As we just discussed, Rand.\ Smooth-Leaky often improves \textsc{Grow} in regimes where rapid adaptation is difficult. 
Next, we ask whether this performance gain is accompanied by the mechanistic consequences predicted by our main analysis: improved newborn integration after a growth event. 
To test this, we compare ReLU and Rand.\ Smooth-Leaky in the controlled CIFAR-100 \textsc{Grow} setting using the same newborn-to-incumbent signal diagnostics introduced in Sec.~\ref{sec:allocation_mechanisms}.

\begin{figure*}[ht]
    \centering
    \includegraphics[width=0.9\linewidth]{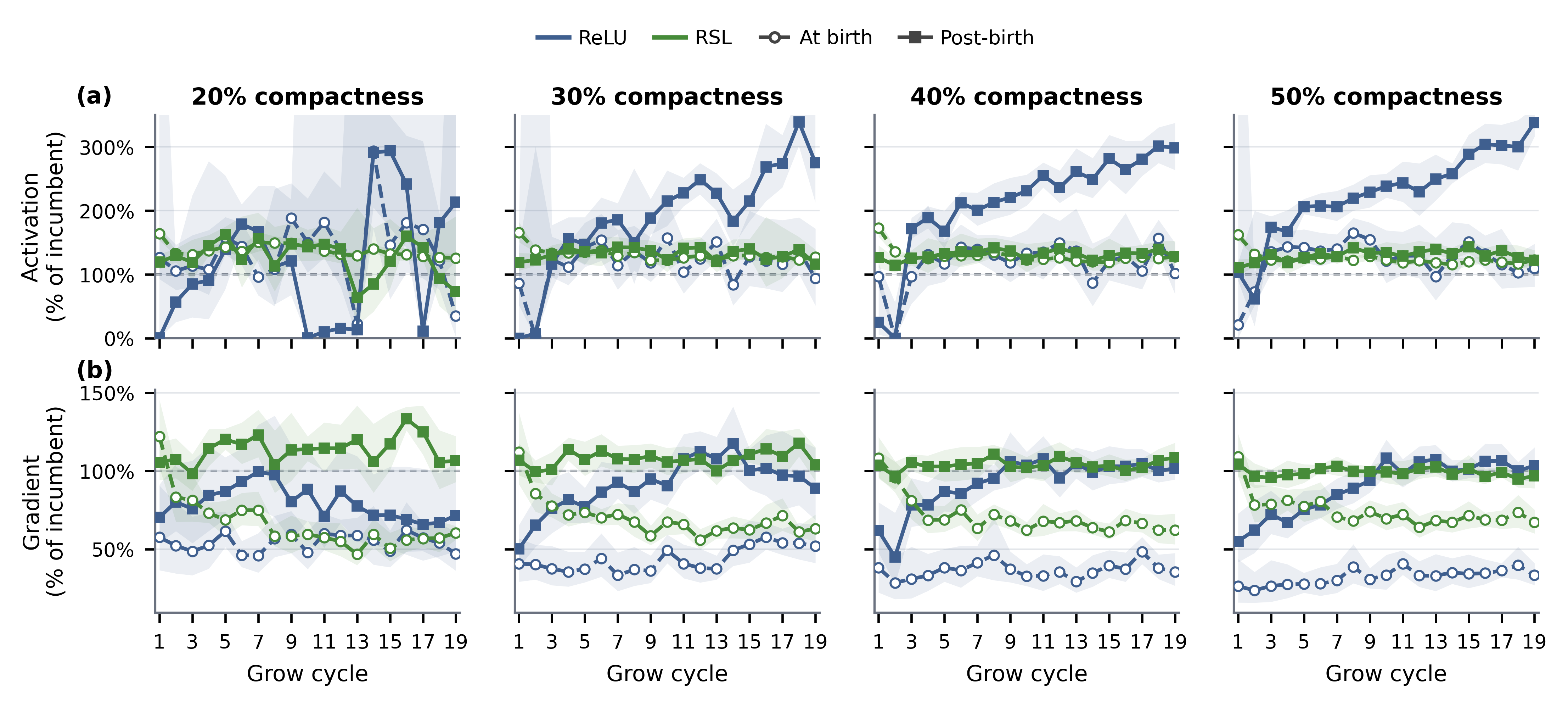}
    \vspace{-1em}
    \caption{
    \textbf{Rand.\ Smooth-Leaky improves newborn integration in Split-CIFAR100 Head GROW.}
    Newborn activation (a) and gradient signal (b) are reported relative to previously active units across compactness and growth cycles; $100\%$ denotes equal per-unit signal.
    Under ReLU, newborn integration is less stable, with large early activation deviations and substantially weaker birth-time gradients.
    Rand.\ Smooth-Leaky keeps forward participation closer to the incumbent level and increases the gradient signal available to newborn units, especially at birth.
    }
    \label{fig:rsl_newborn_integration}
\end{figure*}

\paragraph{Interpretation}
Figure~\ref{fig:rsl_newborn_integration} supports the activation-intervention account much more clearly in Split-CIFAR100 than in i.i.d.\ CIFAR-100.
Rand.\ Smooth-Leaky keeps newborn units better coupled to the forward computation and, more importantly, increases their backward signal relative to incumbent units, especially at birth.
Under ReLU, newborn activation is more variable and birth-time gradients remain substantially below the incumbent level.
This matches the main-text result: in a plasticity-stressing regime, improving activation design helps \textsc{Grow} by making newly added units easier to integrate into ongoing learning.                   
\vspace{-0.5em}
\section{Early-Task Plasticity Under Repeated Shift}
\label{sec:app_cl_early_task_auc}
\vspace{-0.5em}

\begin{figure*}[ht]
    \centering
    \includegraphics[width=\linewidth]{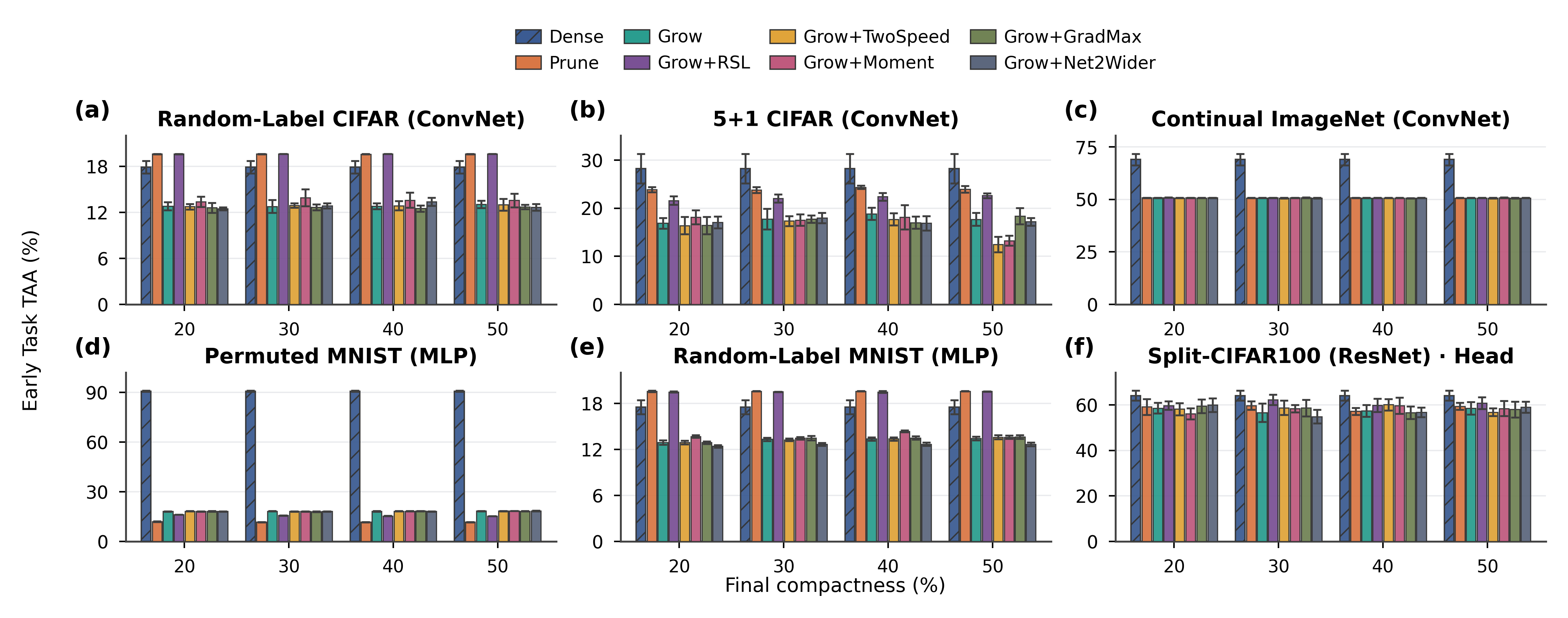}
    \vspace{-1.0em}
    \caption{
    \textbf{Early-task plasticity across sequential-accumulation and repeated-shift benchmarks.}
    We report \emph{Early Task TAA}, defined as the average accuracy over only the initial portion of each task, emphasizing immediate post-shift adaptation rather than late within-task convergence.
    \textsc{Dense} or \textsc{Prune}-based methods retain the strongest early-window performance when adaptation time is severely limited, whereas growth-based methods improve when insertion is made more integration-friendly.
    \textsc{Grow+Rand.\ Smooth-Leaky} is the most consistently effective growth-family intervention, substantially improving over vanilla \textsc{Grow}.
    These results support the view that the utility of growth depends critically on how quickly newborn units can integrate after insertion.
    }
    \label{fig:cl_early_task_auc}
\end{figure*}

We additionally report \emph{Early Task TAA}, the average accuracy over the initial portion of each task, measuring immediate post-shift plasticity before substantial within-task adaptation.
Figure~\ref{fig:cl_early_task_auc} shows a clear dependence on adaptation time.
Under rapid shifts, especially Permuted MNIST and Continual ImageNet, \textsc{Dense} retains a strong early advantage, consistent with mature capacity being immediately usable while newborn units still integrate.
On Random-Label MNIST and CIFAR, \textsc{Grow+Rand.\ Smooth-Leaky} substantially improves over vanilla \textsc{Grow} and is the strongest growth variant, while other interventions provide smaller or less consistent gains (App.~\ref{app:activation_control}).
Under the longer optimization intervals of Split-CIFAR100, growth becomes more competitive, particularly with \textsc{Rand.\ Smooth-Leaky}.
Overall, growth depends critically on whether newborn units become usable before the learning signal changes again.

\begin{table*}[ht]
\centering
\scriptsize
\renewcommand{\arraystretch}{1.08}
\setlength{\tabcolsep}{3.0pt}
\resizebox{\linewidth}{!}{%
\begin{tabular}{llcccccccccccc}
\toprule
\textbf{Comp.} & \textbf{Method} & \multicolumn{2}{c}{\textbf{Random-Label CIFAR}} & \multicolumn{2}{c}{\textbf{5+1 CIFAR}} & \multicolumn{2}{c}{\textbf{Continual ImageNet}} & \multicolumn{2}{c}{\textbf{Permuted MNIST}} & \multicolumn{2}{c}{\textbf{Random-Label MNIST}} & \multicolumn{2}{c}{\textbf{Split-CIFAR100}} \\
\cmidrule(lr){3-4}\cmidrule(lr){5-6}\cmidrule(lr){7-8}\cmidrule(lr){9-10}\cmidrule(lr){11-12}\cmidrule(lr){13-14}
 &  & \textbf{Avg.} & \textbf{Early} & \textbf{Avg.} & \textbf{Early} & \textbf{Avg.} & \textbf{Early} & \textbf{Avg.} & \textbf{Early} & \textbf{Avg.} & \textbf{Early} & \textbf{Avg.} & \textbf{Early} \\
\midrule
\textbf{20\%} & \textbf{Dense} & 57.5 $\pm$ 15.2 & 17.9 $\pm$ 0.8 & 18.7 $\pm$ 6.7 & \textbf{28.2 $\pm$ 3.1} & 72.6 $\pm$ 0.5 & \textbf{68.9 $\pm$ 2.8} & 81.5 $\pm$ 1.2 & \textbf{90.8 $\pm$ 0.3} & 48.7 $\pm$ 12.8 & 17.5 $\pm$ 0.9 & 28.5 $\pm$ 1.0 & \textbf{64.1 $\pm$ 2.2} \\
 & \textbf{Prune} & \textbf{97.9 $\pm$ 0.4} & 19.6 $\pm$ 0.0 & \textbf{73.8 $\pm$ 1.1} & 23.9 $\pm$ 0.6 & \textbf{91.8 $\pm$ 0.9} & 50.7 $\pm$ 0.1 & \textbf{85.9 $\pm$ 0.1} & 12.0 $\pm$ 0.2 & \textbf{94.4 $\pm$ 0.5} & \textbf{19.6 $\pm$ 0.1} & 24.4 $\pm$ 0.7 & 59.1 $\pm$ 3.5 \\
 & \textbf{Grow} & 13.5 $\pm$ 0.9 & 12.8 $\pm$ 0.5 & 23.7 $\pm$ 3.8 & 16.9 $\pm$ 1.1 & 70.7 $\pm$ 1.6 & 50.7 $\pm$ 0.1 & 71.5 $\pm$ 0.2 & 18.1 $\pm$ 0.1 & 16.3 $\pm$ 1.4 & 12.9 $\pm$ 0.3 & 24.5 $\pm$ 1.0 & 58.6 $\pm$ 2.3 \\
 & \textbf{Grow+Rand. Smooth-Leaky} & 94.4 $\pm$ 0.2 & \textbf{19.6 $\pm$ 0.0} & 63.6 $\pm$ 0.9 & 21.6 $\pm$ 0.9 & 81.5 $\pm$ 0.1 & 50.9 $\pm$ 0.1 & 75.1 $\pm$ 0.1 & 16.2 $\pm$ 0.1 & 72.1 $\pm$ 2.3 & 19.5 $\pm$ 0.1 & \textbf{36.2 $\pm$ 0.8} & 59.7 $\pm$ 1.9 \\
 & \textbf{Grow+TwoSpeed} & 13.8 $\pm$ 0.7 & 12.8 $\pm$ 0.4 & 20.8 $\pm$ 4.4 & 16.4 $\pm$ 1.8 & 67.3 $\pm$ 0.6 & 50.7 $\pm$ 0.1 & 71.5 $\pm$ 0.3 & 18.3 $\pm$ 0.2 & 16.1 $\pm$ 1.3 & 12.9 $\pm$ 0.2 & 24.9 $\pm$ 0.6 & 58.2 $\pm$ 2.6 \\
 & \textbf{Grow+Moment} & 15.2 $\pm$ 1.8 & 13.4 $\pm$ 0.7 & 25.0 $\pm$ 2.9 & 18.1 $\pm$ 1.5 & 68.9 $\pm$ 3.2 & 50.8 $\pm$ 0.1 & 72.0 $\pm$ 0.1 & 18.1 $\pm$ 0.1 & 17.7 $\pm$ 0.7 & 13.7 $\pm$ 0.2 & 24.4 $\pm$ 1.0 & 56.1 $\pm$ 2.4 \\
 & \textbf{Grow+GradMax} & 13.7 $\pm$ 1.5 & 12.6 $\pm$ 0.6 & 24.4 $\pm$ 4.4 & 16.4 $\pm$ 1.8 & 71.6 $\pm$ 2.8 & 50.7 $\pm$ 0.0 & 71.5 $\pm$ 0.3 & 18.2 $\pm$ 0.2 & 16.2 $\pm$ 0.9 & 12.9 $\pm$ 0.2 & 24.9 $\pm$ 1.0 & 59.5 $\pm$ 3.0 \\
 & \textbf{Grow+Net2Wider} & 13.1 $\pm$ 0.5 & 12.5 $\pm$ 0.2 & 23.0 $\pm$ 3.9 & 17.1 $\pm$ 1.2 & 70.7 $\pm$ 3.2 & 50.8 $\pm$ 0.1 & 71.3 $\pm$ 0.2 & 18.1 $\pm$ 0.1 & 14.8 $\pm$ 0.8 & 12.4 $\pm$ 0.2 & 22.3 $\pm$ 0.9 & 59.9 $\pm$ 3.0 \\
\midrule
\textbf{30\%} & \textbf{Dense} & 57.5 $\pm$ 15.2 & 17.9 $\pm$ 0.8 & 18.7 $\pm$ 6.7 & \textbf{28.2 $\pm$ 3.1} & 72.6 $\pm$ 0.5 & \textbf{68.9 $\pm$ 2.8} & 81.5 $\pm$ 1.2 & \textbf{90.8 $\pm$ 0.3} & 48.7 $\pm$ 12.8 & 17.5 $\pm$ 0.9 & 28.5 $\pm$ 1.0 & \textbf{64.1 $\pm$ 2.2} \\
 & \textbf{Prune} & \textbf{97.8 $\pm$ 0.4} & 19.6 $\pm$ 0.0 & \textbf{74.6 $\pm$ 1.0} & 23.8 $\pm$ 0.6 & \textbf{93.3 $\pm$ 0.6} & 50.7 $\pm$ 0.1 & \textbf{86.4 $\pm$ 0.1} & 11.7 $\pm$ 0.1 & \textbf{94.7 $\pm$ 0.5} & \textbf{19.6 $\pm$ 0.0} & 24.6 $\pm$ 0.6 & 59.7 $\pm$ 1.8 \\
 & \textbf{Grow} & 14.9 $\pm$ 2.3 & 12.8 $\pm$ 0.9 & 29.4 $\pm$ 4.3 & 17.7 $\pm$ 2.2 & 71.9 $\pm$ 2.1 & 50.7 $\pm$ 0.1 & 73.3 $\pm$ 0.2 & 18.2 $\pm$ 0.1 & 19.4 $\pm$ 1.5 & 13.3 $\pm$ 0.2 & 24.9 $\pm$ 0.8 & 56.6 $\pm$ 4.0 \\
 & \textbf{Grow+Rand. Smooth-Leaky} & 96.1 $\pm$ 0.1 & \textbf{19.6 $\pm$ 0.0} & 65.3 $\pm$ 1.0 & 22.1 $\pm$ 0.8 & 82.8 $\pm$ 0.1 & 50.8 $\pm$ 0.1 & 78.1 $\pm$ 0.0 & 15.6 $\pm$ 0.0 & 83.4 $\pm$ 1.2 & 19.5 $\pm$ 0.0 & \textbf{38.1 $\pm$ 0.9} & 62.3 $\pm$ 2.3 \\
 & \textbf{Grow+TwoSpeed} & 15.4 $\pm$ 2.1 & 12.9 $\pm$ 0.3 & 28.3 $\pm$ 1.7 & 17.3 $\pm$ 1.0 & 69.3 $\pm$ 0.1 & 50.7 $\pm$ 0.2 & 73.3 $\pm$ 0.2 & 18.2 $\pm$ 0.1 & 20.1 $\pm$ 1.6 & 13.3 $\pm$ 0.2 & 26.1 $\pm$ 1.1 & 58.8 $\pm$ 3.1 \\
 & \textbf{Grow+Moment} & 19.5 $\pm$ 5.1 & 13.9 $\pm$ 1.1 & 35.1 $\pm$ 2.8 & 17.5 $\pm$ 1.2 & 72.2 $\pm$ 1.7 & 50.7 $\pm$ 0.1 & 74.0 $\pm$ 0.1 & 18.2 $\pm$ 0.1 & 20.3 $\pm$ 1.4 & 13.5 $\pm$ 0.2 & 25.4 $\pm$ 0.7 & 58.4 $\pm$ 1.6 \\
 & \textbf{Grow+GradMax} & 13.9 $\pm$ 1.8 & 12.7 $\pm$ 0.4 & 31.8 $\pm$ 2.3 & 17.7 $\pm$ 0.8 & 69.9 $\pm$ 0.2 & 50.8 $\pm$ 0.1 & 73.4 $\pm$ 0.3 & 18.1 $\pm$ 0.2 & 20.6 $\pm$ 1.6 & 13.5 $\pm$ 0.3 & 25.3 $\pm$ 1.0 & 58.7 $\pm$ 3.6 \\
 & \textbf{Grow+Net2Wider} & 15.1 $\pm$ 1.5 & 12.9 $\pm$ 0.3 & 30.6 $\pm$ 3.0 & 18.0 $\pm$ 1.1 & 67.9 $\pm$ 0.3 & 50.7 $\pm$ 0.1 & 73.1 $\pm$ 0.2 & 18.1 $\pm$ 0.2 & 16.5 $\pm$ 1.1 & 12.7 $\pm$ 0.2 & 23.2 $\pm$ 0.8 & 54.9 $\pm$ 3.0 \\
\midrule
\textbf{40\%} & \textbf{Dense} & 57.5 $\pm$ 15.2 & 17.9 $\pm$ 0.8 & 18.7 $\pm$ 6.7 & \textbf{28.2 $\pm$ 3.1} & 72.6 $\pm$ 0.5 & \textbf{68.9 $\pm$ 2.8} & 81.5 $\pm$ 1.2 & \textbf{90.8 $\pm$ 0.3} & 48.7 $\pm$ 12.8 & 17.5 $\pm$ 0.9 & 28.5 $\pm$ 1.0 & \textbf{64.1 $\pm$ 2.2} \\
 & \textbf{Prune} & \textbf{98.0 $\pm$ 0.3} & 19.6 $\pm$ 0.0 & \textbf{75.4 $\pm$ 0.9} & 24.4 $\pm$ 0.3 & \textbf{94.0 $\pm$ 0.4} & 50.7 $\pm$ 0.1 & \textbf{86.7 $\pm$ 0.2} & 11.7 $\pm$ 0.1 & \textbf{94.9 $\pm$ 0.5} & \textbf{19.6 $\pm$ 0.0} & 24.3 $\pm$ 0.4 & 57.2 $\pm$ 1.5 \\
 & \textbf{Grow} & 14.1 $\pm$ 0.6 & 12.8 $\pm$ 0.4 & 33.8 $\pm$ 4.3 & 18.8 $\pm$ 1.3 & 70.4 $\pm$ 0.2 & 50.7 $\pm$ 0.1 & 74.7 $\pm$ 0.2 & 18.2 $\pm$ 0.1 & 21.1 $\pm$ 2.0 & 13.4 $\pm$ 0.2 & 25.3 $\pm$ 1.0 & 57.3 $\pm$ 2.6 \\
 & \textbf{Grow+Rand. Smooth-Leaky} & 96.8 $\pm$ 0.1 & \textbf{19.6 $\pm$ 0.0} & 65.0 $\pm$ 0.8 & 22.4 $\pm$ 0.8 & 83.5 $\pm$ 0.1 & 50.7 $\pm$ 0.1 & 80.0 $\pm$ 0.0 & 15.4 $\pm$ 0.0 & 84.1 $\pm$ 0.9 & 19.5 $\pm$ 0.1 & \textbf{36.9 $\pm$ 1.2} & 59.8 $\pm$ 2.9 \\
 & \textbf{Grow+TwoSpeed} & 14.6 $\pm$ 1.4 & 12.9 $\pm$ 0.6 & 30.2 $\pm$ 2.8 & 17.7 $\pm$ 1.3 & 70.6 $\pm$ 0.2 & 50.7 $\pm$ 0.1 & 75.0 $\pm$ 0.2 & 18.2 $\pm$ 0.1 & 21.0 $\pm$ 1.8 & 13.4 $\pm$ 0.2 & 25.9 $\pm$ 0.8 & 60.1 $\pm$ 2.5 \\
 & \textbf{Grow+Moment} & 17.4 $\pm$ 5.9 & 13.6 $\pm$ 1.0 & 38.3 $\pm$ 6.3 & 18.2 $\pm$ 2.5 & 70.2 $\pm$ 2.5 & 50.7 $\pm$ 0.1 & 75.0 $\pm$ 0.1 & 18.3 $\pm$ 0.1 & 22.4 $\pm$ 1.2 & 14.4 $\pm$ 0.1 & 26.1 $\pm$ 0.8 & 59.7 $\pm$ 3.5 \\
 & \textbf{Grow+GradMax} & 13.6 $\pm$ 0.7 & 12.5 $\pm$ 0.4 & 30.0 $\pm$ 2.4 & 17.0 $\pm$ 1.3 & 70.5 $\pm$ 0.3 & 50.6 $\pm$ 0.1 & 74.6 $\pm$ 0.2 & 18.3 $\pm$ 0.1 & 22.2 $\pm$ 2.0 & 13.5 $\pm$ 0.2 & 25.2 $\pm$ 0.7 & 56.6 $\pm$ 2.7 \\
 & \textbf{Grow+Net2Wider} & 18.0 $\pm$ 2.7 & 13.4 $\pm$ 0.5 & 31.8 $\pm$ 4.4 & 16.9 $\pm$ 1.5 & 68.5 $\pm$ 0.3 & 50.7 $\pm$ 0.1 & 74.5 $\pm$ 0.2 & 18.1 $\pm$ 0.1 & 17.5 $\pm$ 1.4 & 12.7 $\pm$ 0.2 & 24.1 $\pm$ 0.8 & 56.8 $\pm$ 2.1 \\
\midrule
\textbf{50\%} & \textbf{Dense} & 57.5 $\pm$ 15.2 & 17.9 $\pm$ 0.8 & 18.7 $\pm$ 6.7 & \textbf{28.2 $\pm$ 3.1} & 72.6 $\pm$ 0.5 & \textbf{68.9 $\pm$ 2.8} & 81.5 $\pm$ 1.2 & \textbf{90.8 $\pm$ 0.3} & 48.7 $\pm$ 12.8 & 17.5 $\pm$ 0.9 & 28.5 $\pm$ 1.0 & \textbf{64.1 $\pm$ 2.2} \\
 & \textbf{Prune} & \textbf{98.1 $\pm$ 0.3} & 19.6 $\pm$ 0.0 & \textbf{75.5 $\pm$ 0.8} & 23.9 $\pm$ 0.7 & \textbf{94.5 $\pm$ 0.3} & 50.7 $\pm$ 0.1 & \textbf{86.9 $\pm$ 0.3} & 11.6 $\pm$ 0.1 & \textbf{95.1 $\pm$ 0.5} & \textbf{19.6 $\pm$ 0.0} & 24.7 $\pm$ 0.3 & 59.4 $\pm$ 1.5 \\
 & \textbf{Grow} & 15.4 $\pm$ 2.4 & 13.0 $\pm$ 0.5 & 34.4 $\pm$ 3.7 & 17.7 $\pm$ 1.3 & 71.3 $\pm$ 0.3 & 50.7 $\pm$ 0.1 & 75.7 $\pm$ 0.1 & 18.3 $\pm$ 0.1 & 23.2 $\pm$ 2.3 & 13.4 $\pm$ 0.2 & 26.0 $\pm$ 0.8 & 58.6 $\pm$ 2.7 \\
 & \textbf{Grow+Rand. Smooth-Leaky} & 97.1 $\pm$ 0.1 & \textbf{19.6 $\pm$ 0.0} & 67.9 $\pm$ 0.5 & 22.6 $\pm$ 0.5 & 84.2 $\pm$ 0.1 & 50.7 $\pm$ 0.1 & 81.3 $\pm$ 0.0 & 15.2 $\pm$ 0.0 & 85.9 $\pm$ 0.7 & 19.6 $\pm$ 0.0 & \textbf{37.8 $\pm$ 1.2} & 60.8 $\pm$ 2.6 \\
 & \textbf{Grow+TwoSpeed} & 15.8 $\pm$ 3.4 & 13.0 $\pm$ 0.8 & 29.3 $\pm$ 1.3 & 12.5 $\pm$ 1.6 & 71.2 $\pm$ 0.3 & 50.6 $\pm$ 0.1 & 75.6 $\pm$ 0.2 & 18.4 $\pm$ 0.1 & 23.6 $\pm$ 1.8 & 13.6 $\pm$ 0.2 & 25.6 $\pm$ 0.8 & 56.8 $\pm$ 1.7 \\
 & \textbf{Grow+Moment} & 17.3 $\pm$ 3.5 & 13.6 $\pm$ 0.9 & 31.6 $\pm$ 0.7 & 13.3 $\pm$ 1.0 & 70.8 $\pm$ 0.2 & 50.8 $\pm$ 0.1 & 76.1 $\pm$ 0.2 & 18.4 $\pm$ 0.1 & 23.7 $\pm$ 2.0 & 13.6 $\pm$ 0.2 & 26.0 $\pm$ 1.2 & 58.4 $\pm$ 3.3 \\
 & \textbf{Grow+GradMax} & 13.8 $\pm$ 0.6 & 12.7 $\pm$ 0.3 & 33.7 $\pm$ 5.2 & 18.4 $\pm$ 1.7 & 71.5 $\pm$ 0.3 & 50.6 $\pm$ 0.1 & 75.5 $\pm$ 0.2 & 18.3 $\pm$ 0.2 & 24.3 $\pm$ 2.4 & 13.6 $\pm$ 0.2 & 25.6 $\pm$ 0.9 & 58.0 $\pm$ 3.5 \\
 & \textbf{Grow+Net2Wider} & 14.8 $\pm$ 2.4 & 12.7 $\pm$ 0.4 & 30.9 $\pm$ 3.9 & 17.2 $\pm$ 0.8 & 69.6 $\pm$ 0.3 & 50.7 $\pm$ 0.1 & 75.4 $\pm$ 0.1 & 18.4 $\pm$ 0.1 & 18.5 $\pm$ 1.7 & 12.6 $\pm$ 0.2 & 24.8 $\pm$ 0.5 & 59.0 $\pm$ 2.5 \\
\bottomrule
\end{tabular}%
}
\vspace{-0.5em}
\caption{
\textbf{Continual-learning summary (cycle)}. 
For each dataset we report Avg.\ Acc.\ and Early Task TAA. Within each compactness block, the best value for each dataset--metric pair is shown in bold. 
Dense is shown as the full-capacity reference within each compactness block, while non-dense methods are compared at matched compactness. 
Dense uses a dataset-specific fixed learning rate; all other methods select the best learning rate by Avg.\ Acc.\ among $\{0.01, 0.001, 0.0001\}$.}
\label{tab:cl_all_benchmarks_summary}
\end{table*}         

\section{The Use of Large Language Models (LLMs)}
We used an LLM as a tool for early brainstorming, code debugging, and writing/editing of the earlier drafts of this paper. During ideation, we used it to enumerate experiment variants and sanity-check design choices; for code, we requested bug-finding hints and refactoring suggestions that we implemented only after manual review and testing; for text, we used it to improve clarity, organization, and grammar. The LLM did not generate novel research ideas, experiments, or results on our behalf; all methodological innovations, analyses, and conclusions are our own. We verified any technical claims, equations, and citations suggested during assisted drafting and we did not include uncited, model-generated text verbatim. No proprietary or personally identifiable data was provided to the LLM. The authors retain full responsibility for the content of this paper, and we affirm the novelty and originality of the work.

\end{document}